\documentclass[acmtog]{acmart}
\acmSubmissionID{475}

\newcommand{\final}{0}

\usepackage{hyperref}
\usepackage{epsfig}
\usepackage{graphicx}
\usepackage{amsmath}
\usepackage{ifthen}
\usepackage{float}
\usepackage{morefloats}
\usepackage{alltt}
\usepackage{amsmath}
\usepackage{amsthm}
\usepackage{ulem}
\usepackage{mathrsfs}
\usepackage[utf8]{inputenc}
\usepackage{newlfont} 
\usepackage{floatflt}
\usepackage{wrapfig}

\usepackage[ruled]{algorithm2e} 

\SetAlFnt{\small}
\SetAlCapFnt{\small}
\SetAlCapNameFnt{\small}
\SetAlCapHSkip{0pt}

\usepackage{booktabs} 
\usepackage{gensymb}
\usepackage{subfig} 
\usepackage{multirow}
\usepackage{cleveref}
\Crefname{equation}{Eq.}{Eqs.}
\Crefname{figure}{Fig.}{Figs.}
\Crefname{section}{Sec.}{Sec.}
\usepackage[T1]{fontenc} 
\usepackage{etoolbox}
\usepackage{xfrac}
\usepackage{booktabs} 
\usepackage[export]{adjustbox}
\usepackage{algpseudocode}
\usepackage[acronym,nomain]{glossaries}
\usepackage{adjustbox}
\usepackage{xcolor}
\usepackage{wrapfig}
\usepackage[percent]{overpic}
\usepackage{soul}
\usepackage{paralist}
\usepackage{pict2e}

\newcommand{\etal}{{et~al.}}

\newcommand{\eg}{{e.g.,\ }}
\newcommand{\ie}{{i.e.,\ }}
\newcommand{\wide}{\textbf{W}}
\newcommand{\ultrawide}{\textbf{UW}}

\definecolor{YodaColor}{rgb}{0,0.6,0}
\definecolor{ChiakaiColor}{rgb}{0.15, 0.60, 0.16}
\definecolor{YichangColor}{rgb}{0.70, 0.30, 0.20}
\definecolor{JasonColor}{rgb}{0.84 0.31 0.87}
\definecolor{LuncColor}{rgb}{0.7 0.7 0.1}
\definecolor{ReviseColor}{rgb}{0 0 1}
\definecolor{DarkGreen}{rgb}{0.008, 0.294, 0.188}

\newif\ifdrafting
\draftingtrue 
\ifdrafting
    \newcommand{\ds}[1]{{\color{red}[Deqing: #1]}}
    
    \newcommand{\ckliang}[1]{{\color{ChiakaiColor} [Chia-Kai: #1]}}
    \newcommand{\yichang}[1]{{\color{YichangColor} [Yichang: #1]}}
    \newcommand{\wslai}[1]{{\color{JasonColor} [Jason: #1]}}
    
    \newcommand{\reviewer}[2][]{{\color{YodaColor} [Reviewer#1: #2]}}
\else
	\newcommand{\ds} [1] {}
	
    \newcommand{\ckliang}[1]{}
    \newcommand{\yichang}[1]{}
    \newcommand{\wslai}[1]{}
    \newcommand{\reviewer}[2][]{}
\fi

\newcommand{\warning}[1]{{\it\color{red} #1}}
\newcommand{\note}[1]{{\it\color{blue} #1}}
\newcommand{\nothing}[1]{}

\definecolor{AudioColor}{rgb}{0, 0, 0}
\newcommand{\audio}[2]{{\color{AudioColor} [#1] #2 $\qed$}}
\definecolor{VideoColor}{rgb}{0.44,0.66,0.38}
\newcommand{\video}[1]{{\color{VideoColor} Video: #1 $\qed$}}
\definecolor{DeadlineColor}{rgb}{0.9,0.4,0}
\newcommand{\deadline}[1]{{\bf\color{DeadlineColor} ETA: #1}}

\definecolor{OldColor}{rgb}{0.5,0.5,0.5}
\newcommand{\old}[1]{{\color{OldColor} #1}}
\definecolor{NewColor}{rgb}{0.9,0.4,0}

\definecolor{DeleteColor}{rgb}{0.1,0.6,1.0}
\newcommand{\delete}[1]{{\color{DeleteColor} #1}}
\definecolor{MoveColor}{rgb}{0.5,0.1,0.5}

\definecolor{figred}{rgb}{1,0,0}
\definecolor{figgreen}{rgb}{0,0.6,0}
\definecolor{figblue}{rgb}{0,0,1}
\definecolor{figpink}{rgb}{1,0.63,0.63}

\ifthenelse{\equal{\final}{1}}
{
\renewcommand{\ckliang}[1]{}
\renewcommand{\yichang}[1]{}
\renewcommand{\minghsuan}[1]{}
\renewcommand{\wslai}[1]{}
\renewcommand{\reviewer}[2][]{}

\renewcommand{\warning}[1]{}
\renewcommand{\note}[1]{}
\renewcommand{\old}[1]{}
\renewcommand{\audio}[2][]{}
\renewcommand{\video}[1]{}
\renewcommand{\deadline}[1]{}
}

\newcommand{\pseudocode}{Pseudocode}
\floatstyle{plain}
\newfloat{algorithm}{tbhp}{lop}
\floatname{algorithm}{\pseudocode}

\ifthenelse{\isundefined{\diff}}
{

\renewcommand{\delete}[1]{}

}
{
\ifthenelse{\equal{\diff}{0}}
{

\renewcommand{\delete}[1]{}

}
{}
}

\ifdefined\google

\else

\fi

\newcommand{\figref}[1]{Figure~\ref{fig:#1}}
\newcommand{\eqnref}[1]{\eqref{eq:#1}}
\newcommand{\secref}[1]{Section~\ref{sec:#1}}

\newcommand{\blue}[1]{\textcolor{blue}{#1}}

\newcommand{\white}[1]{\textcolor{white}{#1}}

\def\srcimage{I_{\text{src}}}
\def\refimage{I_{\text{ref}}}
\def\colormatchedrefimage{\hat{I}_{\text{ref}}}
\def\warpedrefimage{I'_{\text{ref}}}
\def\fusedimage{I_{\text{fused}}}
\def\finalimage{I_{\text{final}}}
\def\gtimage{I_{\text{GT}}}
\def\gammasrcimage{\tilde{I}_{\text{src}}}
\def\gammafusedimage{\tilde{I}_{\text{fused}}}
\def\srcccm{CCM_{\text{src}}}
\def\refccm{CCM_{\text{ref}}}
\def\forwardflow{F_{\text{fwd}}}
\def\backwardflow{F_{\text{bwd}}}
\def\facemask{M_{\text{face}}}
\def\occlusionmask{M_{\text{occ}}}
\def\reprojectionmask{M_{\text{reproj}}}
\def\blendingmask{M_{\text{blending}}}
\def\blurkernel{k}
\def\blurmask{M_{\text{blur}}}
\def\noise{n}

\def\pwcnet{\text{PWCNet}}
\def\fusionnet{\text{FusionNet}}
\def\warp{\mathbb{W}}
\newcommand{\onenorm}[1]{||#1||_1}
\newcommand{\twonorm}[1]{||#1||^2_2}

\def\contentloss{\mathcal{L}_{\text{content}}}
\def\vggloss{\mathcal{L}_{\text{vgg}}}

\def\colorloss{\mathcal{L}_{\text{color}}}
\def\contentweight{w_{\text{content}}}
\def\vggweight{w_{\text{vgg}}}

\def\colorweight{w_{\text{color}}}

\def\imagewidth{W}
\def\imageheight{H}

\def\website{\blue{\url{https://www.wslai.net/publications/fusion_deblur}}}

\setcopyright{none} 
\begin{document}
\acmJournal{TOG}
\acmYear{2022}\acmVolume{41}\acmNumber{4}\acmArticle{148}\acmMonth{7}
\acmDOI{10.1145/3528223.3530131}

\normalem
\title{Face Deblurring using Dual Camera Fusion on Mobile Phones}

\author{Wei-Sheng Lai}
\email{wslai@google.com}
\author{YiChang Shih}
\email{yichang@google.com}
\author{Lun-Cheng Chu}
\email{lunc@google.com}
\author{Xiaotong Wu}
\email{abbywu@google.com}
\author{Sung-Fang Tsai}
\email{sungfang@google.com}
\author{Michael Krainin}
\email{mkrainin@google.com}
\author{Deqing Sun}
\email{deqingsun@google.com}
\author{Chia-Kai Liang}
\email{ckliang@google.com}
\affiliation{%
 \institution{Google}
 \country{USA}
}

\renewcommand\shortauthors{Lai, et al}

\begin{abstract}
Motion blur of fast-moving subjects is a longstanding problem in photography and very common on mobile phones due to limited light collection efficiency, particularly in low-light conditions. 
While we have witnessed great progress in image deblurring in recent years, most methods require significant computational power and have limitations in processing high-resolution photos with severe local motions.
To this end, we develop a novel face deblurring system based on the dual camera fusion technique for mobile phones. 
The system detects subject motion to dynamically enable a reference camera, \eg ultrawide angle camera commonly available on recent premium phones, and captures an auxiliary photo with faster shutter settings. 
While the main shot is low noise but blurry (\figref{teaser}(a)), the reference shot is sharp but noisy (\figref{teaser}(b)).
We learn ML models to align and fuse these two shots and output a clear photo without motion blur (Figure~\ref{fig:teaser}(c)). 
Our algorithm runs efficiently on Google Pixel 6, which takes 463 ms overhead per shot. 
Our experiments demonstrate the advantage and robustness of our system against alternative single-image, multi-frame, face-specific, and video deblurring algorithms as well as commercial products.
To the best of our knowledge, our work is the first mobile solution for face motion deblurring that works reliably and robustly over thousands of images in diverse motion and lighting conditions.
\end{abstract}

%
%


%
%


\begin{teaserfigure}
    \footnotesize
    \renewcommand{\tabcolsep}{1pt} 
	\newcommand{\figurewidth}{0.3\linewidth} 
	\newcommand{\image}{XXXX_20210929_203446_233} 
    \centering
    \begin{tabular}{ccc}
        \includegraphics[width=\figurewidth]{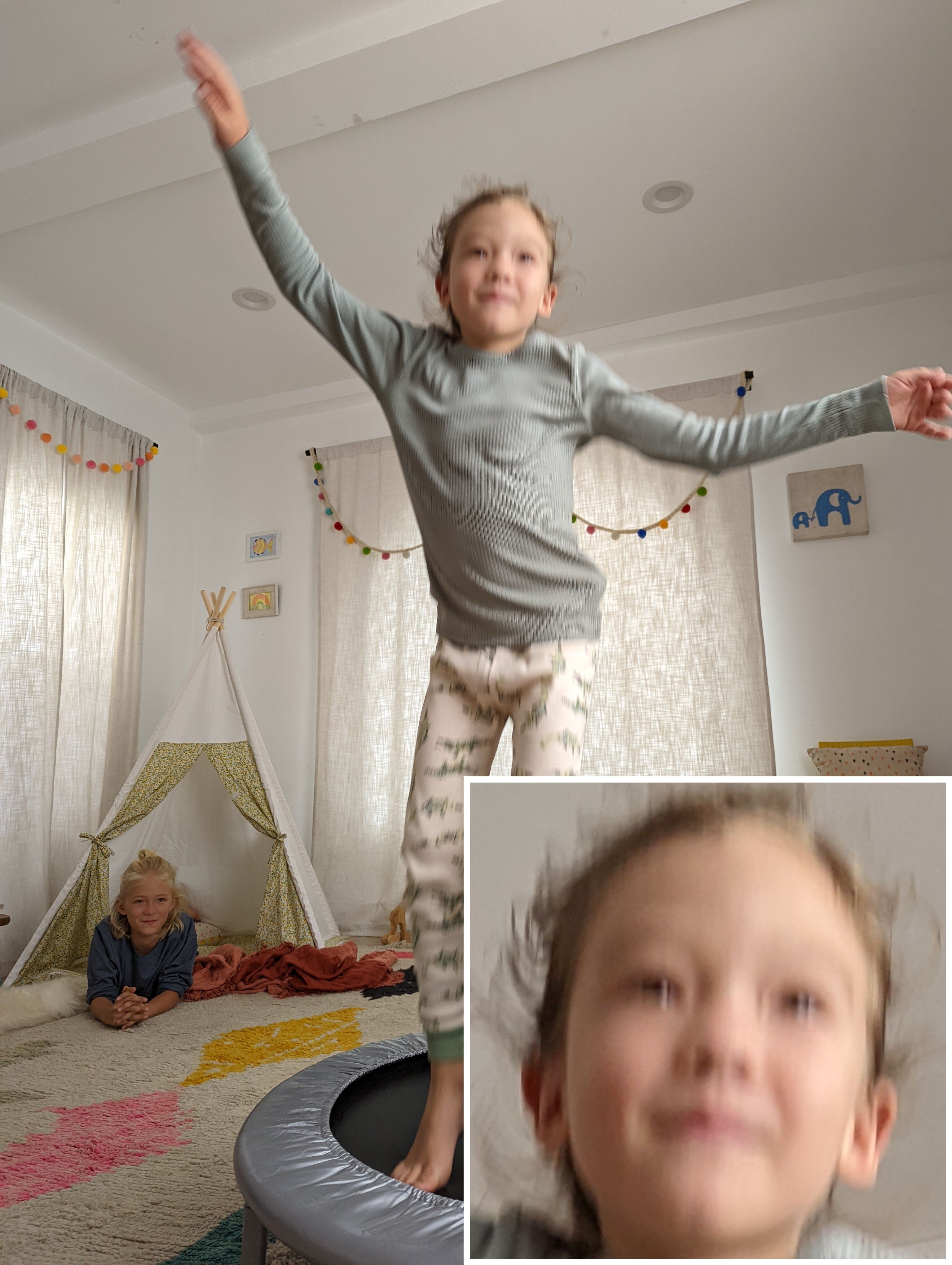} & 
        \includegraphics[width=\figurewidth]{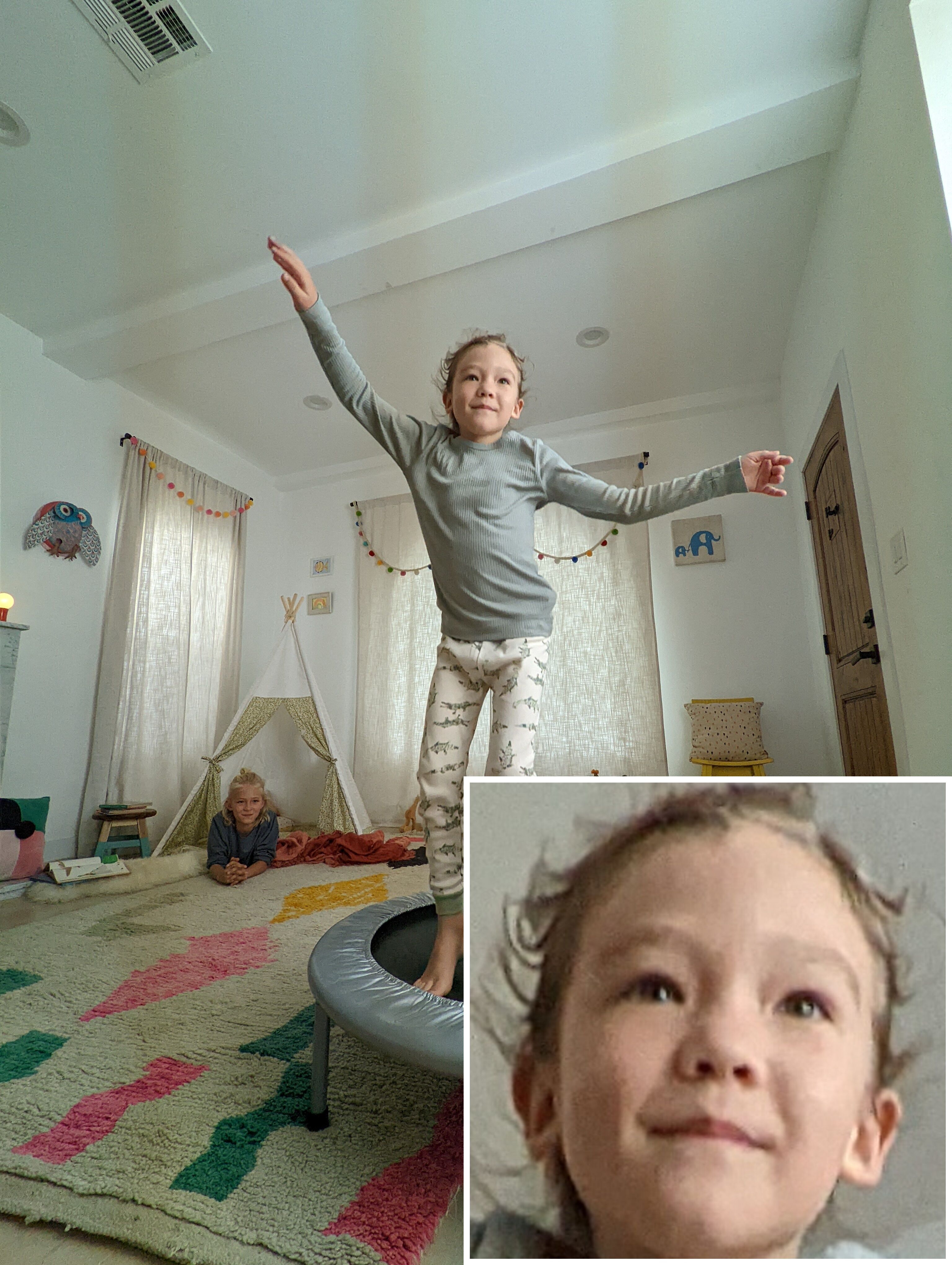} & 
        \includegraphics[width=\figurewidth]{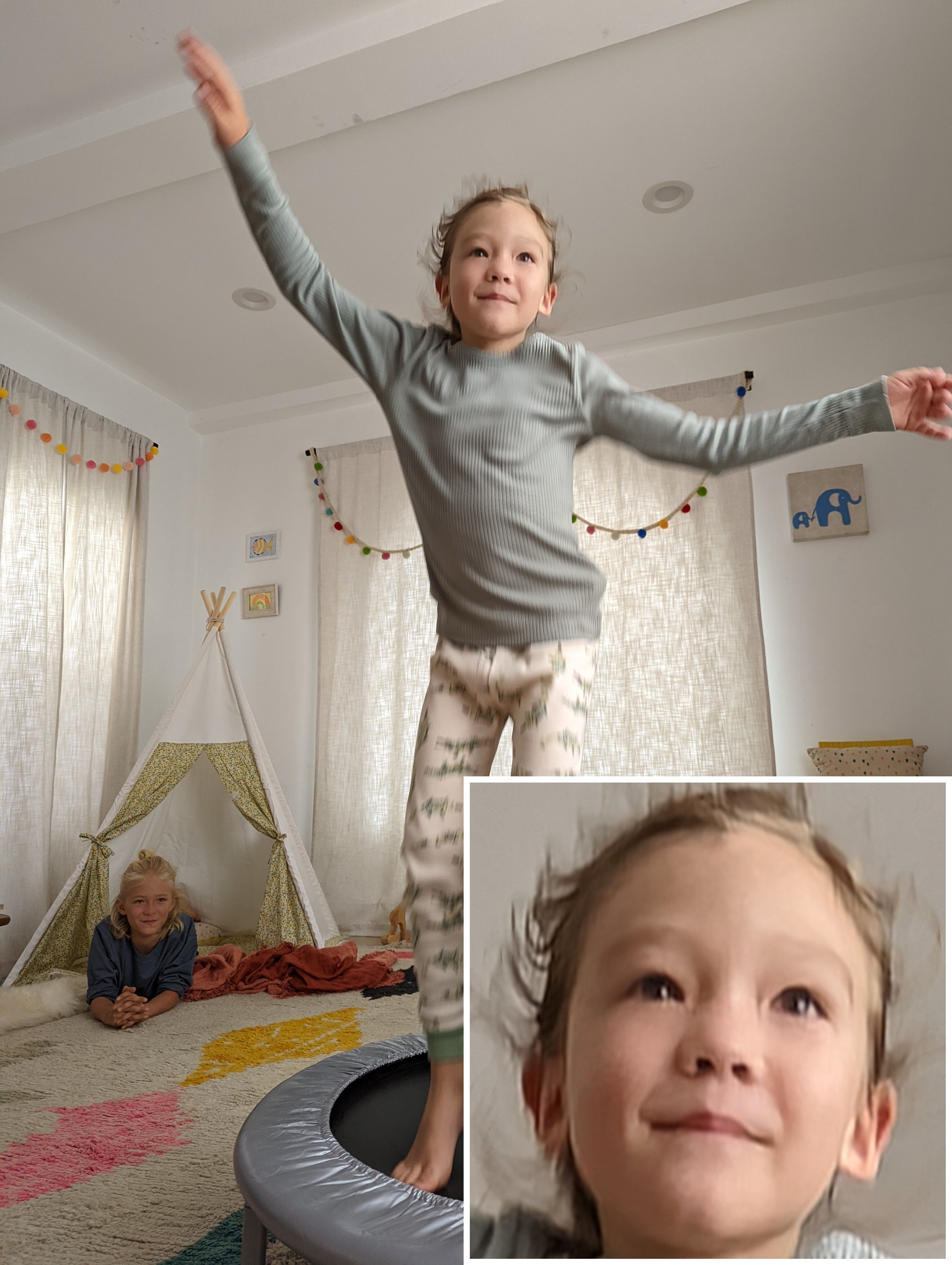} \\
        (a) Input: 26mm, 1/120s, ISO 101 &
        (b) Reference: 16mm, 1/480s, ISO 688 & 
        (c) Our deblurred result using (a) and (b).
    \end{tabular}
    \captionof{figure}{
        We present a robust and efficient system that leverages synchronized dual capture, which is commonly available on mobile phones, to deblur faces at an interactive rate on mobile devices. 
        \textbf{Input (a):} In this setting where the subject performs ordinary exercise (jumping on a trampoline), the commercial-grade auto-exposure system equipped with high-sensitivity sensor from a modern premium phone still produces objectionable motion blur on the face region. 
        \textbf{Reference (b):} Our system detects the subject motion and uses the ultrawide camera to capture a short-exposure shot as reference simultaneously. While the image appears noisy, has low-resolution, and has wrong color, it preserves the subject's facial details.
        \textbf{Output (c):} Our system recovers the sharp and authentic facial details by aligning and fusing the input with the reference. 
        }  
    \label{fig:teaser}
\end{teaserfigure}
\maketitle

\section{Introduction}
Taking photos of moving human subjects such as kids running or jumping has been a popular photography topic yet notorious challenge due to its well-known motion blur artifacts. While motion blur on limbs or body conveys the dynamic nature of human actions, it can distort or even totally remove the subject's facial identity or expression which are critical to the photo.
To capture crisp and clean facial details from the subject, sophisticated skills are required to predict types and strengths of motion, carefully estimate the lighting conditions, and set the perfect exposure time. 
Such skills take well-trained photographers years to master and would be otherwise difficult for casual users.

\begin{figure}
    \centering
    \footnotesize
    \renewcommand{\tabcolsep}{1pt} 
	\newcommand{\figurewidth}{0.49\linewidth} 
    \begin{tabular}{cc}
        \includegraphics[width=\figurewidth,trim={5cm 0 4cm 6.5cm},clip]{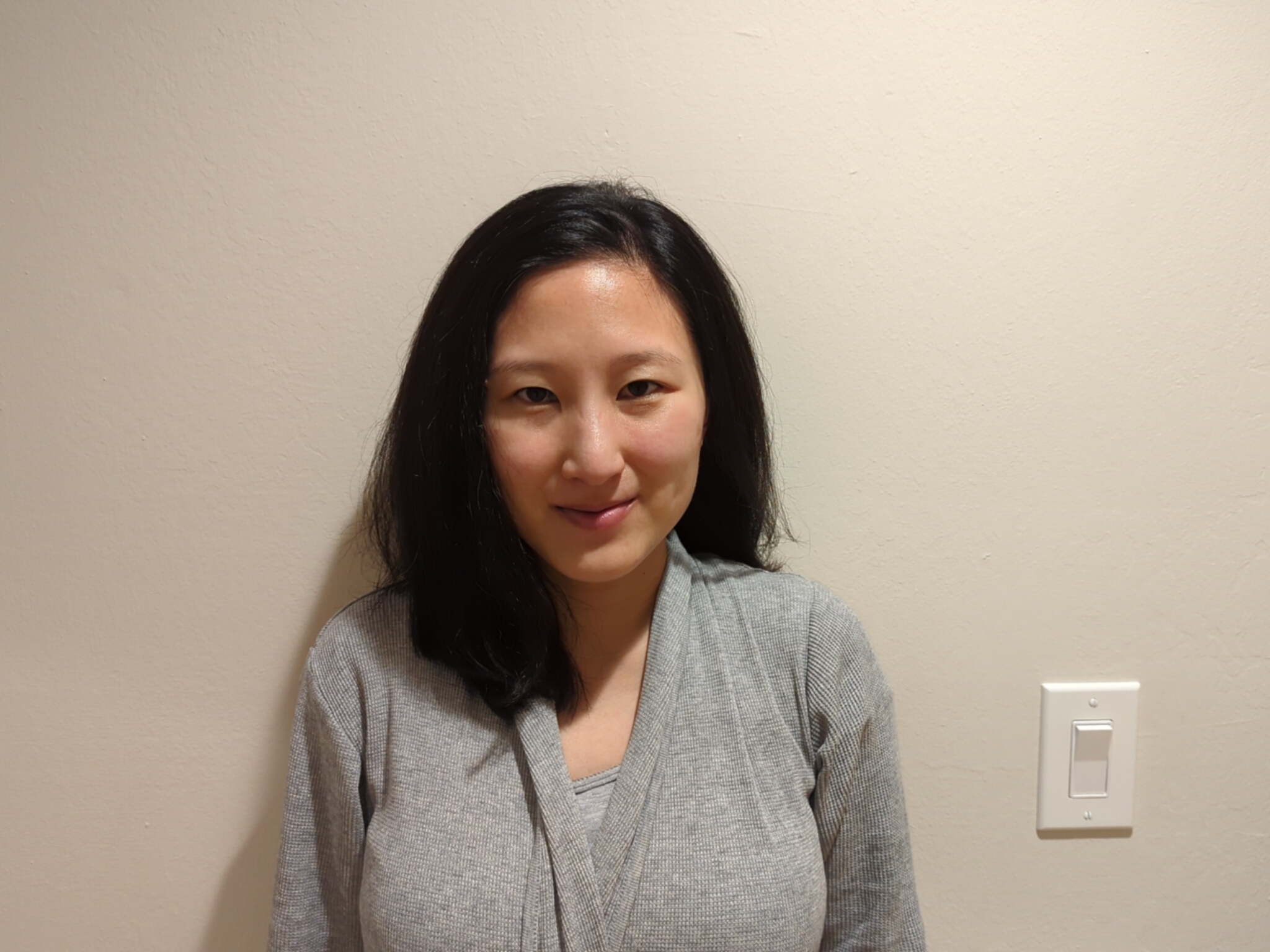} &
        \includegraphics[width=\figurewidth]{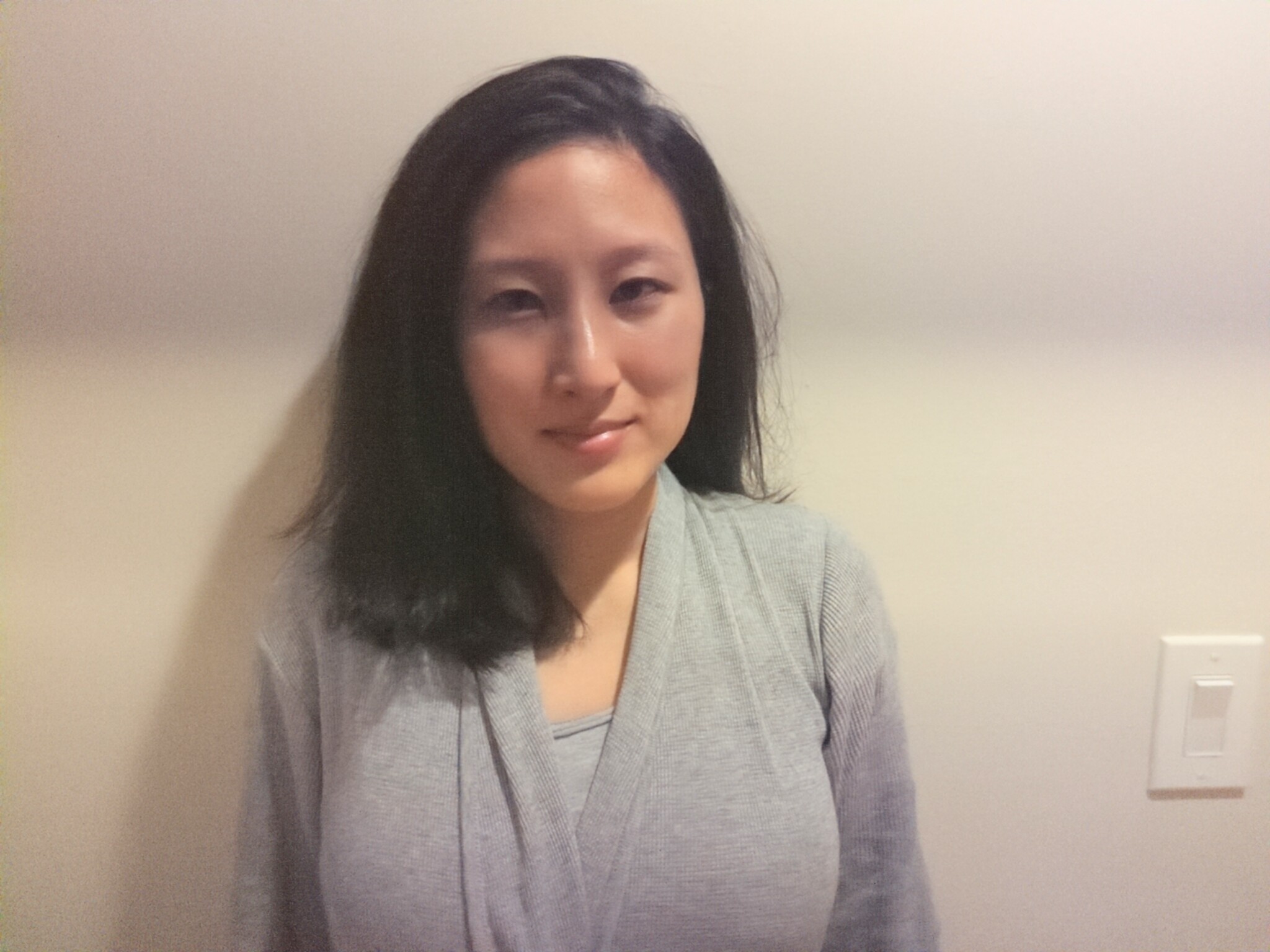} \\
        (a) 1/120s &
        (b) 1/1000s 
    \end{tabular}
    \vspace{-2mm}
    \caption{\textbf{Flickering and banding.}
    When a subject is illuminated by artificial lighting, \eg 60Hz, increasing shutter speed faster than 1/120s (a) not only leads to excessive noise but also objectionable banding artifacts as shown in (b) due to the rolling shutter effects in CMOS sensors on mobile phones.
    }
    \label{fig:flickering}
    \vspace{-3mm}
\end{figure}

Given the difficulty in precise motion estimation, premium mobile cameras today employ large sensors and aggressively reduce the exposure time when motion is detected~\cite{liba2019handheld}. 
Unfortunately, increasing the shutter speed introduces stronger noise~\cite{plotz2017benchmarking}, and brings objectionable flickering and banding artifacts under common AC-supplied lighting, due to the rolling shutter read of CMOS sensors~\cite{dietz2019shuttering}. 
For example, the 60Hz AC supply in the U.S. prevents the exposure time from being faster than 1/120s, as illustrated in~\figref{flickering}.

Recovering sharp details from blurry faces through post-capture processing is fundamentally challenging. Blind deconvolution methods often model motion blur as a global or slowly-varying local kernels~\cite{yang2019variational,ren2016image}, and cannot handle face motions that may contain non-rigid and out-of-plane movements, saturation at eye highlights, camera ISP non-linearity, lens aberration, and self-occlusion between hairs and facial landmarks.
Deep learning methods using a single image~\cite{nah2017deep,cho2021rethinking}, a multi-frame burst~\cite{aittala2018burst}, or a video sequence~\cite{su2017deep, son2021recurrent} have made significant progresses over the years. However,  high-quality solutions on high-resolution data require expensive computation that is prohibitive on mobile phones with limited memory and processing budget.

This paper describes a system to deblur faces of moving subjects captured by handheld mobile phones. We tailor the system to faces given its importance in mobile photography. 
Our system, which launched to the public as Face Unblur on Google Pixel 6 smartphones, aims to achieve the following goals: 

\begin{itemize}[-]

\item Reliably produce high-quality, high-resolution (2.3MP), realistic-looking, and artifact-free deblur results on moving subjects in the wild. We handle any form of face motion in real-world, where the motion blur size can reach hundreds of pixels.

\item Require no additional users inputs other than pressing the shutter button. The system detects motion blur, identifies the blurry areas, and applies deblurring automatically.

\item Run on mobile phone at interactive rate. In other words, users see the result immediately after the photo is taken.

\item Operate within the power, memory, and computation budget provided by the SoC on Google Pixel 6. 

\item Support zero-shutter lag capture for motion subject. The deblurring is applied to the moment that the user intends to capture, not nearby frames.
\end{itemize}

Our solution is built on (1) the heterogeneous dual camera system adopted by the majority of phone vendors, \ie a wide-angle main camera and a secondary ultrawide-angle camera, denoted as \wide~ and \ultrawide~ throughout this paper, and (2) the ability to synchronize timestamp, auto-exposure, and auto white-balancing between \wide~ and \ultrawide~ cameras offered by most smartphones. 
When the user captures a moving subject using the main \wide~ camera, our system captures a synchronized reference raw using the \ultrawide~ camera at a faster shutter speed to acquire sharp but noisy facial details. The main \wide~ and reference \ultrawide~ shots exhibit different viewpoint and sensor characteristics, \eg noise, gain, color, etc. We train deep CNNs to align and fuse the low noise but blurry linear RAW from \wide, and noisy but sharp linear RAW from \ultrawide, generating a clean and sharp linear raw of the face region from the \wide~'s point of view (Section~\ref{sec:fusion}). The fused linear RAW is then fed to a standard image processing pipeline to output the tone-mapped result. Compared to the alternative that takes long and short exposure frames from the same camera, our dual camera approach leverages temporal synchronization to greatly improve the quality of image alignment between \wide~ and \ultrawide. 

However, dual \wide~ and \ultrawide~ streaming consumes prohibitively high power and memory footprint. To reduce the amortized power, we propose an adaptive stream system to dynamically enable and disable the \ultrawide~ sensor based on real-time motion analysis. We integrate the algorithm into the camera driver to avoid excessive activation delay (Section~\ref{sec:adaptive}) \nothing{(12 frames if the algorithm were in Application layer)}. Finally, we describe the requirements of the camera image signal processing pipeline to enable our system (Section ~\ref{sec:isp}).

Section~\ref{sec:results} evaluates our deblur results over 1783 images exhibiting motions in daily tasks, such as jumping, walking, and exercising, containing subjects of diverse skin color, gender, and face poses.
The inputs are challenging and consist of local, non-rigid, and large motion, where blur kernel may reach up to hundreds of pixels.
We compare our results against state-of-the-art methods in single-image, multi-frame, face-specific, and video-based deblurring papers, as well as commercial products such as iPhone13 Pro, Shake Reduction filter in Adobe PhotoShop, and Samsung Gallery's Remaster function. 
Our method achieves higher scores on no-reference image quality metrics and generates results that look visually more pleasing.
Our images used for comparisons in Section 6 are available at the project website\footnote{\website}.
To the best of our knowledge, our system is the first practical face deblur solution that runs on mobile phones.

\section{Related Work}

\begin{figure*}[t!]
    \centering
    \includegraphics[width=1.0\textwidth]{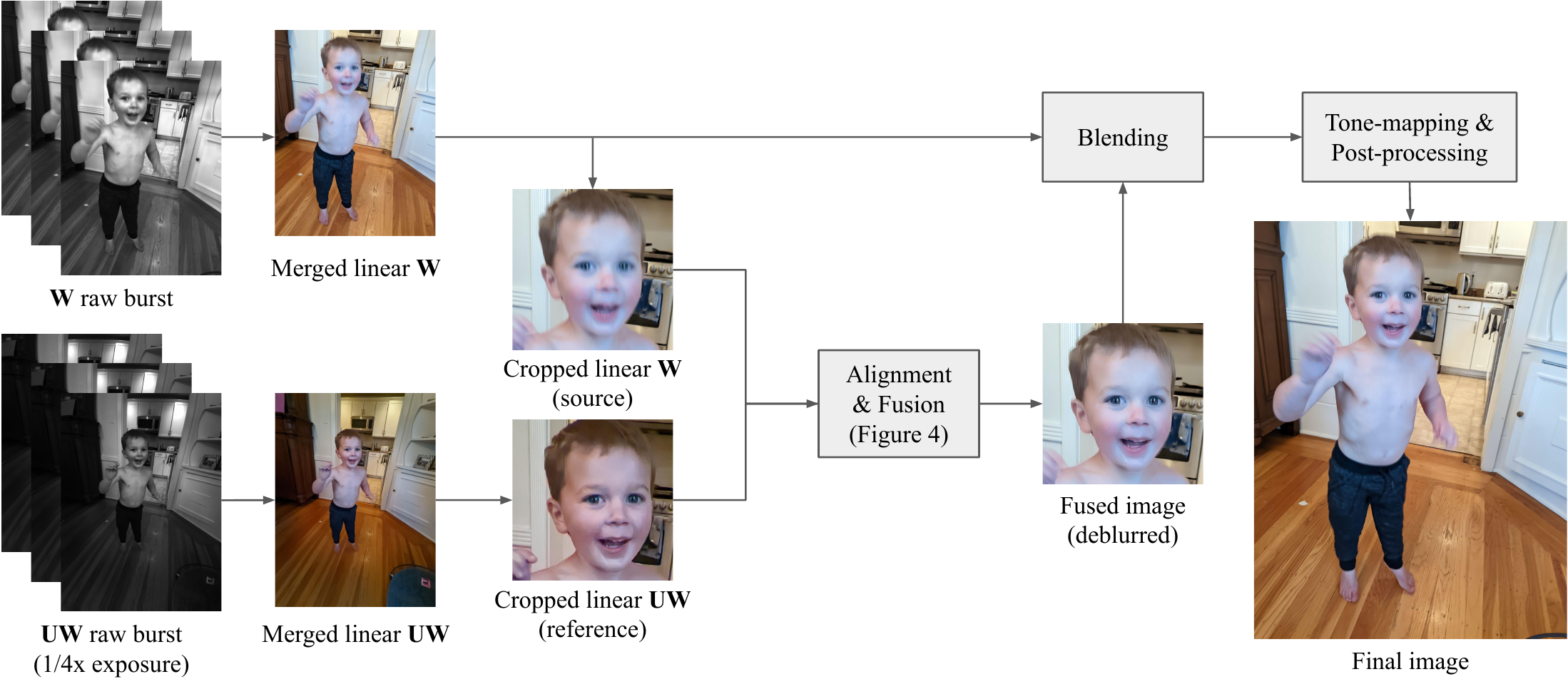}
    \vspace{-5mm}
    \caption{
        \textbf{System overview.} 
        Our system takes raw bursts from \wide~ and \ultrawide. We control the shutter speed of \ultrawide~ to be 4$\times$ faster than \wide. \ultrawide~ image is sharper but appears noisier and may contain flickering, \ie the vertical bands in the image, due to rolling shutter under artificial lighting. We merge the burst raw into a linear RAW, and crop the face region for deblurring using ML-based alignment (\secref{alignment}) and fusion (\secref{fusion-fusion}). Detailed steps are described in ~\figref{fusion_algorithm}. We then blend the deblurred face back to \wide~ (\secref{blending}), and outputs the final image through tone mapping and post-processing (\secref{post-processing}).
        Note that the reference image size is $2\times$ smaller than the source. We enlarge the reference image for better visualization.
    }
    \label{fig:overview}
\end{figure*}

Our system addresses a long-standing computer vision problem: motion deblurring. 
The full review of this highly active field is outside the scope of this paper, and we refer users to an overview paper~\cite{koh2021single}.
The state-of-the-art methods using variants of UNet ~\cite{cho2021rethinking} require a large and realistic dataset for training ~\cite{nah2017deep, rim2020real}. However, collecting large-scale real-world blur images paired with ground truth is still impractical. In contrast, we show that our system based on dual camera fusion only requires training on synthetic blur images to generate high-quality results. We cover the most relevant papers related to our work in this section.

\subsection{Single-image motion deblurring}
Conventional methods typically rely on the MAP optimization framework~\cite{fergus2006removing,shan2008high, cho2009fast, xu2010two} and natural image priors~\cite{krishnan2011blind, xu2013unnatural, sun2013edge, lai2015blur, pan2016blind} for single-image blind deblurring.
However, using complicated image priors requires iterative optimization steps, which lead to high computational cost, and thus makes it difficult to be deployed to mobile devices with limited computing resources. 

With the advance of deep learning techniques in recent years, several learning-based approaches adopt deep CNNs for motion deblurring.
Early methods train CNNs to predict local blur kernels~\cite{sun2015learning, gong2017motion} or learn image priors~\cite{li2018learning}, but still involve conventional optimization-based framework to deblur images.
Recent approaches train end-to-end CNNs for single-image debluring by exploring different architecture designs~\cite{nah2017deep, tao2018scale, gao2019dynamic, li2020dynamic, zhang2019deep, suin2020spatially, zhang2020deblurring, zamir2021multi, cho2021rethinking, kaufman2020deblurring} or GAN-based training~\cite{kupyn2018deblurgan, kupyn2019deblurgan}.
While state-of-the-art methods achieve great performance on benchmark datasets (e.g., GoPro~\cite{nah2017deep} and RealBlur~\cite{rim2020real}), single-image deblurring methods typically are not robust enough to be generalized well to real images in the wild, e.g., either fail to remove large motion blur or leave unpleasing visual artifacts (see~\figref{compare_academic}).
In contrast, we develop an effective reference-based deblurring approach that achieves high-quality results on mobile devices, which enables processing 2.3MP images, while most single-image deblur methods are limited to 720p~\cite{nah2017deep,hu2021pyramid} due to computational constraints.
Furthermore, unlike GAN-based approaches~\cite{xu2017learning, kupyn2019deblurgan} that have risks of generating a different face, our system takes a reference image from the same subject captured at the same time, which can preserve the subject identity well.

On the other hand, when deblurring low-light images, saturated light streaks are commonly visible on eyeglasses, eyeballs, or reflections.
Optimization-based approaches handle saturation by excluding saturated pixels~\cite{whyte2014deblurring, pan2016robust, dong2017blind} or estimating blur kernels from light streaks~\cite{hu2014deblurring}.
We found that our model trained on synthetic data may not be able to remove saturated light streaks and leave undesired ringing artifacts (see~\figref{highlight}).
Therefore, we add synthetic light streaks to our training data and find this simple data augmentation sufficient to remove saturated blur streaks.

\subsection{Multi-frame and video deblurring}
As single-image deblurring is a challenging ill-posed problem, several methods have exploited additional input images for deblurring.
Multi-frame deblurring approaches merge a burst of images via Fourier burst accumulation~\cite{delbracio2015burst} or deep CNNs~\cite{aittala2018burst}.
Early video deblurring methods use optical flow to model linear motion blur~\cite{hyun2015generalized, ren2017video} or merge sharp pixels or patches from neighboring frames~\cite{cho2012video, delbracio2015hand}.
Recent learning-based approaches learn to aggregate information from neighboring frames using optical flow~\cite{su2017deep}, recurrent networks~\cite{hyun2017online, nah2019recurrent, son2021recurrent, zhou2019spatio}, spatial-temporal transformers~\cite{kim2018spatio}, deformable convolutions~\cite{wang2019edvr}, and multi-patch architectures~\cite{deng2021multi}.
Our system captures \wide~ and \ultrawide~ raw bursts consisting of $7-9$ frames and uses the robust local alignment in~\cite{hasinoff2016burst} to merge raw frames to a single linear image before deblurring.
The robust local alignment step avoids additional blur in the raw-merging step.
Compared to deblurring on raw burst frames, deblurring on the merged linear image avoids the challenging temporal alignment and excessive memory cost.

\subsection{Reference-based deblurring}
Reference-based approaches take a pair of blurry and sharp, short-exposed images as inputs~\cite{yuan2007image,chang2021low, gu2020blur, zhuo2010robust}.
Early approaches adopt conventional Richardson–Lucy deconvolution~\cite{yuan2007image} or MAP framework~\cite{zhuo2010robust} to estimate a blur kernel for deblurring.
These methods assume the input and reference images are well-aligned, and motion blur is spatially-invariant \ie a global blur kernel exists. The assumption does not apply to dynamic subject motions in our problem.
Gu~\etal~\shortcite{gu2020blur} use optical flow to align the noisy reference image and build GMM priors to model the sharp latent image.
This work requires expensive iterative optimization (10+ minutes as reported in~\cite{gu2020blur}), which makes it difficult to run efficiently on mobile devices.
Chang~\etal~\shortcite{chang2021low} learns CNNs to align and fuse long and short exposure frames for deblurring. They capture frames sequentially, which makes alignment challenging for fast-moving subjects that often involve non-rigid motion.
On the contrary, our system captures a pair of blurry-noisy inputs simultaneously.
Our alignment only needs to handle the parallax from the camera baseline, which is typically a rigid transform and is easier to produce satisfying alignment results.

\subsection{Face Deblurring}
Compared to natural images, face images are highly structural and consist of rich semantic information such as facial contours and landmarks.
Conventional optimization approaches~\cite{hacohen2013deblurring, pan2014deblurring} search for a reference face to guide the image deblurring.
These methods often require careful alignment between the blurry and reference images to estimate accurate blur kernels.
Recent CNN-based approaches learn semantic labels~\cite{shen2018deep, shen2020exploiting, yasarla2020deblurring}, facial sketches~\cite{lin2020learning}, or 3D facial reconstructions~\cite{ren2019face} for deblurring. 
However, such methods based on facial priors could fail when a subject wears a mask (e.g., the third example in~\figref{compare_academic}).
Instead of learning face priors or searching for a reference face from an external database, we propose an end-to-end system that concurrently captures a sharp reference face and learns ML models to fuse the blurry and reference images.

\section{Fusion Deblurring}
\label{sec:fusion}

\begin{figure}[t!]
    \centering
    \includegraphics[width=1.0\linewidth]{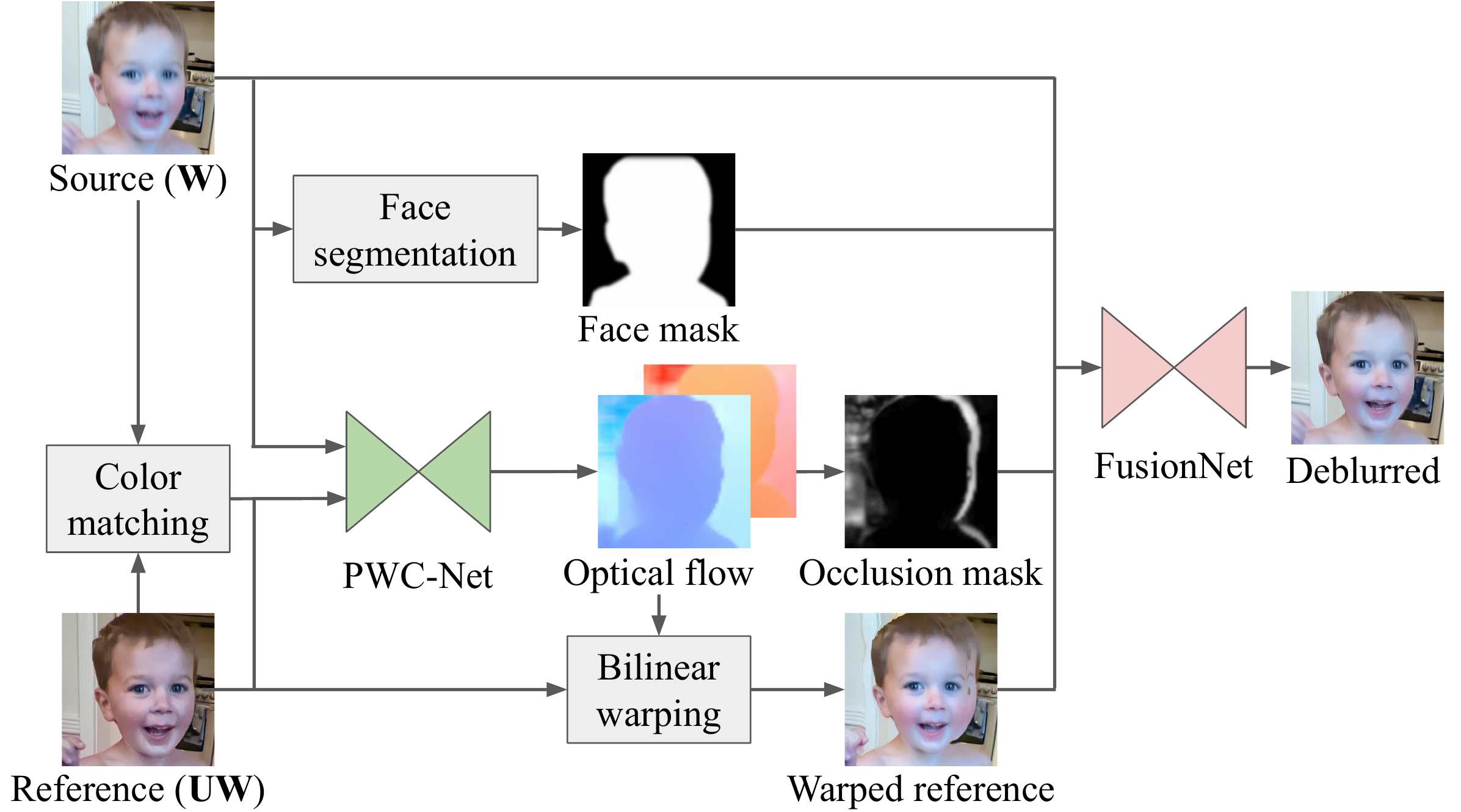}
    \caption{
        \textbf{Alignment and fusion}. Given crops of blurry faces on \wide, and sharp yet noisy faces on \ultrawide, we produce the sharp and clean face by matching the color of \ultrawide~ to \wide, densely aligning the images using PWC-Net optical flow, and feed the \wide~ and aligned \ultrawide~ to a UNet-based FusionNet. We use face mask and occlusion mask to reduce the warping artifacts in the fused outputs. 
        Note that the reference image is $2\times$ smaller than the source. We enlarge the reference image for better visualization.
    }
    \label{fig:fusion_algorithm}
\end{figure}

We consider the image acquisition system consisting of a main wide camera \wide, and a reference fixed-focus ultrawide camera \ultrawide. The field-of-view (FOV) are 80\degree and 107\degree, respectively, throughout the results in the paper. \wide~ and \ultrawide~ are roughly fronto-parallel such that \ultrawide's FOV covers \wide's. The sensor resolutions of both \wide~ and \ultrawide~ are 12MP. \ultrawide~ is noisier than \wide, since modern phones often optimize the sensitivity for the main \wide~ camera. Note that the dual \wide+\ultrawide~setup is common for all mid-tier and premium smartphones.

Our fusion algorithm takes two images as input, where we treat \wide~ as the source and \ultrawide~ as the reference image.
When subject motion is detected, we enable dual camera streaming to concurrently capture \wide~ and \ultrawide~ raw bursts ~\cite{hasinoff2016burst}.
The exposure time of \wide~ is determined by the auto-exposure algorithm for the optimal SNR and sharpness performance in averaged cases, but is usually too long for fast-moving subjects, and leads to blurry faces.
The \ultrawide~ image is deliberately captured at a higher shutter speed than \wide~ to minimize motion blur, at the cost of more noise and artifacts. We lock the \ultrawide's exposure time to be $\frac{1}{N}$ of \wide's exposure time. We test $N=2$ and $N=4$ in this paper.
Note that our system design is independent of the choice of raw burst capture. We adopt the raw capture in our pipeline since many phones employ raw burst capture in the post-capture processing pipeline for high-dynamic-range enhancements.

\figref{overview} shows an overview of our dual camera system pipeline.
First, we merge the raw bursts into linear raw images~\cite{hasinoff2016burst}.
Then, we apply face detection and subject segmentation to determine a face mask and ROI crop that will be deblurred.
The cropped \wide~ and \ultrawide~ images are then fused through our fusion ML models to generate the result in the linear raw format, as illustrated in ~\figref{fusion_algorithm}.
Finally, we blend the cropped fusion result with the full-resolution linear raw image and apply post-processing, which includes sharpness enhancement, global, and local tone-mapping, to output a final 8-bit RGB image. 
In this section, we will discuss the key components of the fusion algorithm.

\begin{figure}[t!]
    \centering
    \footnotesize
    \renewcommand{\tabcolsep}{1pt} 
	\newcommand{\figurewidth}{0.49\linewidth} 
    \begin{tabular}{cc}
        \includegraphics[width=\figurewidth]{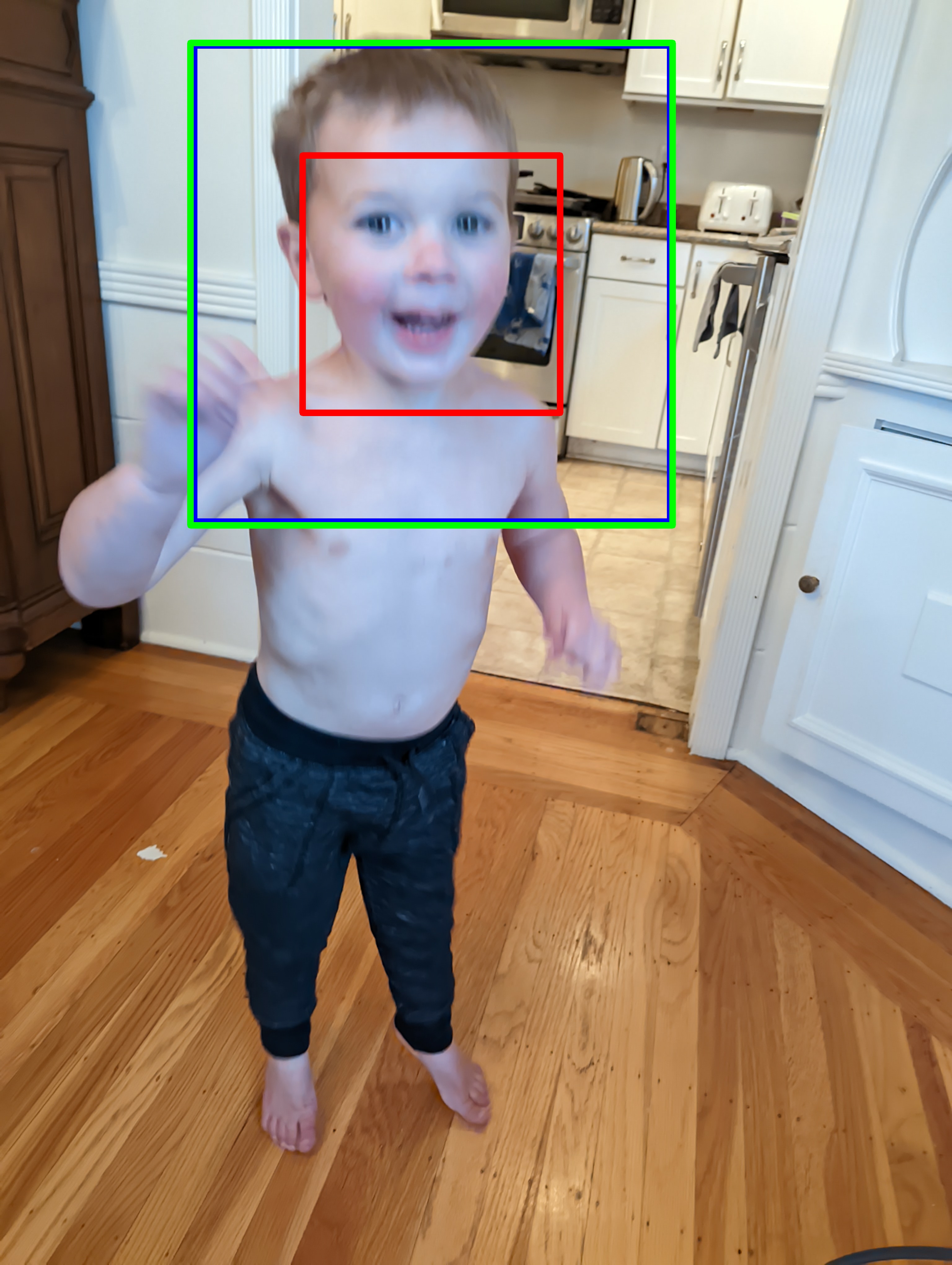} &
        \includegraphics[width=\figurewidth]{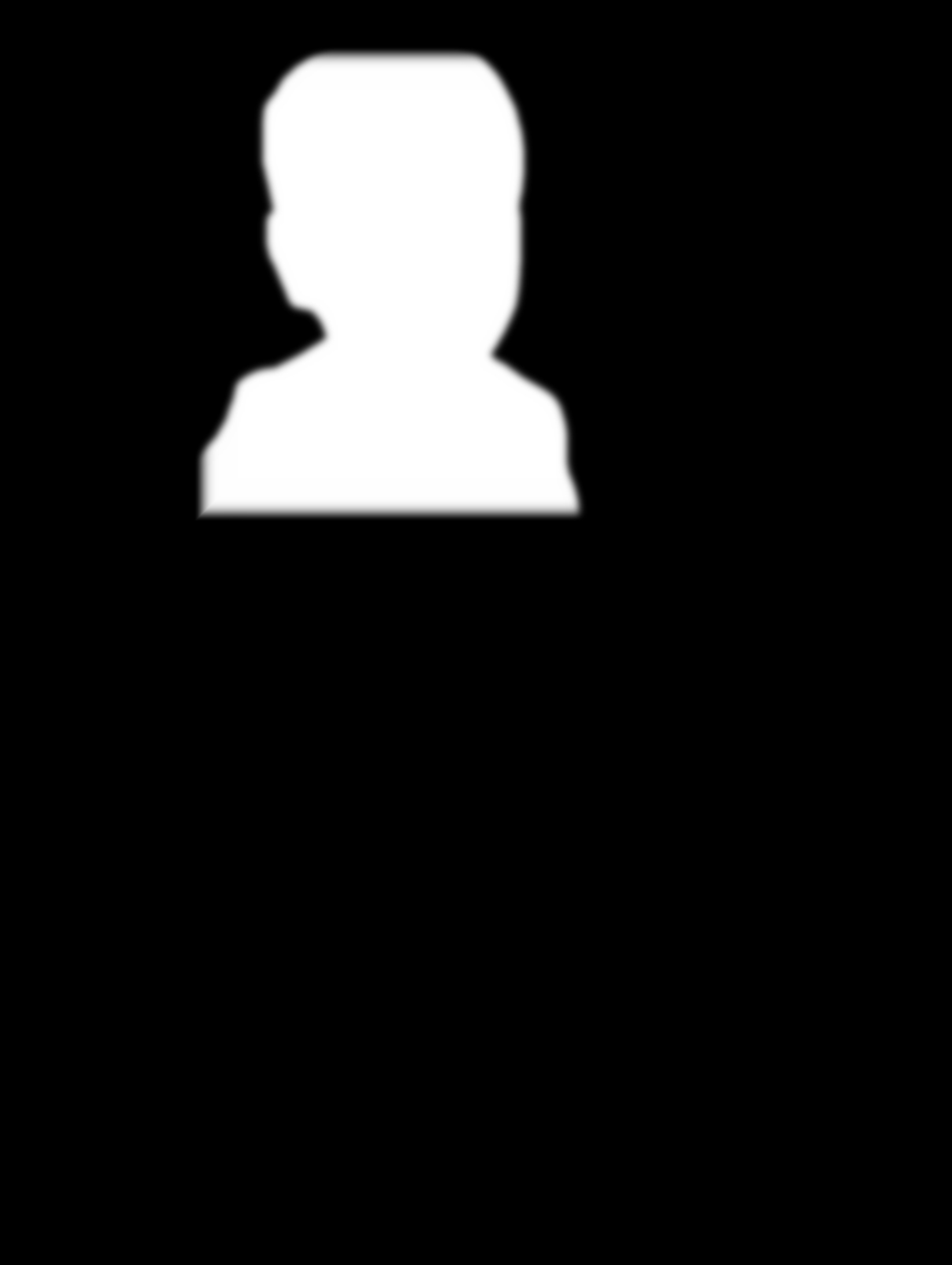}
        \\
        (a) Fusion ROI &
        (b) Face mask
    \end{tabular}
    \caption{\textbf{Fusion ROI and face mask.}
        (a) We first detect the largest face box (red box) in the source image, extend it by $1.75\times$ to cover the entire face (blue box), and round up to $1536\times1536$ for efficient inference (green box).
        Fusion ROI is the minimum box covering the extended face box while staying within the source image domain.
        (b) After we determine the fusion ROI, we apply subject segmentation to obtain a face mask and smooth the boundaries to avoid sharp transition after blending. 
    }
    \label{fig:fusion_roi}
\end{figure}

\subsection{Pre-processing}
\label{sec:preprocessing}

\paragraph{Fusion ROI and face mask.}
As we focus on faces, we restrict the processing area to a rectangular region-of-interest (ROI) around the largest face. 
We determine an ROI for fusion by detecting the largest face box from the source image (red box in~\figref{fusion_roi}(a)), extends it by a margin to cover face, hairs, ears, and chin (blue box in~\figref{fusion_roi}(a)), and round up to a maximum $1536\times1536$ for efficient ML inference (green box in~\figref{fusion_roi}). 
We generate a face mask through a UNet-based ML subject segmenter~\cite{wadhwa2018synthetic} over the fusion ROI to identify the facial pixels (\figref{fusion_roi}).
Finally, we smooth the mask boundaries to avoid sharp transition after blending.

\paragraph{Color matching.}
While we synchronize the total exposure, \ie exposure time $\times$ total gain, and auto white-balance between \wide~ and \ultrawide~ cameras, the source and reference images can still show color mismatches given different sensor characteristics and calibration tolerance.
The color mismatch leads to alignment error and undesired color shift in the fusion output.
We match the color of reference $\refimage$ to source by matching the image statistics ~\cite{reinhard2001color} in linear space.
We first normalize the color space using the following:
\begin{equation}
    \refimage^{n} = \srcccm^{-1} \cdot \refccm \cdot \refimage,
    \label{eq:ccm_normalization}
\end{equation}
where $CCM$ is the $3 \times 3$ color-conversion matrix obtained from camera calibration in image metadata.
As we operate in the linear image space, RGB color is calibrated to the same color space after applying $CCM$.
Eq.~\eqnref{ccm_normalization} normalizes the color space of reference image to the source image such that $\srcccm \cdot \refimage^{n}$ would look visually similar to $\srcccm \cdot \srcimage$.
Then, we match the global mean of source and reference images by:
\begin{equation}
    \colormatchedrefimage = \frac{\refimage^n}{\mu_{\text{ref}}} \cdot \mu_{\text{src}},
    \label{eq:color_matching}
\end{equation}
where $\mu_{\text{src}}$ and $\mu_{\text{ref}}$ are the global color mean of the source image $\srcimage$ and normalized reference image $\refimage^n$, respectively.
We apply ~\eqnref{color_matching} to each RGB channel separately. 
We choose division instead of subtraction in~\eqnref{color_matching} since the color mismatch between source and reference comes from different sensor sensitivities, which can be approximated through the intensity ratio. 
Subtracting the RGB means leads to negative intensity values, and clamping-to-zero loses pixel information and de-saturates images.

\begin{figure}
    \centering
    \footnotesize
    \renewcommand{\tabcolsep}{1pt} 
	\renewcommand{\imagewidth}{0.49\linewidth} 
    \newcommand{\addimage}[1]{
        \begin{tabular}{cc}
        \includegraphics[width=\imagewidth]{figures/ablation/pwcnet/#1/source.jpg}  &
        \includegraphics[width=\imagewidth]{figures/ablation/pwcnet/#1/reference.jpg} \\
        (a) Source & (b) Reference \\
        \includegraphics[width=\imagewidth]{figures/ablation/pwcnet/#1/warped_reference_and_flow_pwcnet_full.jpg} &
        \includegraphics[width=\imagewidth]{figures/ablation/pwcnet/#1/warped_reference_and_flow_pwcnet_small.jpg} \\
        (c) Warped-reference and flow from & 
        (d) Warped-reference and flow from  \\
        PWC-Net (full-size input) & 
        PWC-Net ($\frac{1}{4}\times$-size input) \\
        \end{tabular}
	}
    \addimage{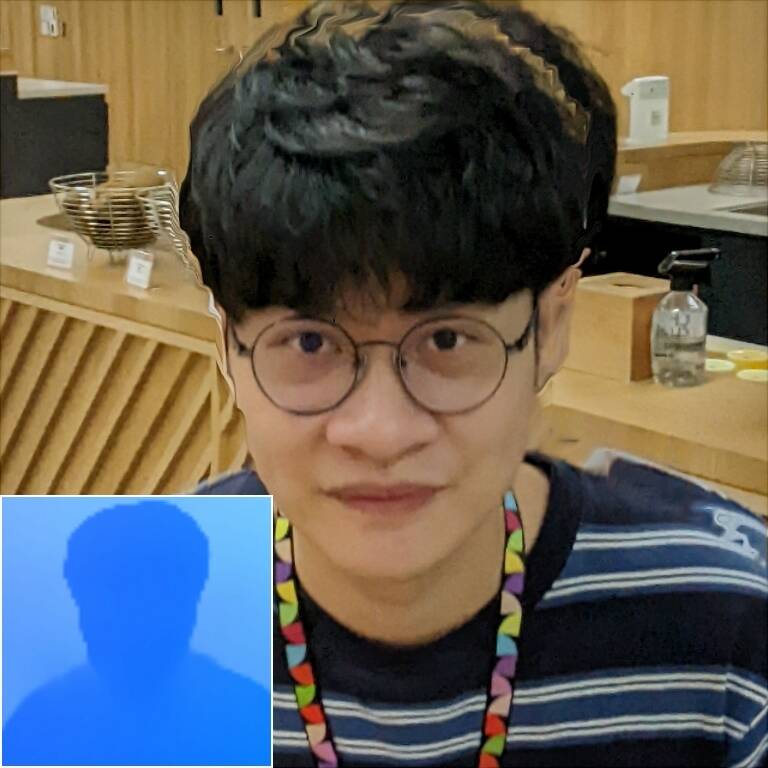}
    \caption{\textbf{Alignment quality.}
        As the shifts between (a) source and (b) reference images are larger than the motion magnitude of the optical flow training data, directly applying the PWC-Net to the full-size input leads to inaccurate flow estimation and severe distortion on the warped reference image in (c). 
        To account for the discrepancy, we downsample the source and reference images by $\frac{1}{4}\times$ for the inputs to the PWC-Net, obtaining more accurate flow to align the face in (d).
    }
    \label{fig:flow_magnitude}
\end{figure}

\subsection{Image Alignment}
\label{sec:alignment}
As the \wide~ and \ultrawide~ cameras do not have the same optical center, the parallax introduces misalignment and occlusion between the source and reference images.
To align the reference to source for image fusion, we use PWC-Net~\cite{sun2018pwc} to estimate optical flow, since it shows similar performance as recent state-of-the-art methods, \eg RAFT~\cite{teed2020raft}, while running faster with less memory consumption~\cite{sun2021autoflow}.
As the source image is blurry and the reference image is typically noisy, directly applying PWC-Net to full-resolution images may lead to inaccurate flow estimation. 
Furthermore, optical flow networks are typically trained on image sizes at 720p, \ie $1280 \times 720$, or smaller\footnote{{$576 \times 448$} in Autoflow~\cite{sun2021autoflow}, {$960 \times 540$} in FlyingThings3D~\cite{mayer2016large}, and {$1024 \times 436$} in Sintel~\cite{butler2012naturalistic}.}, where the maximal motion magnitude is typically less than 100 pixels (c.f.,~\cite{sun2021autoflow} Figure 8).
Our full-resolution image size is $4080 \times 3072$, where the average motion magnitude between the source and reference images is larger than 300 pixels, and cropping to $1536 \times 1536$ does not change the motion magnitude that we need to handle.
Directly aligning the source and reference images lead to severe distortion as shown in~\figref{flow_magnitude}.
To account for this, we downsample both the source and reference images by $4\times$ and upsample the estimated flow to warp the reference image:
\begin{equation}
    \forwardflow = \pwcnet( (\srcimage)_{\downarrow}, (\colormatchedrefimage)_{\downarrow} )_{\uparrow},
\end{equation}
where $(\cdot)_{\downarrow}$ and $(\cdot)_{\uparrow}$ represent the bilinear downsampling and upsampling operator, respectively.
The warped reference image is generated via bilinearly resampling $\warp$:
\begin{equation}
    \warpedrefimage = \warp(\colormatchedrefimage; \forwardflow).
\end{equation}
\figref{flow_magnitude} shows that the flow predicted from $\frac{1}{4}\times$ resolution images is much cleaner, resulting in better alignment quality.

\paragraph{Optimizing PWC-Net for mobile devices.}
The original PWC-Net~\cite{sun2018pwc} uses 7-level pyramids and a DenseNet~\cite{huang2017densely} at the prediction level.
The on-device inference takes 113 ms on $384 \times 512$ input images, which is too slow for our use cases.
We optimize the architecture of PWC-Net by (1) reducing the number of pyramid levels from 7 to 5, (2) removing the DenseNet structure, (3) limiting the search range of cost volume from 4 to 2, and (4) replacing each $3\times3$ conv layer with a $3\times3$ depth-wise conv layer followed by an $1\times1$ conv layer~\cite{howard2017mobilenets}.
We re-train the optimized PWC-Net on the AutoFlow dataset~\cite{sun2021autoflow}.
Our optimization reduces the inference latency from 113 ms to 13 ms, the memory usage from 600 MB to 34 MB, and the model size from 40 MB to 1.27 MB, which are suitable for mobile platform.
While the average EPE on the Sintel training final set increases from 2.91 to 3.73 after our optimization, we find that the warping quality on faces remains on par with the original PWC-Net. 
We also quantize the model weights from fp32 to fp16 using post-training quantization, which saves another $50\%$ of model binary size without affecting the flow quality. 

\paragraph{Occlusion mask.}
We compute the occlusion mask using the forward-backward flow consistency check. 
We first generate the forward $\forwardflow$ and the backward flow $\backwardflow$ using PWC-Net. 
Then, we project the pixel coordinates from the source image domain to the reference using the forward flow and re-project them back to the source image domain using the backward flow. 
The occlusion mask $\occlusionmask$ is defined as the re-projected distance multiplied by a scaling factor $s$ as the following:
\begin{equation}
    \occlusionmask(\textbf{x}) = \min( s \cdot || \warp(\warp(\textbf{x}; \forwardflow); \backwardflow) - \textbf{x} ||_2, 1.0),
    \label{eq:occlusion}
\end{equation}
where $\warp$ is the warping operator, $\textbf{x}$ is the 2D image coordinate on the source, and $s$ controls the strength of the occlusion mask.

\begin{figure}[t!]
    \centering
    \includegraphics[width=1.0\linewidth]{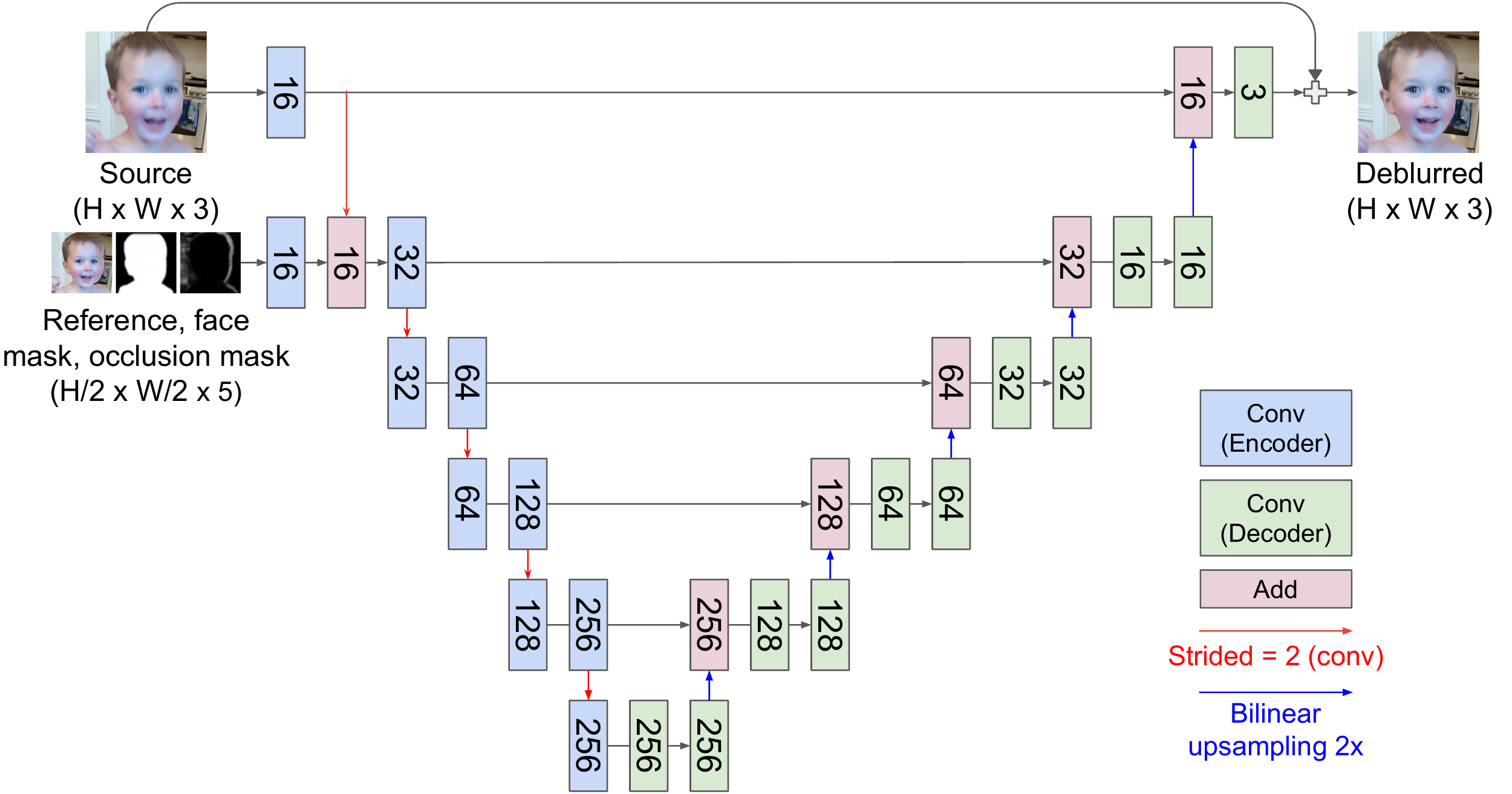}
    \caption{
        \textbf{FusionNet architecture.} The model is a variant of residual UNet for multi-scale processing. The model takes inputs from the blurry source and stacked tensors consisting of the reference image, face mask, and occlusion mask at half the size of the source image. 
    }
    \label{fig:fusion_net}
\end{figure}

\subsection{Image Fusion}
\label{sec:fusion-fusion}
We learn a UNet variant network, called FusionNet, for reference-based deblurring.
Our FusionNet takes inputs from the blurry source image $\srcimage$, warped reference image $\warpedrefimage$, face mask $\facemask$, and occlusion mask $\occlusionmask$.
\nothing{We describe the dimension of each input image.}
Since the focal length of \ultrawide~ camera is nearly half of the \wide~ in our camera system, the size of cropped face on the reference image is roughly half of the source image. 
The face mask is generated from a thumbnail resolution ($640\times480$). The optical flow and occlusion mask are computed at the $\frac{1}{4}\times$ source image resolution.
Instead of resizing all input images to the same resolution, we resize the reference image, face mask, and occlusion mask to $\frac{1}{2}\times$ size of the source to save the memory usage.

\paragraph{Model architecture.}
Our FusionNet illustrated in \figref{fusion_net} has N-level pyramids for the encoder and decoder. 
At each level, the encoder block uses 1 \textsc{Conv} layer, followed by another \textsc{Conv} that uses a stride of 2 to downsample the features by 2. 
The decoder block at each layer uses 2 \textsc{Conv} layers followed by a bilinear upsampling layer to upsample the features. 
Except the output layer, each \textsc{Conv} layer is followed by a \textsc{ReLU} activation layer.
The features in each encoder block are fed to the decoder block at the same level through skip connections.
In the first level, the encoder extracts features from the source image and downsamples them by $2\times$. 
In the second level, we concatenate the warped reference image, face mask, and occlusion mask as inputs and extract reference features. 
The features from source and reference are fused and become the input features to the second pyramid level.
In the last decoder block, the upsampled feature is fused with the source feature (before downsampling). 
We apply another \textsc{Conv} layer to generate a 3-channel image as the output.
We also adopt global residual learning such that the output image is a summation of the source image and the model prediction. 
The output image is represented by:
\begin{equation}
    \fusedimage = \fusionnet(\srcimage, \warpedrefimage, \facemask, \occlusionmask) + \srcimage.
\end{equation}

\paragraph{Optimizing FusionNet for mobile devices.}
We start exploring the architecture configuration from a 7-level FusionNet.
While the 7-level FusionNet (the numbers of feature channels at each level are 16, 32, 64, 128, 256, 512, and 1024) can produce high-quality results, the model size is too heavy (102 MB after quantizing to fp16) to be deployed on mobile devices.
We first optimize our FusionNet by reducing the number of pyramid levels, and we find the 5-level FusionNet (16, 32, 64, 128, and 256 channels in each level) can generate similar quality results as a 7-level FusionNet.
Further reducing the pyramid level or the number of feature channels (e.g., from 16 to 8 at the first level) will degrade deblurring quality significantly.
We also explore a similar architecture change as PWC-Net to replace each $3 \times 3$ conv layer with depth-wise conv layers.
However, we also observe quality drop despite the reduction in latency and memory usage.
Finally, we use \textsc{Add} instead of \textsc{Concat} to fuse skip connections with the decoder features to avoid extra buffer.
\textsc{Concat} requires an extra buffer of $H \times W \times 32$ in the first pyramid level, which takes about 288 MB memory for an input size of $1536 \times 1536$.
Instead, we use \textsc{Add} to reuse the existing buffer for skip connection.
This extra 288 MB memory is critical to system stability on mobile phones.

\begin{figure}[t!]
    \centering
    \footnotesize
    \renewcommand{\tabcolsep}{1pt} 
    \includegraphics[width=\linewidth]{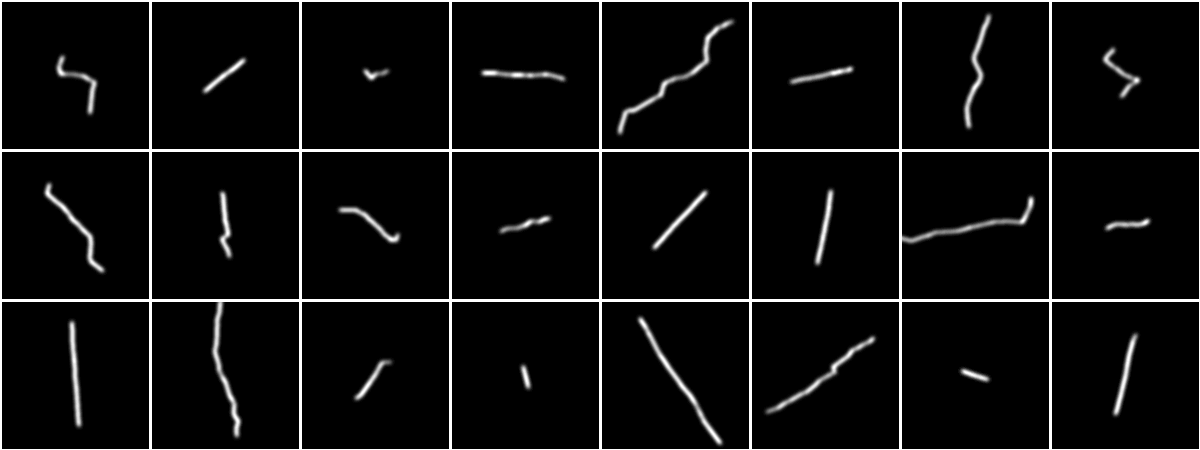}
    \caption{\textbf{Synthetic motion blur kernel.}
        We randomly sample a 2D camera trajectory to generate motion blur kernels for generating synthetic motion-blurred images.
    }
    \label{fig:blur_kernels}
\end{figure}

\paragraph{Training data.}
It is nearly impossible to collect paired blurry input, \ie face with motion blur, and the ground-truth, \ie static face of the same scene at the same moment, in a large scale. Therefore, we resort to synthetic data to train our FusionNet.
We first capture a pair of \wide~ and \ultrawide~ images from static subjects and treat \wide~ and \ultrawide~ images as ground-truth $\gtimage$ and reference, respectively.
We then generate motion blur kernels by randomly sampling a camera trajectory~\cite{boracchi2012modeling} and convolving the ground-truth image with the blur kernel to generate synthetic blurred images as the source. 
\figref{blur_kernels} shows examples of our synthetic motion blur kernels.
The maximum kernel size is $150 \times 150$ with a non-linearity of 0.5 (non-linearity = 0 and 1 correspond to linear lines and highly curved trajectories, respectively).
We restrict the motion blur to the area covered by the face mask and apply a ghosting effect around the subject boundary through alpha blending between the blurred image and ground-truth, as described below:
\begin{equation}
    \srcimage = \blurmask \cdot (\gtimage \otimes \blurkernel) + (1 - \blurmask) \cdot \gtimage + \noise
    \label{eq:blur_model}
\end{equation}
where $\blurkernel$ is blur kernel, $\otimes$ denotes 2D convolution, $\blurmask = \facemask \otimes \blurkernel$, and $\noise$ is realistic Gaussian-Poisson noise~\cite{foi2008practical}.

\begin{figure}[t!]
    \centering
    \footnotesize
    \renewcommand{\tabcolsep}{1pt} 
	\newcommand{\figurewidth}{0.49\linewidth} 
    \begin{tabular}{cc}
        \includegraphics[width=\figurewidth]{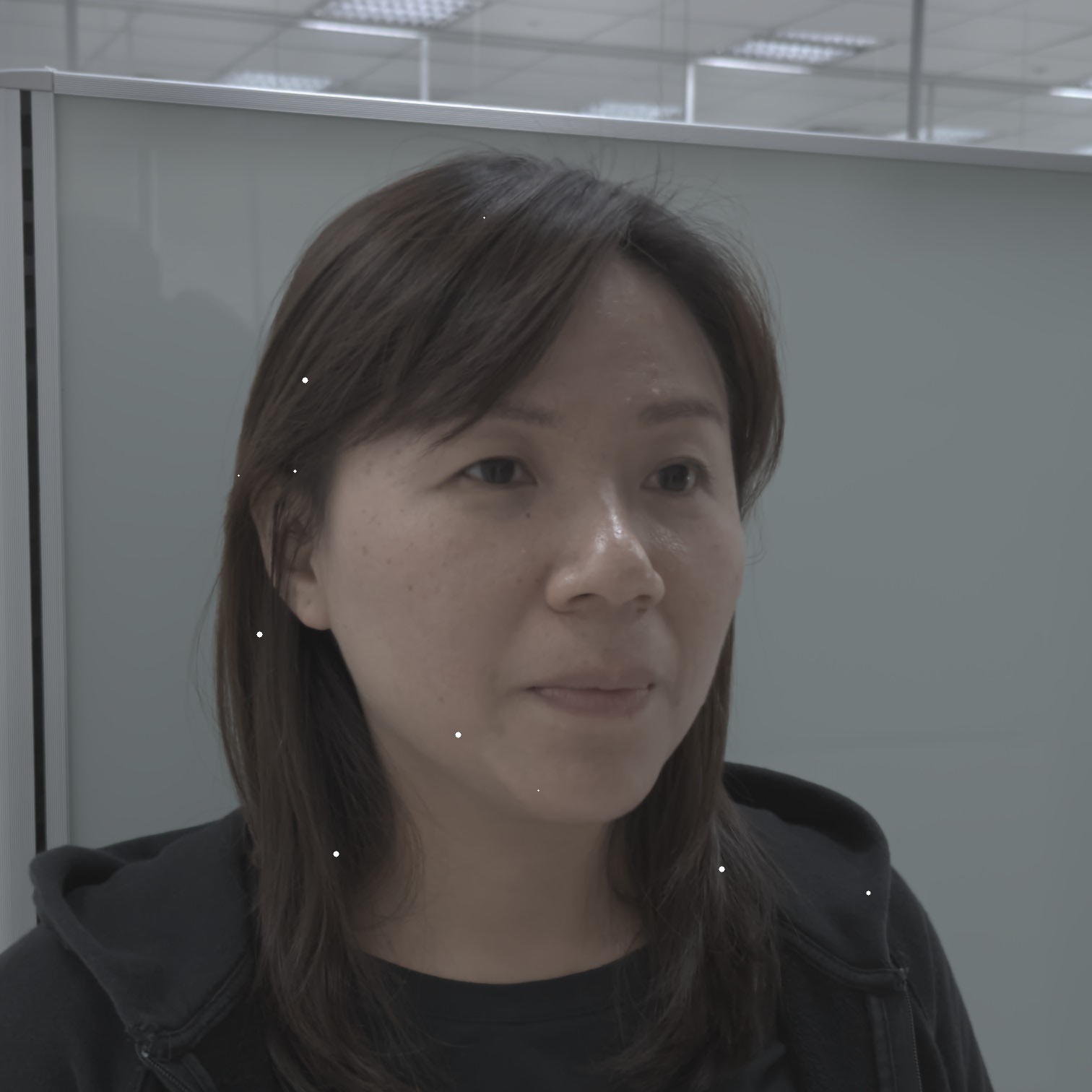} &
        \includegraphics[width=\figurewidth]{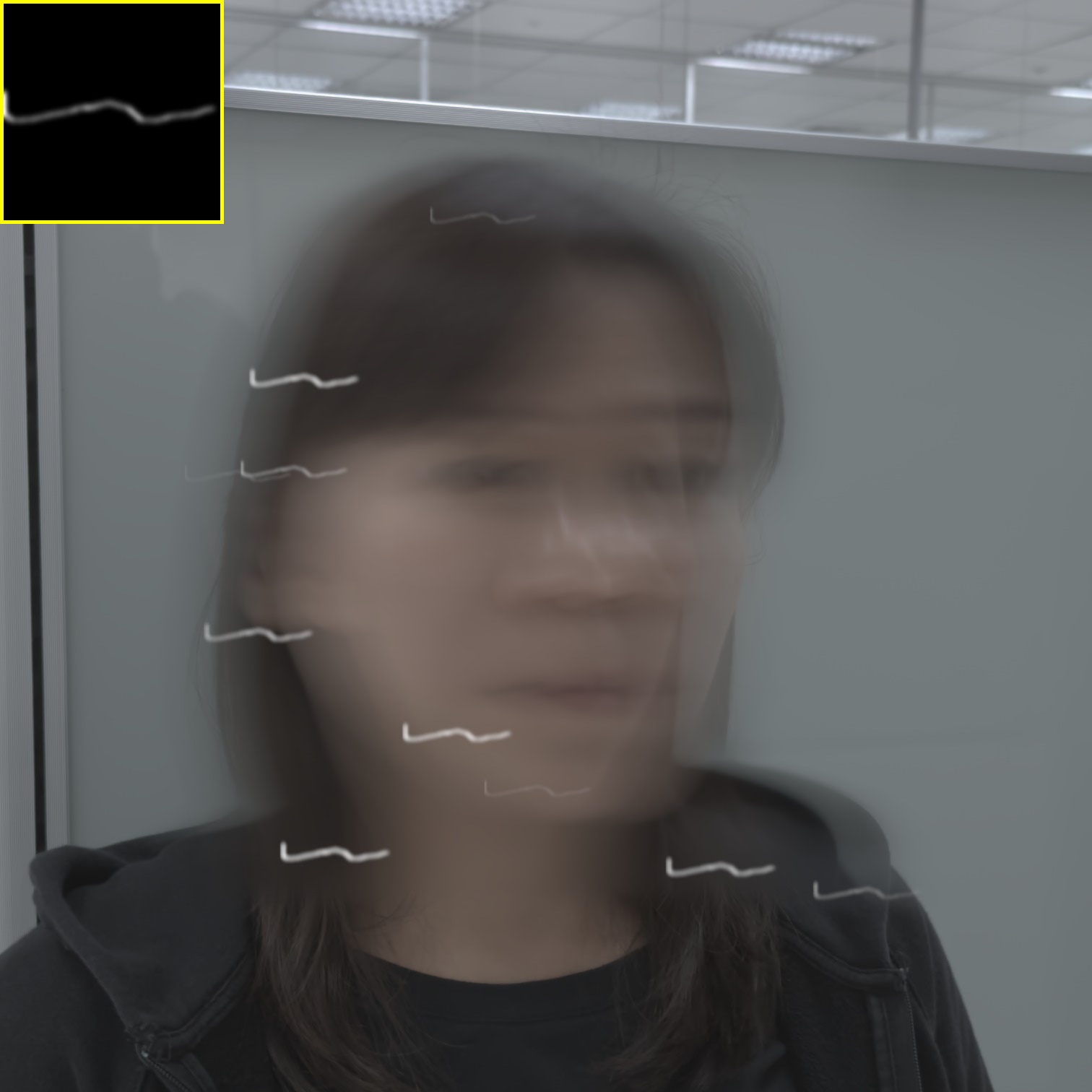} 
        \\
        (a) Ground truth image &
        (b) Synthetic blur 
    \end{tabular}
    \caption{\textbf{Synthetic training data.} 
        (a) We add synthetic highlight on the sharp image captured from \wide~ camera as ground truth. (b) We then randomly sample a blur kernel to simulate motion blur and saturated light streaks from (a). The pairs of (a) and (b) form our synthetic data for supervised training.   
    }
    \label{fig:training_data}
\end{figure}

\paragraph{Synthetic highlights.} In real images, reflection highlight from the eyes or other accessories often leads to saturated light streaks due to motion blur.
These light streaks often saturate and do not follow the linearity assumed by the convolution model in Eq.~\eqnref{blur_model}, resulting in poor performance when training our model on synthetic data.
To tackle this problem, we add random light dots to the ground-truth image before applying motion blur. 
We sample the synthetic light dots through uniform distribution over the radius between $[1, 3]$ pixels, and intensity between $[2, 5]$.
After blurring, we create realistic light streaks as shown in~\figref{training_data}. 
Training with synthetic light streaks effectively reduces motion blur on light streaks during test time (see~\figref{highlight}).
In total, we create about 2594 pairs of \wide~ and \ultrawide~ images for training data.

\paragraph{Training losses.}
We adopt the following loss functions to train our FusionNet:

\emph{- Content loss}: We compute the L1 difference between the fusion result and ground-truth:
\begin{equation}
    \contentloss = \onenorm{\fusedimage - \gtimage}.
\end{equation}

\emph{- Perceptual loss}: We also compute the weighted L1 difference between the features of the fusion result and ground-truth extracted from a pre-trained VGG19 model~\cite{simonyan2014very}:
\begin{equation}
    \vggloss = \sum_j w_j \onenorm{VGG_j(\fusedimage) - VGG_j(\gtimage)}.
\end{equation}
We use features from $\textsc{conv1-2}$, $\textsc{conv2-2}$, $\textsc{conv3-2}$, $\textsc{conv4-2}$, and $\textsc{conv5-2}$, and $w_j = 1/2.6, 1/4.8, 1/3.7, 1/5.6$ and 10/1.5, respectively\footnote{We follow the settings in the source code of~\cite{chen2017photographic}.}.

\emph{- Color-consistency loss}: 
Even with AE, AWB synchronization, and our color matching step, the reference image may still have a different color than the source image. 
We observe color-shifting artifacts on hair or forehead in our fusion results. 
To alleviate this issue, we introduce a color consistency loss by matching the local patch mean between the model output and the source image. 
Specifically, we apply a Gaussian filter with a large kernel to blur both the model output and source image, and then compute the L1 loss between the smoothed images:
\begin{equation}
    \colorloss = \onenorm{\mathcal{G}_\sigma(\fusedimage) - \mathcal{G}_\sigma(\srcimage)},
\end{equation}
where $\mathcal{G}_\sigma$ is Gaussian filter with a standard deviation $\sigma$.
We set $\sigma = 20$ in our experiments.

The total loss is a weighted sum of the above losses:
\begin{equation}
    \contentweight \contentloss + \vggweight \vggloss + \colorweight \colorloss
\end{equation}
where we set $\contentweight = 1$, $\vggweight = 2$, and $\colorweight = 1$ in our experiments. We train our model using Tensorflow and ADAM~\cite{kingma2014adam} optimizer with decayed learning rate. The model converges after 1M iterations. The PSNR reaches 37.97 dB on the validation set.

\subsection{Blending}
\label{sec:blending}
To focus the deblurring on faces and keep the background unchanged, we adopt an alpha blending to merge the FusionNet output and source image, as described below:
\begin{equation}
    \finalimage = \blendingmask \cdot \fusedimage + (1 - \blendingmask) \cdot \srcimage,
    \label{eq:blending}
\end{equation}
where the blending mask $\blendingmask$ is the face mask subtracted by the occlusion mask and reprojection error map:
\begin{equation}
    \blendingmask = \min(\facemask - \alpha \cdot \occlusionmask - \beta \reprojectionmask, 0).
\end{equation}
We set $\alpha = 5.0$ and $\beta = 2.0$.
The reprojection error map is calculated from the difference between the source image and the warped reference image:
\begin{equation}
    \reprojectionmask = \max_{R,G,B}(| \srcimage - \warpedrefimage | ).
\end{equation}
We use the reprojection error map and occlusion mask to gracefully fallback error-prone pixels to the source pixels. 
With the soft blending, the results look natural and artifact-free.

\subsection{Post-processing}
\label{sec:post-processing}

After blending the fusion result with the source linear image, we apply chroma denoising, local and global tone-mapping, and dehazing described in ~\cite{hasinoff2016burst} to generate the final 8-bit gamma-corrected image in the sRGB color space.
While FusionNet removes motion blur, the results look slightly over-smooth due to the fact that \ultrawide~ has lower optical resolution than \wide.
To remove mild blur and enhance sharpness, we resort to a single-image sharpening method based on Polyblur~\cite{delbracio2021polyblur}, which estimates a Gaussian blur kernel from the input image and approximates the deconvolution with a series of polynomial filters. 
We modify Polyblur by limiting the support of the kernel estimation step to the face area, so that the estimated kernel accurately represents the local blur on the face area.

\subsection{Fallback Mechanism}
\label{sec:fallback}
The reference quality, \ie \ultrawide, is critical to the deblur quality. 
When the reference image shows poor quality, \eg is too noisy or fails to align to the source image, we observe objectionable artifacts. Our strategy in these cases is to automatically fall back to the source image based on image quality heuristics.

We calculate face motion from the center of the face rectangle between adjacent frames in \wide~ raw burst. (1) If the motion is too small, we will skip fusion. (2) If the timestamp difference between \wide and \ultrawide~ is larger than 20ms, we fall back since the alignment is too challenging for our optical flow. (3) When the sensor gain is higher than a threshold (=160 in our work), \eg low-light conditions, excessive noise due to insufficient exposure time, we skip fusion.

Finally, we check (4) the difference between the gamma-corrected fusion output and source image over the face area based on a masked mean square error (MSE) described below:
\begin{equation}
    \frac{(\facemask) \cdot \twonorm{\gammafusedimage - \gammasrcimage}}{\sum (\facemask )}.
\end{equation}
If the MSE is smaller than a threshold (0.25 in our work), we fall back to the source image because the fusion result improvement is insignificant. 

\subsection{Hardware acceleration}
Our fusion pipeline contains 3 ML models: optical flow (PWC-Net), fusion deblur (FusionNet), and subject segmentation (UNet). To deploy them on the mobile phone, we convert the models into TensorFlow lite format~\cite{tensorflow} 
and delegate inference to the accelerators. The SoC in our experiment has a Mali G78 GPU and a neural processing unit (NPU) for efficient convolution ~\cite{yazdanbakhsh2021evaluation}. 
We choose to delegate segmentation to NPU and other operations to GPU. The latter performs more efficiently for the gather operators used by PWC-Net and the large memory footprint required by FusionNet.

\section{Adaptive Streaming}
 \label{sec:adaptive}
 
\begin{figure}
    \centering
    \footnotesize
    \includegraphics[width=\linewidth]{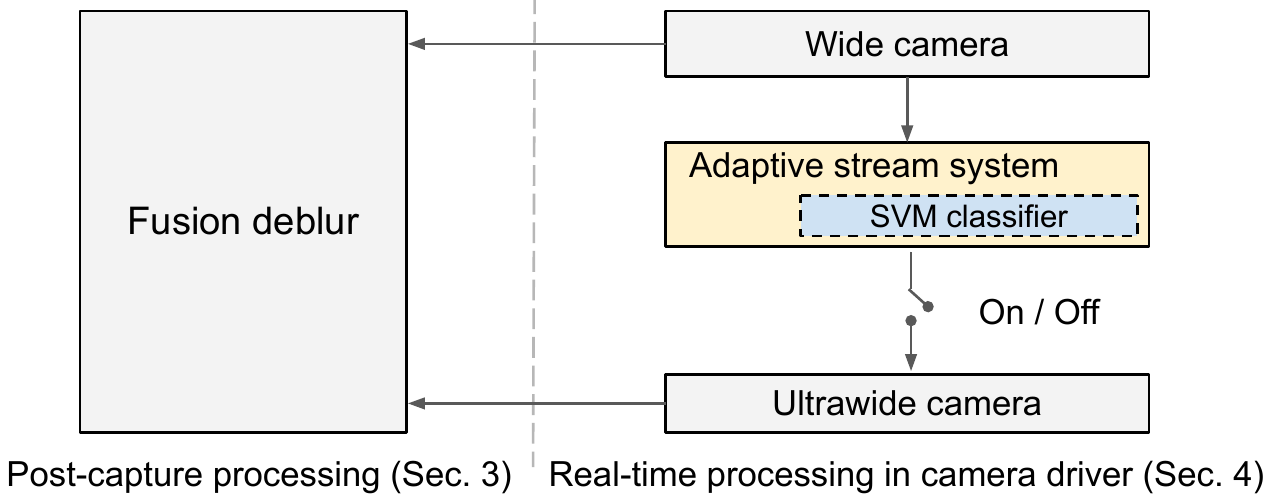}
    \caption{\textbf{Adaptive Streaming Design.}
        The adaptive streaming system resides in the camera driver, passing camera metadata to the ML classifier to identify if the human face is suffering from motion blur, and turns on the \ultrawide~ camera dynamically.
    }
    \label{fig:adaptive_streaming_design}
\end{figure}

Our algorithm requires a mobile phone to stream both \wide~ and \ultrawide~ cameras. 
However, continuously streaming both \wide~ and \ultrawide~ camera consumes high power (+400 mW) unsustainable for thermal status. Moreover, it requires a large additional memory footprint (+160 MB). 
To save amortized power, we present an adaptive stream system that dynamically enables and disables \ultrawide~ camera based on real-time motion analysis, as illustrated in ~\figref{adaptive_streaming_design}.
Our adaptive streaming system determines whether the subject in the frame $t$ demonstrates excessive motion that leads to motion blur, and, if so, streams \ultrawide~ at $t+1$ to capture the short-exposed reference. 
The motion analysis and decision need to be made within the duration of 1-frame, \ie 33ms when the camera streams at 30 FPS. 
To meet the real-time requirement, we propose a lightweight SVM model that takes feature vectors from camera metadata relevant to motion blur and outputs a binary flag indicating whether to turn on \ultrawide~ camera.
We now describe our SVM model and implementation for mobile camera systems.

\subsection{Real-time Motion Analysis using SVM}
Modern ISPs in smartphones offer a temporal denoising block based on real-time multi-frame alignment and merge. We take the alignment map computed from the patch-based inverse Lucas-Kanade algorithm~\cite{ehmann2018real} between the adjacent \wide~frames as the motion flow and use the (1) average flow strength in the face region, and (2) the maximal flow as the input vector for the SVM model. We augment the face motion feature by calculating the (3) average gradient strength in the face area. We include the (4) exposure time in the feature vector due to the high correlation to motion blur. Furthermore, we include (5) sensor gain in the feature vector since excessive noise is detrimental to the deblur quality. 

If motion blur is insignificant in \wide, the fusion result appears similar to \wide, and we prefer to turn off \ultrawide~ to save the power. We leverage this observation to label the ground truth of the output for the SVM model. 
We collect 1777 pairs of \wide, \ultrawide and the feature vectors consisting of (1)-(5) above at various motion strengths and lighting conditions, and generate the deblur results from the pairs. 
We then manually examine the deblur output and \wide, and label positive if the results are visually better than \wide. Our SVM model achieves $6\%$ and $11\%$ false positive and negative rate, respectively, on the validation set consisting of 500 captures.

\subsection{Implementation}
The Android camera framework (Camera2 API)~\cite{camera2api} allows the camera app developer to stream each camera dynamically, but leads to an 11-frame delay between motion analysis results and the moment when \ultrawide~ starts to stream in our experiment. 
If the users capture a motion shot within these 11 frames, our system may fail to stream \ultrawide~ for fusion deblurring.
Alternatively, we reduce the frame delay to 7 frames by integrating the adaptive stream system into the camera driver. In our experiment over 600K shots by our users, only $8.1 \%$ shots were captured during the 7 frames.

\section{Camera image processing pipeline}
\label{sec:isp}

We describe the requirements of the image signal processing (ISP) pipeline for integrating our fusion deblurring (\secref{fusion}) with photo capture.  
Before the shutter is pressed, fusion deblurring requires careful auto exposure, auto white-balance, and temporal synchronization between \wide~ and \ultrawide~ camera in a real-time processing pipeline.

\paragraph{Auto exposure.}
Auto-exposure in the camera ISP decides a target total exposure time (TET) that represents the scene brightness. 
TET is factorized into the exposure time $t_e$ and sensor gain $\gamma$, \ie, $\text{TET}=\gamma t_e$ ~\cite{liba2019handheld}. 
Our system determines the $t_e$ and $\gamma$ for \ultrawide~ from \wide~ camera.
To capture facial details, we lock the exposure time of \ultrawide~$N$ times faster than \wide, \ie $t_{e,UW} = \frac{1}{N}t_{e,W}$. To match the total scene brightness, we determine the TET for \ultrawide~ by calibrating the sensor sensitivity using a synchronization ratio $\mu$  at ISO100 gain, \ie $TET_{UW}=\mu TET_{W}$, where $\mu=3.7$ in our testing device. Finally, we determine the sensor gain of \ultrawide~ by dividing $TET_{UW}$ by the exposure time of \ultrawide, \ie $\gamma_{UW} = N\mu \gamma_{W}$. We test $N=2$ and $N=4$ and find both lead satisfying deblurred results.

\paragraph{Auto white-balance.}
In the camera ISPs, the linear colors are determined by the color correction matrix ($CCM$) and lens-shading correction ($LSC$) table. We apply Eq.~\ref{eq:ccm_normalization} to match $CCM$. In theory, the sensor-dependent $LSC$ varies by scene illumination. In practice, since faces shown in \wide~ often appears near the center in \ultrawide, and $LSC$ is nearly fixed at the camera center, we apply the factory-calibrated $LSC$ table to \ultrawide~ and ignore the scene-dependent effect. 

\paragraph{Timestamp synchronization.}
Our fusion deblurring requires the maximal timestamp difference between \wide~ and \ultrawide~ camera to be less than 20ms when streaming at 30FPS. 
In practice, frames may be misaligned due to frame drop, camera timestamp drifts, and different sensor integration time margins. 
Most SoCs on premium phones support hardware sensor synchronization across multiple cameras and achieve tolerance of less than 1ms. 
However, we found the synchronization causes preview jank when \ultrawide~ runs in adaptive mode (\secref{adaptive}), since our SoC only allows syncing \ultrawide~ to \wide~ but not the other way, and drops frames in \wide~ during camera preview.
Therefore, we implemented a software synchronization by calculating the frame time difference between \ultrawide~ and \wide, and send the difference to \ultrawide~ camera sensor to adjust the frame rate dynamically. 

\paragraph{Power-efficient \ultrawide~ camera ISP.} 
Even with the proposed adaptive streaming system (\secref{adaptive}), streaming two cameras simultaneously still consumes significant power. While the \wide~ camera ISP needs to produce the tone-mapped frames for preview, the \ultrawide~ camera ISP only needs to produce the RAW frames. We take the optimization opportunity by customizing the \ultrawide~ camera ISP. We disable the ISP blocks after RAW, including camera 3A, face detection, demosaic, and RGB-to-YUV conversion.

\paragraph{Dual zero shutter lag burst.}
We adopt zero shutter lag (ZSL) capture to avoid missing the fleeting moment of motion due to shutter lag. ZSL employs a ring buffer that continuously captures the RAW frame produced by the sensor, then determines the frame for post-capture processing at the shutter time, referred to as the reference frame. We adopt ZSL for both \wide~ and \ultrawide~ captures. We pair the frames of the two ZSL ring buffers based on the nearest-timestamp criteria. At shutter time, we determine the reference frame from the \wide~ ZSL ring buffer using the algorithm described in ~\cite{hasinoff2016burst} and then use the reference frame timestamp from \wide~ to determine reference frame in \ultrawide~ ZSL ring buffer.

\section{Results}
\label{sec:results}

\begin{figure*}
    \centering
    \footnotesize
    \renewcommand{\tabcolsep}{1pt} 
	\newcommand{\figurewidth}{0.235\linewidth} 
    \newcommand{\addimage}[1]{
        \begin{overpic}[width=\figurewidth]{figures/our_results/#1_source.jpg}
            \put (2,2) {\white{\textbf{Input}}}
        \end{overpic} &
        \begin{overpic}[width=\figurewidth]{figures/our_results/#1_fusion.jpg}
            \put (2,2) {\white{\textbf{Our result}}}
        \end{overpic}
	}
    \begin{tabular}{cccc}
        \addimage{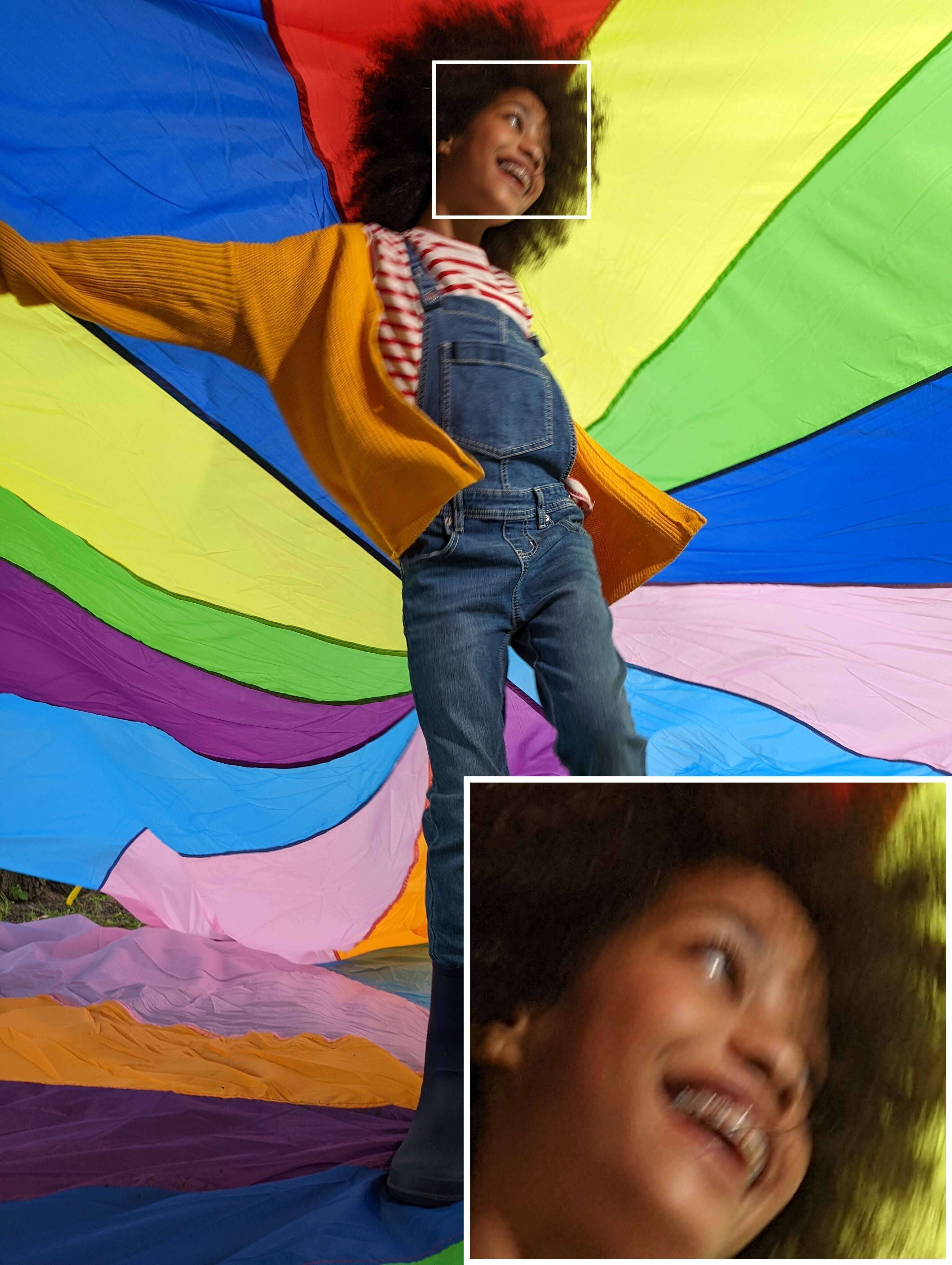} &
        \addimage{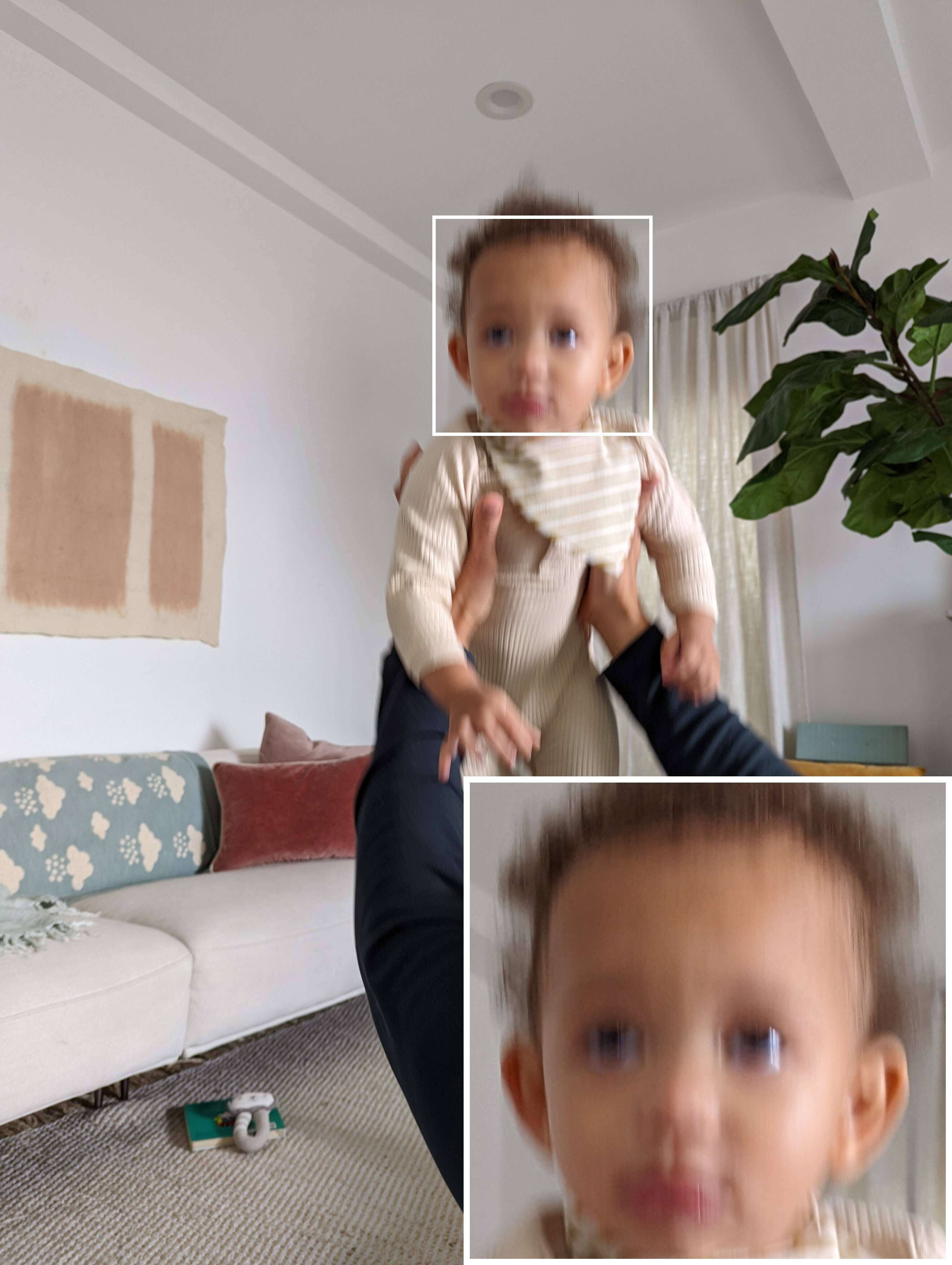} \\
        \addimage{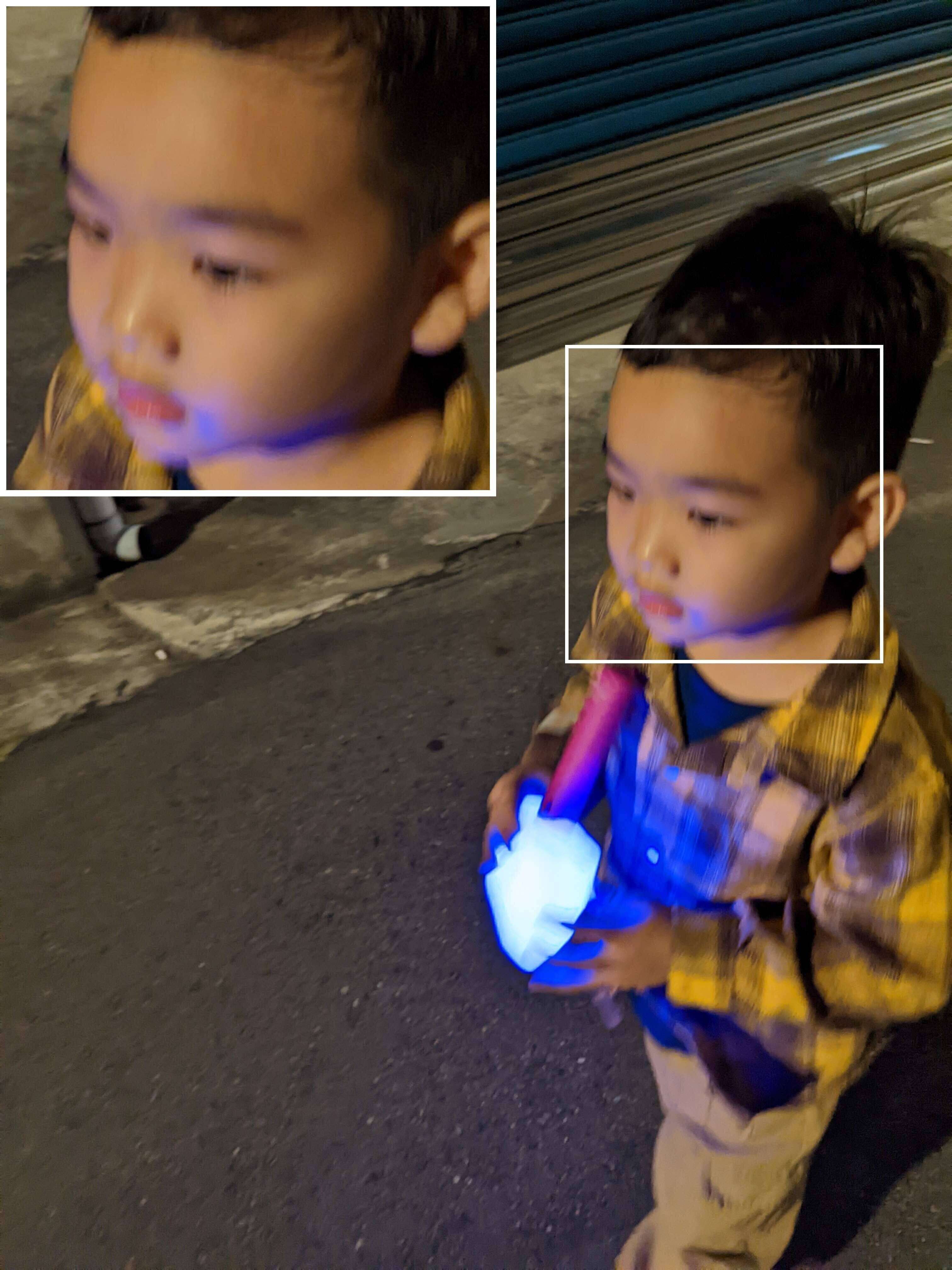} &
        \addimage{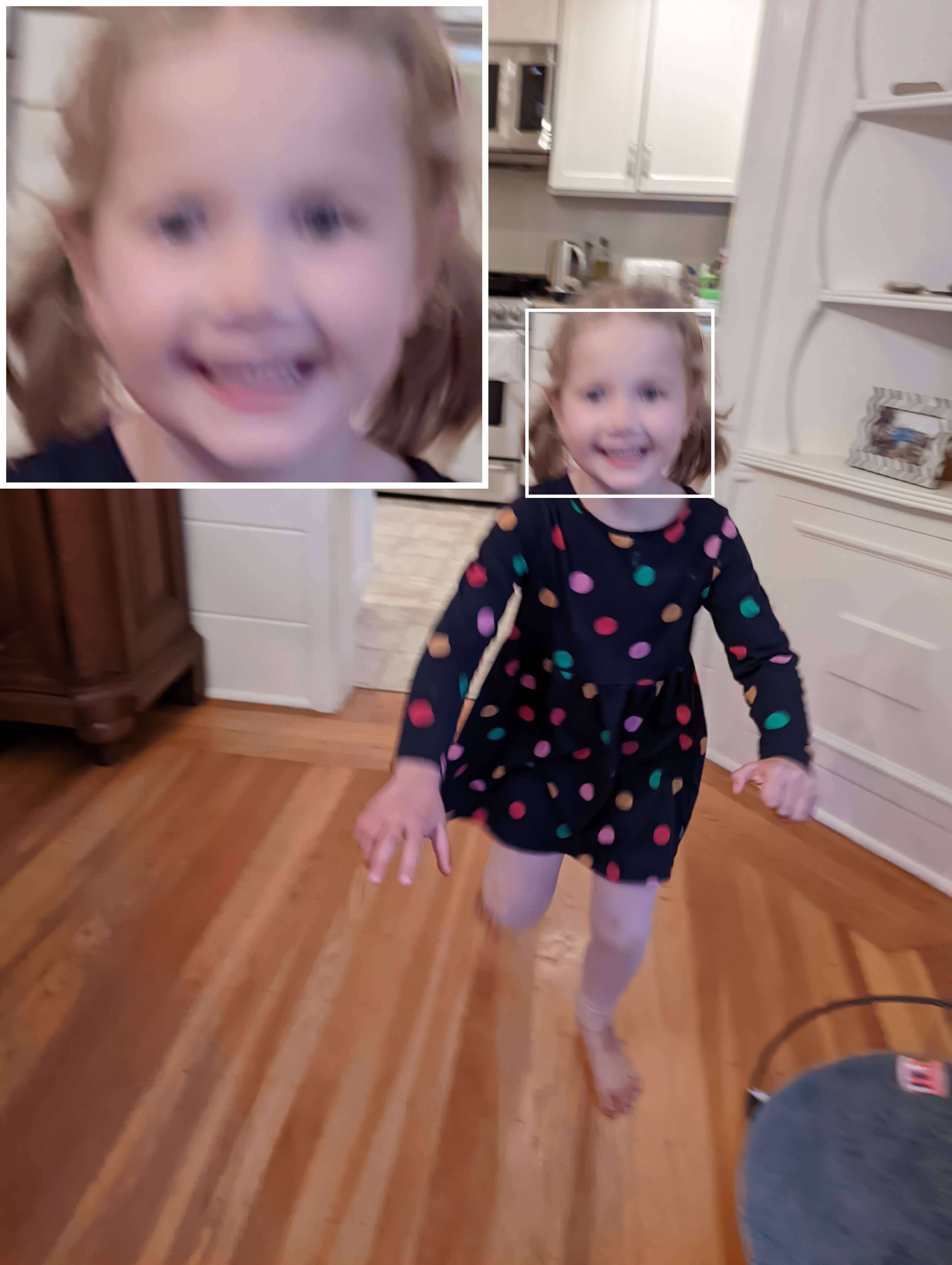} \\
        \addimage{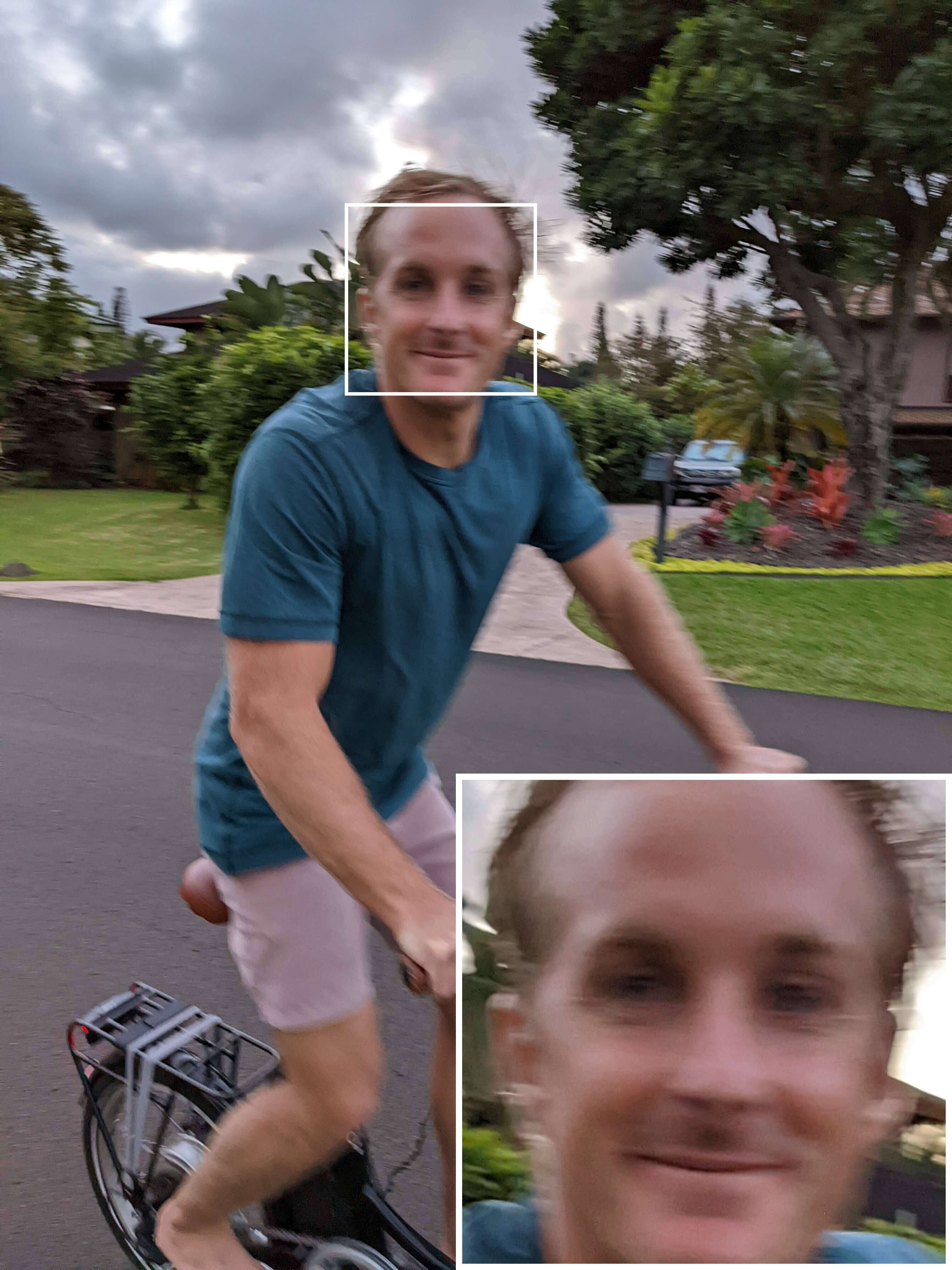} &
        \addimage{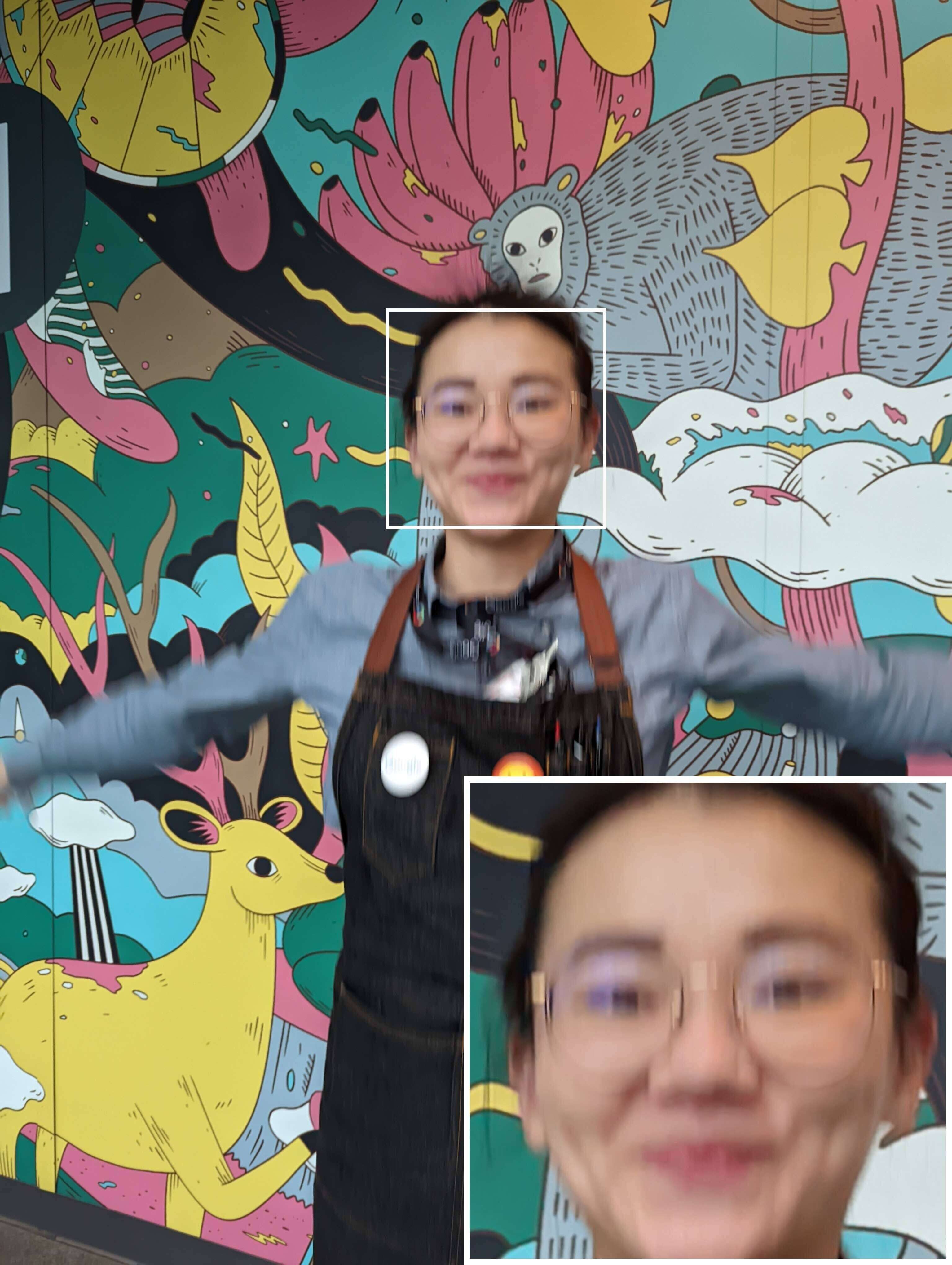} \\
        \addimage{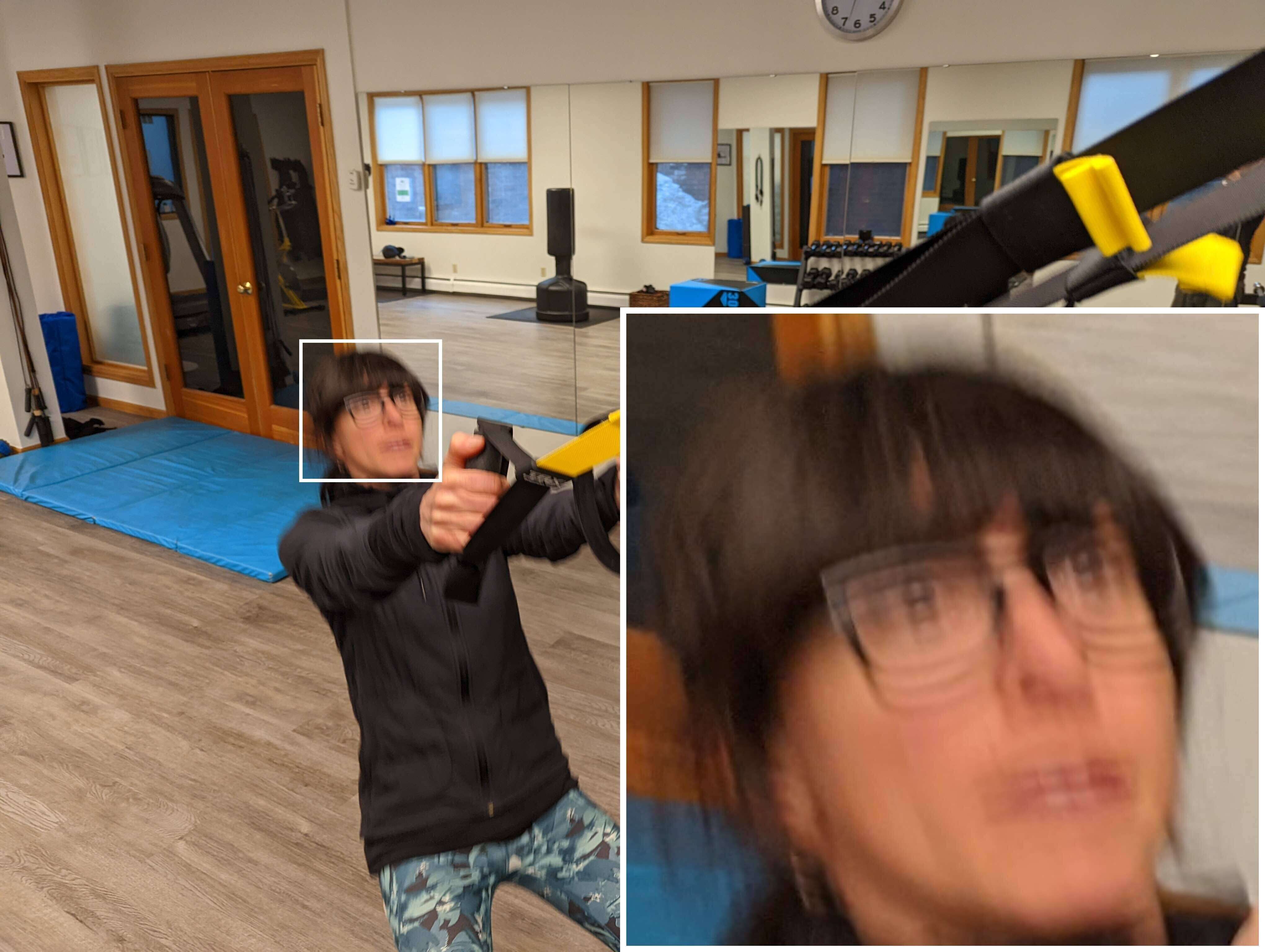} &
        \addimage{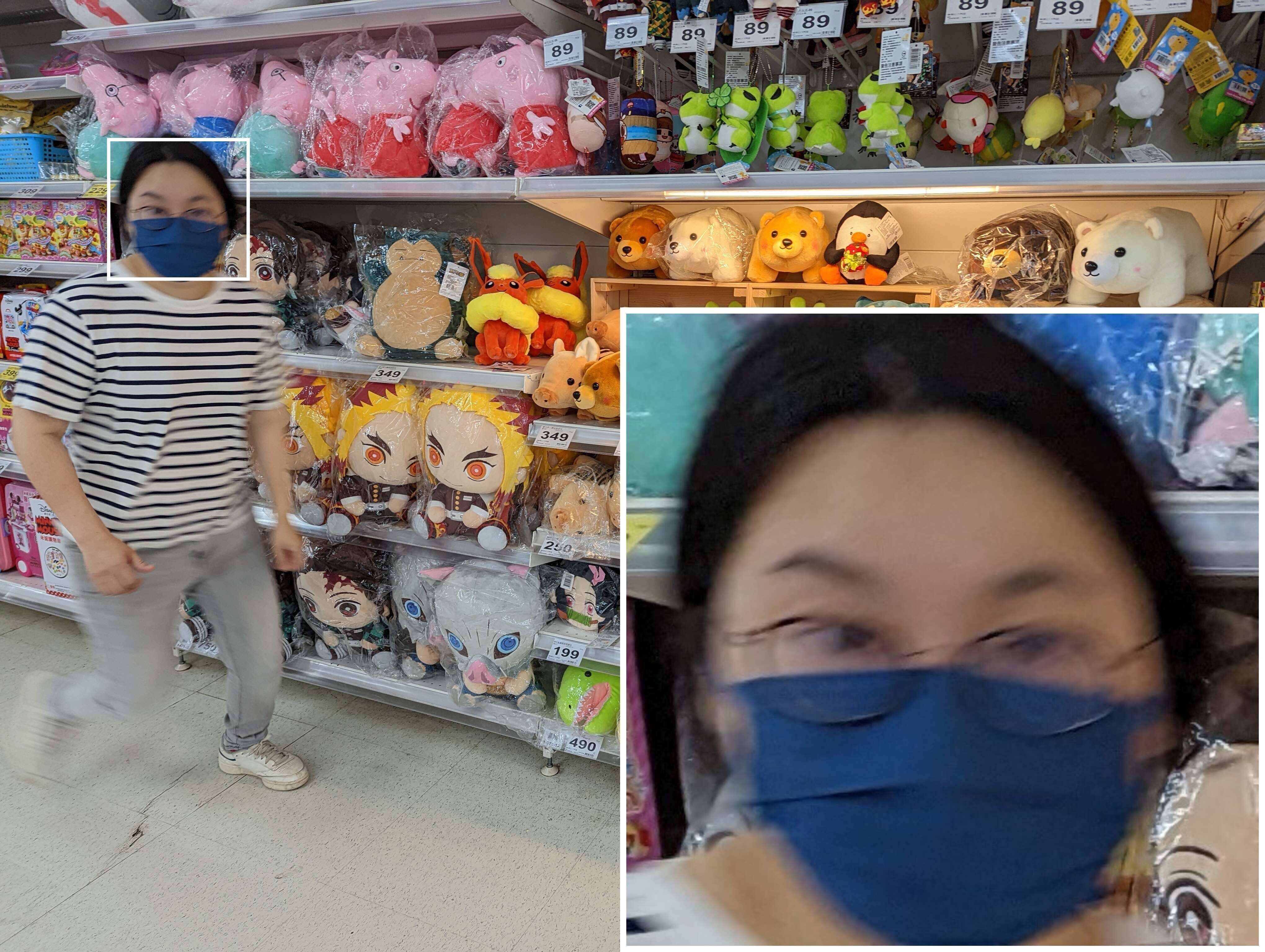} \\
    \end{tabular}
    \vspace{-2mm}
    \caption{\textbf{Our results.} In each pair, we show blurry input at left, and our deblurred result at right. We include the cases of challenging and fast motion from day-to-day scenarios, over subjects of diverse skin colors, face poses, ages, and genders, and under indoor, outdoor, and low-light conditions. The insets clearly show that our method recovers authentic and sharp details, making the captures usable and memorable. Please refer to the supplemental material for a complete list of results.
    }
    \label{fig:our_results}
\end{figure*}

\begin{figure*}
    \centering
    \footnotesize
    \renewcommand{\tabcolsep}{3pt} 
	\renewcommand{\arraystretch}{0.9} 
	\renewcommand{\imagewidth}{0.173\linewidth} 
    \newcommand{\addimage}[9]{
        \includegraphics[width=\imagewidth]{figures/compare_academic/#1_source.jpg} &
        \includegraphics[width=\imagewidth]{figures/compare_academic/#1_DeblurGANv2.jpg} &
        \includegraphics[width=\imagewidth]{figures/compare_academic/#1_MPRNet.jpg} &
        \includegraphics[width=\imagewidth]{figures/compare_academic/#1_MIMO-UNetPlus.jpg} &
        \includegraphics[width=\imagewidth]{figures/compare_academic/#1_UMSNFaceDeblurring.jpg} \\
        Source (NIMA, $\Delta$NIMA) & DeblurGANv2 (#2) & MPRNet (#3) & MIMO-UNet+ (#4) & UMSN (#5) \\
        \includegraphics[width=\imagewidth]{figures/compare_academic/#1_reference.jpg} &
        \includegraphics[width=\imagewidth]{figures/compare_academic/#1_BurstImageDeblurring.jpg} &
        \includegraphics[width=\imagewidth]{figures/compare_academic/#1_UHDVD.jpg} &
        \includegraphics[width=\imagewidth]{figures/compare_academic/#1_PVDNet_DVD.jpg} &
        \includegraphics[width=\imagewidth]{figures/compare_academic/#1_ours.jpg} \\
        Reference & BurstDeblurring (#6) & UHDVD (#7) & PVDNet (#8) & Ours (\textbf{#9})
	}
    \begin{tabular}{c|cccc}
        \addimage{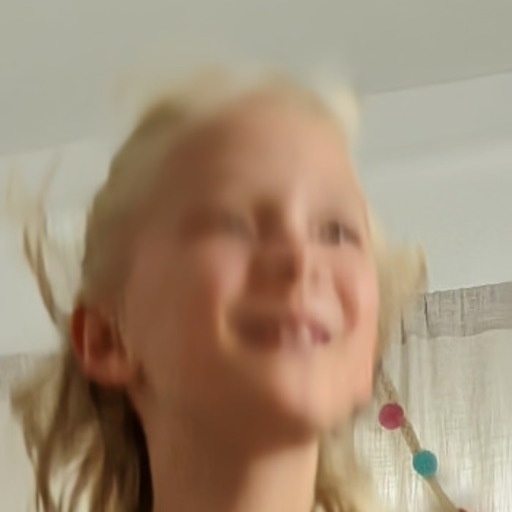}{1.539, 0.449}{1.092, 0.002}{1.289, 0.199}{1.236, 0.146}{1.133, 0.043}{1.268, 0.178}{1.863, 0.774}{2.652, 1.562} \\
        \addimage{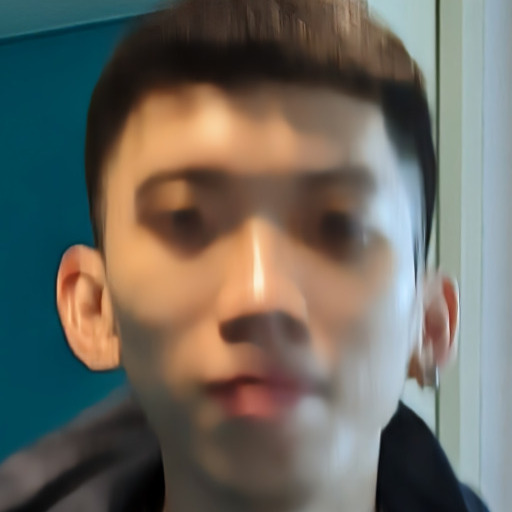}{1.774, 0.657}{1.315, 0.198}{1.400, 0.283}{1.567, 0.450}{1.742, 0.624}{1.261, 0.144}{1.501, 0.383}{4.086, 2.969} \\
        \addimage{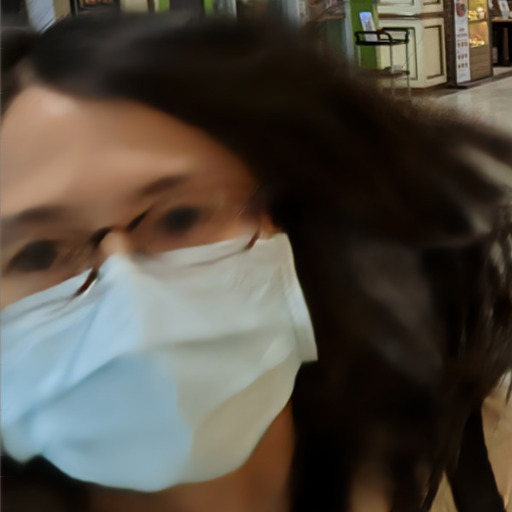}{1.457, 0.399}{1.202, 0.144}{1.151, 0.092}{1.401, 0.343}{1.231, 0.172}{1.424, 0.365}{1.569, 0.510}{2.166, 1.107} \\
    \end{tabular}
    \vspace{-2mm}
    \caption{\textbf{Comparisons with existing methods.} We compare our method with single-image deblurring (MPRNet~\cite{zamir2021multi},
    MIMO-UNet~\cite{cho2021rethinking}, single-image face deblurring (UMSN~\cite{yasarla2020deblurring}),
    multi-frame deblurring (BurstDeblurring~\cite{aittala2018burst}), and video deblurring (UHDVD~\cite{deng2021multi}, PVDNet~\cite{son2021recurrent}) approaches. Our method outperforms other approaches both in $\Delta$NIMA and visual quality.
    }
    \label{fig:compare_academic}
\end{figure*}

\begin{figure}
    \centering
    \footnotesize
    \renewcommand{\tabcolsep}{1pt} 
	\renewcommand{\arraystretch}{0.8} 
	\includegraphics[width=\linewidth]{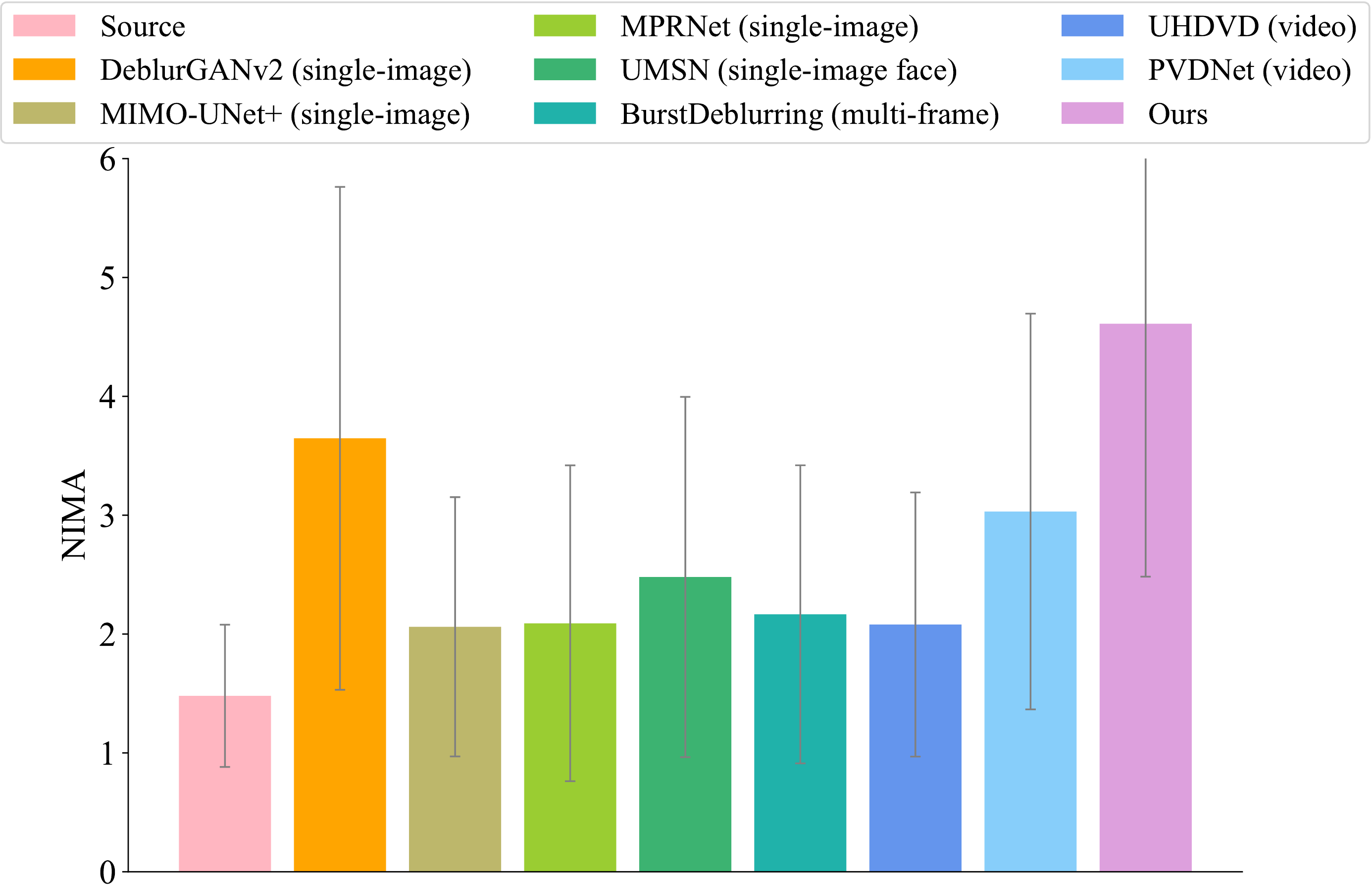} \\
	(a) Average NIMA scores \\
	\hspace{-5mm}\includegraphics[width=0.86\linewidth]{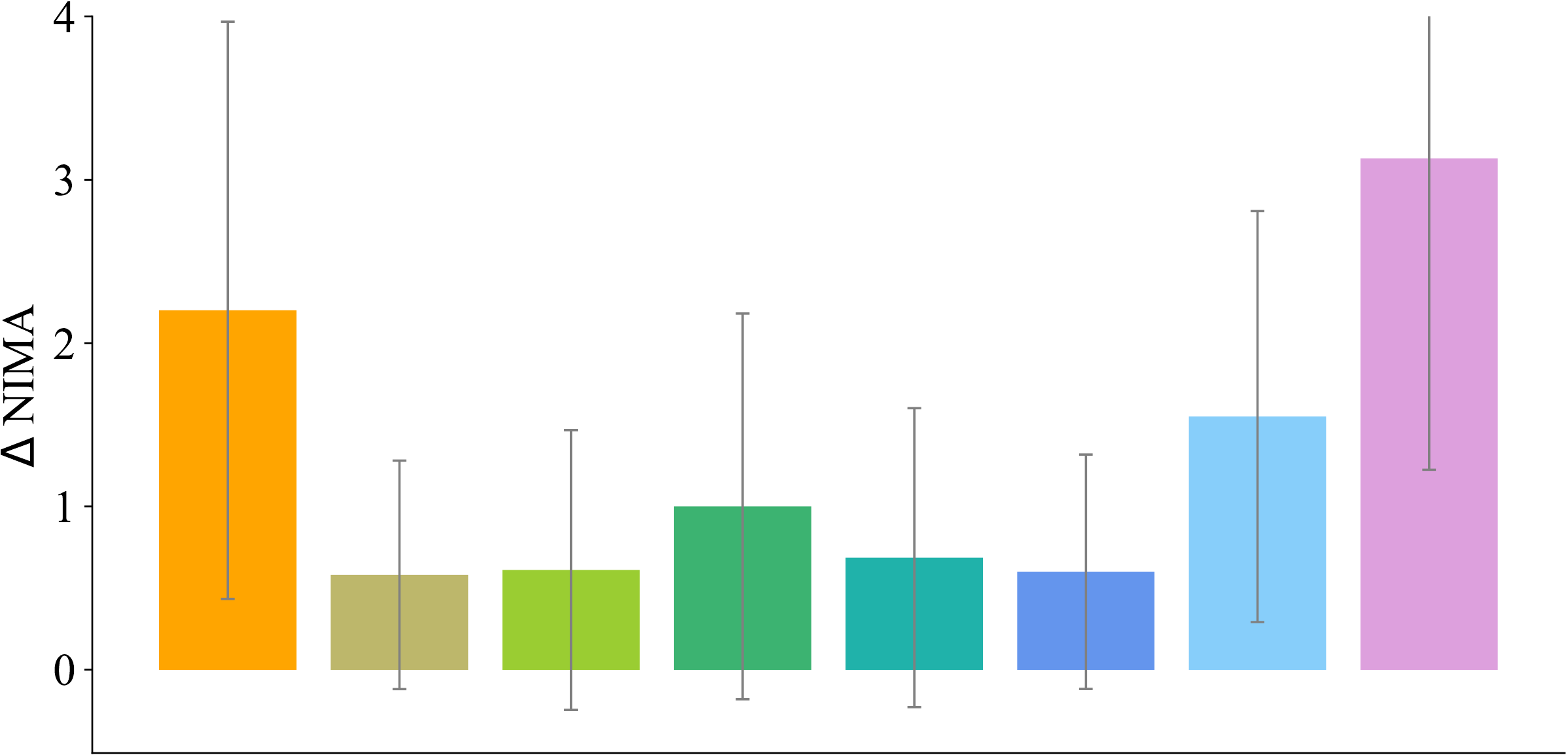} \\
	(b) Average $\Delta$NIMA \\
    \caption{\textbf{Quantitative evaluation.}
        (a) We calculate the no-reference image quality metric using NIMA~\cite{talebi2018nima}.
        Our method outperforms others and achieves the best perceptual quality.
        (b) We compute the $\Delta$NIMA by subtracting the NIMA score before and after the deblurring.
        Some methods lead to NIMA score regression ($\Delta$NIMA$<0$ within one standard deviation range), while our method robustly shows improvement.
    }
    \label{fig:evaluation}
\end{figure}

\figref{our_results} shows the results of our fusion deblurring. Under challenging motions, lighting conditions, and face poses, our method successfully reconstructs facial details without objectionable artifacts. 
We can see that in these examples, the subjects wear glasses and masks (bottom row), or move so fast that the facial features are barely identifiable (left in the bottom row and right in the top row).  
Our method robustly outputs clear faces in these difficult conditions. 
While our training images are synthetic and spatially-invariant blurred images, our FusionNet still generalizes well to dynamic motion blur.
For example, in the left example in the first row of \figref{our_results}, the light streaks on the eyes and teeth show different orientations due to the head rotation.
Our method can reduce the motion blur by fusing the details from the reference image.
The full-resolution outputs and intermediate results are provided in the supplementary material.

\begin{figure*}
    \centering
    \footnotesize
    \renewcommand{\tabcolsep}{1pt} 
	\newcommand{\figurewidth}{0.19\linewidth} 
    \newcommand{\addimage}[3]{
        \begin{overpic}[width=\figurewidth]{figures/compare_commercial/#1_source_crop.jpg}
            \put (2,2) {\white{\textbf{#2 s}}}
        \end{overpic} &
        \begin{overpic}[width=\figurewidth]{figures/compare_commercial/#1_iphone13pro_crop.jpg}
            \put (2,2) {\white{\textbf{#3 s}}}
        \end{overpic} &
        \includegraphics[width=\figurewidth]{figures/compare_commercial/#1_photoshop_crop.jpg} &
        \includegraphics[width=\figurewidth]{figures/compare_commercial/#1_samsung_remastered_crop.jpg} &
        \includegraphics[width=\figurewidth]{figures/compare_commercial/#1_fusion_crop.jpg} 
	}
    \begin{tabular}{ccccc}
        \addimage{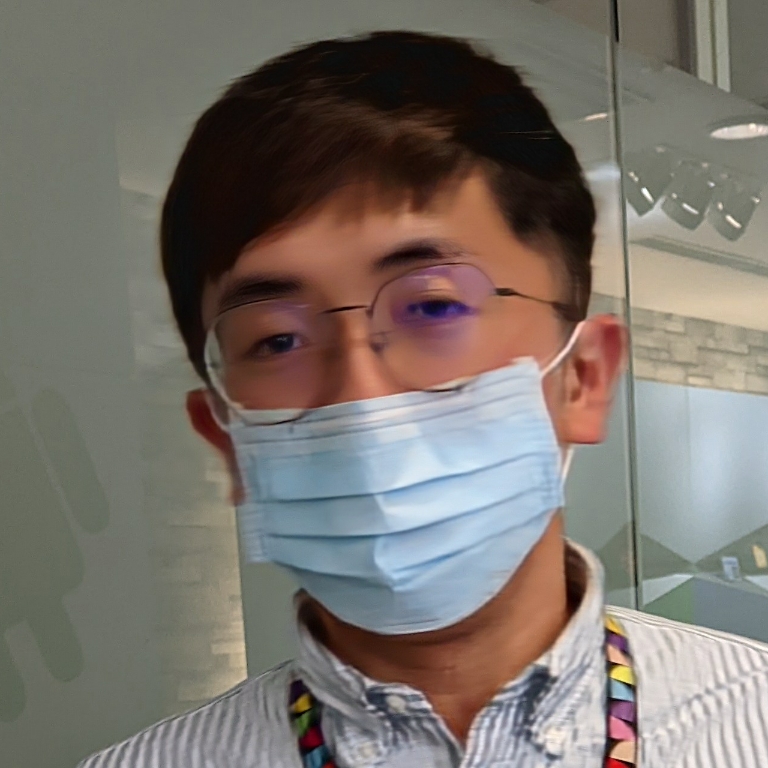}{1/120}{1/120} \\
        \addimage{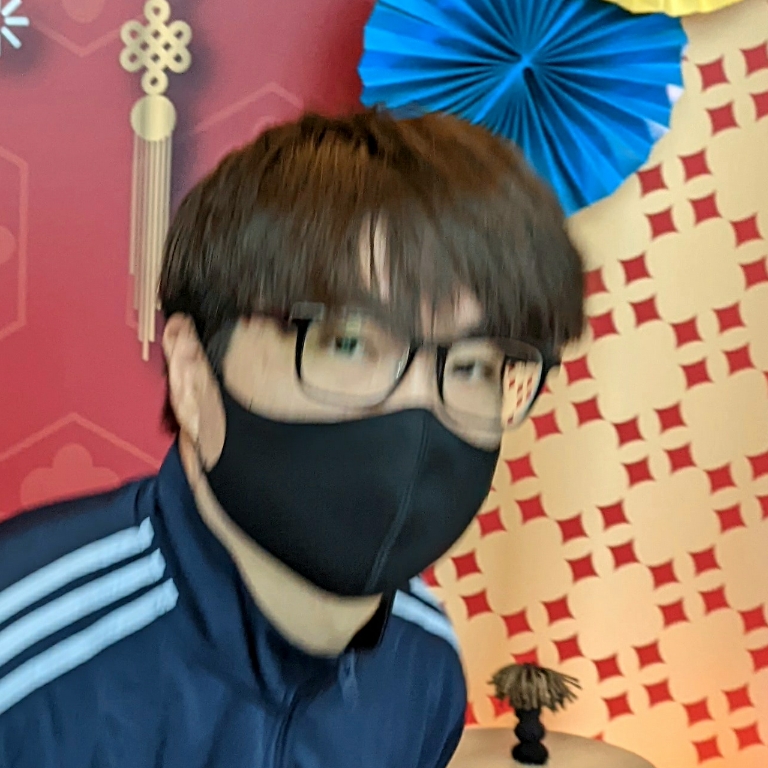}{1/120}{1/60} \\
        (a) Source & (b) iPhone13Pro & (c) Adobe Photoshop  & (d) Samsung Remaster & (e) Ours 
    \end{tabular}
    \caption{\textbf{Comparisons with commercial product.} We evaluate our method on (a) blurry images against (b) iPhone 13 Pro, (c) Shake Reduction filter in Adobe Photoshop, and (d) Samsung Gallery's Remaster feature. 
    Our method removes motion blur effectively and generates the cleanest results in (e).
    }
    \label{fig:compare_commercial}
\end{figure*}

\subsection{Dataset}
To the best of our knowledge, there are no datasets that contain triplets of blurry, noisy, and sharp images. 
Therefore, we collect our own test dataset for evaluation.
We implement our system on an Android-based mobile camera to capture moving subjects under diverse scenes and lighting conditions, including cloudy days, evening, and indoor scenes. 
Subject motions include walking, running, jumping, biking, skateboarding, working out, and playing various sports: basketball, badminton, table tennis, volleyball, etc. We also include subjects of diverse ages and ethnic groups.
Among the 6080 diverse test images collected by us, 2021 images trigger our adaptive streaming (\secref{adaptive}), and 1783 images pass our fallback checks (\secref{fallback}) and trigger our system to apply fusion deblurring.

\begin{table}
    \centering
    \footnotesize
    \caption{\textbf{Runtime comparisons.} 
        We measure the per-frame latency on $768 \times 768$ image resolution using NVidia Quadro P5000 GPU on a desktop. Note that the Pytorch execution time is measured by synchronizing the GPU and CPU executions~\cite{cudasync}.
    }
    \begin{tabular}{ccc}
        \toprule
        Method & Implementation & Latency (ms) \\
        \midrule
        BurstDeblurring &
        \multirow{2}{*}{TensorFlow} &\multirow{2}{*}{1924} \\
        \cite{aittala2018burst} & & \\
        \midrule
        Deblur-GAN-v2 & 
        \multirow{2}{*}{Pytorch} &\multirow{2}{*}{126} \\
        \cite{kupyn2019deblurgan} & & \\
        \midrule
        UMSN Face Deblurring & 
        \multirow{2}{*}{Pytorch} &\multirow{2}{*}{1601} \\
        \cite{yasarla2020deblurring} & & \\
        \midrule
        MPRNet & 
        \multirow{2}{*}{Pytorch} & \multirow{2}{*}{1653} \\
        \cite{zamir2021multi} & & \\
        \midrule
        MIMO-UNet+ & 
        \multirow{2}{*}{Pytorch} &\multirow{2}{*}{581} \\
        \cite{cho2021rethinking} & & \\
        \midrule
        PVDNet & 
        \multirow{2}{*}{Pytorch} &\multirow{2}{*}{456} \\
        \cite{son2021recurrent} & & \\
        \midrule
        UHDVD & 
        \multirow{2}{*}{Pytorch} &\multirow{2}{*}{277} \\
        \cite{deng2021multi} & & \\
        \midrule
        Ours & TensorFlow & 27 \\
    \end{tabular}
    \label{tab:latency_comparison}
\end{table}

\subsection{Comparisons}
\paragraph{Comparisons with academic approaches.}
We compare our results to the recent deblurring approaches listed below:
\begin{itemize}[-]
\item Single-image: DeblurGAN-v2~\cite{kupyn2019deblurgan}, MPRNet~\cite{zamir2021multi}, and MIMO-UNet+~\cite{cho2021rethinking}.
\item Video-based: UHDVD~\cite{deng2021multi}, PVDNet~\cite{son2021recurrent}.
\item Face-specific: UMSN~\cite{yasarla2020deblurring}.
\item Multi-frame: BurstDeblurring~\cite{aittala2018burst}.
\end{itemize}
Note that the existing approaches are typically trained on 8-bit gamma images, \eg GoPro dataset~\cite{nah2017deep}, while our training images are 16-bit linear before any tone-mappings.
It is non-trivial to re-train or fine-tune existing models with our training data.
Therefore, we use the pre-trained model released by the authors of each method for fair comparisons.

We collect 100 representative images that cover diverse scenes and identities and generate 8-bit RGB images from \wide~raw burst using~\cite{hasinoff2016burst}.
Note that several CNN-based methods in other works require excessive memory usage. 
To avoid out-of-memory, we further crop test images to fusion ROI defined in~\secref{preprocessing} and resize the cropped images to $768 \times 768$ for inference. 
For multi-frame and video-based methods, we create a sequence of 8-bit RGB images by applying~\cite{hasinoff2016burst} to each frame in the burst.
The sequence length is between 5 to 9 frames.
\figref{compare_academic} shows the visual comparisons.
Existing methods typically fail to remove large motion blurs and produce overly smooth results or artifacts such as ringings or halo.
In contrast, our method generates sharper and clearer faces with plausible facial details.
We provide all the visual comparisons in the supplementary material.

For quantitative evaluation, as there are no ground-truth images in our testing dataset, we rely on visual comparisons and no-reference quality metrics using NIMA~\cite{talebi2018nima}, which learns a CNN to predict human perceptual quality\footnote{We use the Inception-v2 model in our experiments.}.
\figref{evaluation} plots the average NIMA scores and average NIMA difference between deblurred and source images of these methods. 
Our result achieves the best perceptual scores among others in~\figref{evaluation}(a).
In \figref{evaluation}(b), existing approaches except PVDNet~\cite{son2021recurrent} show negative NIMA difference on $34.4$ percentile, \ie within 1-standard deviation.
Our method is more robust and achieves consistent perceptual improvement.
Table~\ref{tab:latency_comparison} reports the per-frame inference latency on $768 \times 768$ input images.
We use NVidia Quadro P5000 GPU on a desktop.
Note that the implementation libraries are different for these methods, and our model is optimized for mobile GPU inference.
Since our source code and training data involve proprietary information, we are not able to release them. Instead, we release the test images used for comparisons at the project website\footnote{\website}.

\paragraph{Comparisons with commercial products.}
We further compare our method against commercial products using post-capture editing tools from the Shake Reduction feature in Adobe Photoshop CC 2021 and Samsung Gallery's Remaster function, as well as straight-out-of-camera capture from Apple iPhone 13 Pro.
The first two methods are single-image deblurring, while ours and iPhone are processed at capture time.
There is no software to connect iPhones and Android devices for synchronized photo capture.
Therefore, we use our Android device and iPhone 13 Pro to capture the same moving subject side-by-side until we find the best-synchronized pairs.
In~\figref{compare_commercial}, the iPhone produces blurry results on the fast-moving subject. 
Photoshop Shake Reduction cannot reduce blur effectively and leaves ringing artifacts in the output (bottom row).
Samsung Remaster shows promising deblurred results, but misses the thin edges near the glasses frame (top row) and cannot remove large motion blur (bottom row).
Their results also look overly smooth and lose many facial details. 
It takes about 8 to 10 seconds for the Samsung Remaster function to process a $3072 \times 4080$ image.
Compared to other commercial products, our system generates artifact-free results with the best details and sharpness at low latency.

\begin{table}
    \centering
    \caption{\textbf{Latency overhead.} Compared to a normal single-camera shot, the latency overhead of our system is 463 ms on Google Pixel 6.
    }
    \begin{tabular}{c|c}
        \toprule
        Stage           & Latency (ms) \\
        \midrule
        Pre-processing  &   78.0 \\
        Alignment       &  124.3 \\
        Fusion          &  121.0 \\
        Polyblur        &  139.7 \\
        \midrule
        Total overhead  &  463.0
    \end{tabular}
    \label{tab:latency}
\end{table}

\subsection{System Performance}
\paragraph{Latency and memory.} 
We implement our system on Google Pixel 6 released in 2021. The SoC consists of Mali-G78 GPU, Cortex multi-core, and an NPU. 
The latency overhead is 463ms, which consists of preprocessing, alignment, fusion, and the Polyblur in post-processing, as reported in Table~\ref{tab:latency}.
We exclude merging \wide~ burst raw and tone-mapping from the overhead since they are already required in the single camera processing without our fusion deblur.  Merging \ultrawide~ burst raw is scheduled concurrently with merging \wide~ burst and does not take additional latency. Our system's peak memory usage happens at the FusionNet inference stage, which takes an extra 264 MB compared to the peak memory usage in single camera processing pipeline. 

\paragraph{Power.}
When dual camera streaming at 30 FPS, the power overhead is 468mW, which splits between power rail from \ultrawide~ sensor (332mW) and SoC (136mW) for RAW data transaction between DRAM and the \ultrawide~ camera system. 
On average, our adaptive system streams \ultrawide~ 1.8\% of the camera session time, which results in 8.4mW amortized power cost.

\subsection{Ablation Studies}
In this section, we validate the contributions of key design choices of our fusion algorithm, including the color-consistency loss, synthetic light streaks in training data generation, and compare our method with single-image deblurring.
Additional ablation analysis are provided in the supplementary material, including analysis on the perceptual loss, mask smoothing, Polyblur, optical flow, and how our FusionNet utilizes each input for deblurring.

\begin{figure}
    \centering
    \footnotesize
    \renewcommand{\tabcolsep}{1pt} 
	\renewcommand{\imagewidth}{0.24\linewidth} 
    \newcommand{\addimage}[3]{
        \includegraphics[width=\imagewidth]{figures/ablation/color_loss/#1_source_crop.jpg} &
        \includegraphics[width=\imagewidth]{figures/ablation/color_loss/#1_reference_crop.jpg} &
        \begin{overpic}[width=\imagewidth]{figures/ablation/color_loss/#1_fusion_no_color_loss_crop.jpg}
            \linethickness{1pt}
            \put(#2,#3){\color{yellow}\vector(-1,-1){10}}
        \end{overpic}
        &
        \includegraphics[width=\imagewidth]{figures/ablation/color_loss/#1_fusion_tot_crop.jpg}
	}
    \begin{tabular}{cccc}
        \addimage{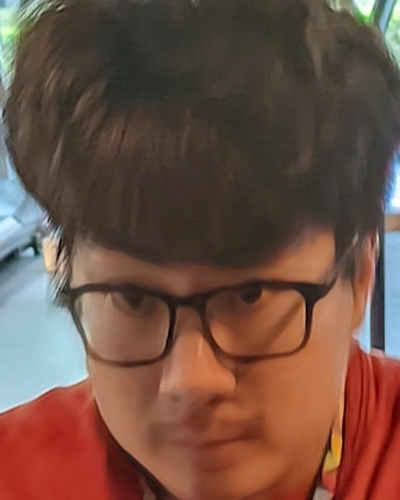}{75}{80} \\
        \addimage{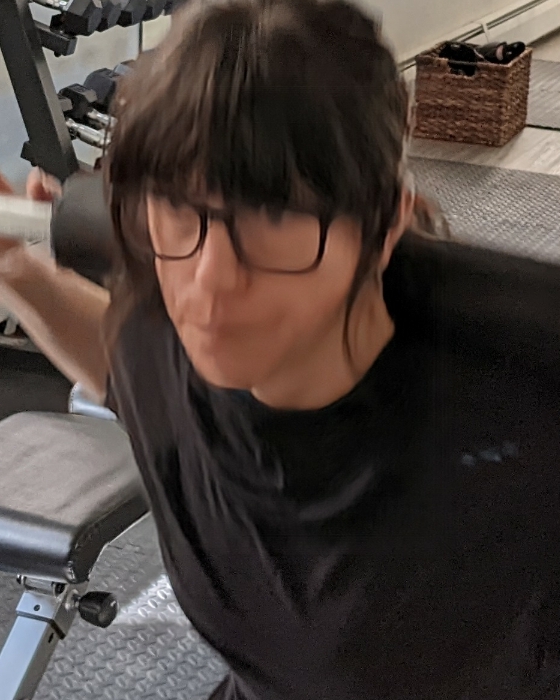}{80}{50} \\
        Source & Reference & Ours w/o $\colorloss$ & Ours 
    \end{tabular}
    \caption{\textbf{Effectiveness of color-consistency loss.} Without the color-consistency loss $\colorloss$, inconsistent color may be transferred from the reference image to the fusion result, \eg green color on hair in both cases (zoom-in to see the details).
    }
    \label{fig:color_loss}
\end{figure}

\paragraph{Color-consistency loss.}
As \wide~ and \ultrawide~ use different camera sensors, there will be inevitable color mismatch even after our color matching and AE synchronization.
Without the color-consistency loss $\colorloss$, inconsistent color may be transferred from the reference image to the fusion result, such as green color on the hair of the two examples in~\figref{color_loss}.
By training with the color-consistency loss, our FusionNet learns to keep the color of the source image after deblurring.

\begin{figure}
    \centering
    \footnotesize
    \renewcommand{\tabcolsep}{1pt} 
	\renewcommand{\imagewidth}{0.24\linewidth} 
    \newcommand{\addimage}[1]{
        \includegraphics[width=\imagewidth]{figures/ablation/single_image/#1_source_crop.jpg} &
        \includegraphics[width=\imagewidth]{figures/ablation/single_image/#1_reference_crop.jpg} &
        \includegraphics[width=\imagewidth]{figures/ablation/single_image/#1_fusion_single_image_crop.jpg} &
        \includegraphics[width=\imagewidth]{figures/ablation/single_image/#1_fusion_tot_crop.jpg}
	}
    \begin{tabular}{cccc}
        \addimage{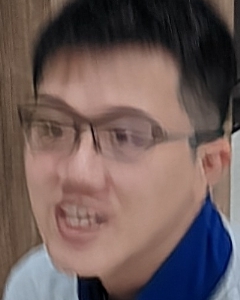} \\
        \addimage{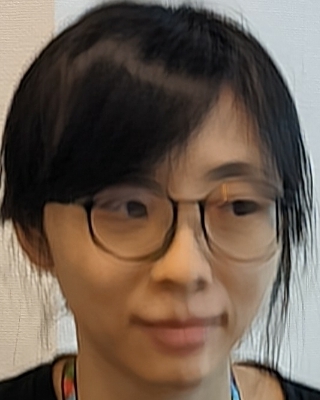} \\
        Source & Reference & Our fusion  & Our fusion \\
        & & w/o reference & with reference
    \end{tabular}
    \caption{\textbf{Effectiveness of reference image.}
    Without the reference image, our single-image deblurring model is not able to remove large motion blur and restore facial details well, leaving undesirable visual artifacts.
    }
    \label{fig:single_image}
\end{figure}

\paragraph{Single-image vs. reference-guided deblurring.}
To validate the importance of the reference image, we train our FusionNet by removing the reference image from the input. The configuration effectively becomes a single-image deblurring approach.
As shown in the third column of~\figref{single_image}, this approach fails to remove large motion blur and produces artifacts on the output.
The eyeglasses in the top row show obvious ringing artifacts, and the face in the bottom row is still significantly blurred.
On the contrary, our dual camera fusion approach restores sharp faces with rich details.

\begin{figure}
    \centering
    \footnotesize
    \renewcommand{\tabcolsep}{1pt} 
	\renewcommand{\imagewidth}{0.32\linewidth} 
    \newcommand{\addimage}[1]{
        \includegraphics[width=\imagewidth]{figures/ablation/highlight/#1_source_crop.jpg} &
        \includegraphics[width=\imagewidth]{figures/ablation/highlight/#1_fusion_no_highlight_training_crop.jpg} &
        \includegraphics[width=\imagewidth]{figures/ablation/highlight/#1_fusion_tot_crop.jpg} 
	}
    \begin{tabular}{ccc}
        \addimage{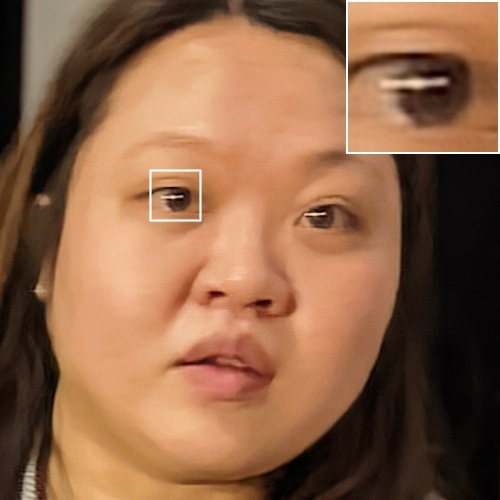} \\
        \addimage{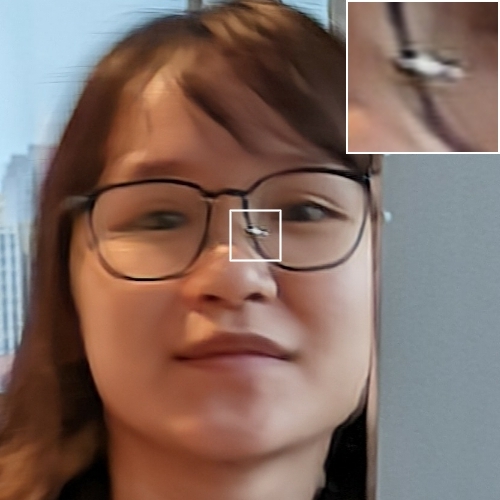} \\
        Source & Ours w/o synthetic lights & Ours
    \end{tabular}
    \caption{\textbf{Effectiveness of synthetic lights streak.} 
        Without adding synthetic highlights to the training data, our model cannot suppress light streaks on eyeballs (top row) or reflection (bottom row), and often results to visible ringing artifacts around the saturated pixels.
    }
    \label{fig:highlight}
\end{figure}

\paragraph{Synthetic light streaks.}
As shown in the second column of~\figref{highlight}, our FusionNet may not be able to remove the motion blur on saturated pixels very well, resulting in visible light streaks or ringing artifacts.
By adding synthetic highlights to the training data, our model can reduce the light streaks on eyeballs or glasses reflection and suppress ringing artifacts, providing more visually pleasing deblurred results.

\subsection{Limitations}
\begin{figure}
    \centering
    \footnotesize
    \renewcommand{\tabcolsep}{1pt} 
	\renewcommand{\imagewidth}{0.32\linewidth} 
    \newcommand{\addimage}[3]{
        \begin{overpic}[width=\imagewidth]{figures/limitation/#1_source_crop.jpg}
            \put (3,3) {\white{\textbf{#2}}}
        \end{overpic} &
        \begin{overpic}[width=\imagewidth]{figures/limitation/#1_reference_crop.jpg}
            \put (3,3) {\white{\textbf{#3}}}
        \end{overpic} &
        \includegraphics[width=\imagewidth]{figures/limitation/#1_fusion_crop.jpg}
	}
    \begin{tabular}{ccc}
        \addimage{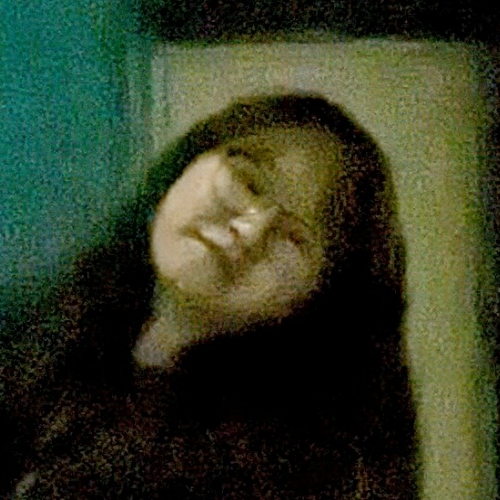}{1/24, ISO 2556}{1/96, ISO 6140} \\
        Source (\wide) & Reference (\ultrawide) & Ours
    \end{tabular}
    \caption{\textbf{Limitation.} Reference quality is critical to our deblurred results. Under extremely low-light conditions, the \ultrawide~ camera suffers from severe sensor noise, leading to objectionable artifacts on the fusion output.
    }
    \label{fig:limitation}
\end{figure}

\begin{figure}
    \centering
    \footnotesize
    \renewcommand{\tabcolsep}{1pt} 
    \renewcommand{\imageheight}{0.255\linewidth}
    \newcommand{\addimage}[1]{
        \begin{tabular}{cccc}
        \begin{overpic}[height=\imageheight]{figures/limitation_small_face/#1_source_full.jpg}
            \linethickness{1pt}
            \put(50,70){\color{red}\vector(-1,-1){10}}
        \end{overpic} &
        \includegraphics[height=\imageheight]{figures/limitation_small_face/#1_source_crop.jpg} &
        \includegraphics[height=\imageheight]{figures/limitation_small_face/#1_reference_crop.jpg} &
        \includegraphics[height=\imageheight]{figures/limitation_small_face/#1_fusion_crop.jpg} \\
        Full-size source & Source & Reference & Ours \\
        \end{tabular}
	}
    \addimage{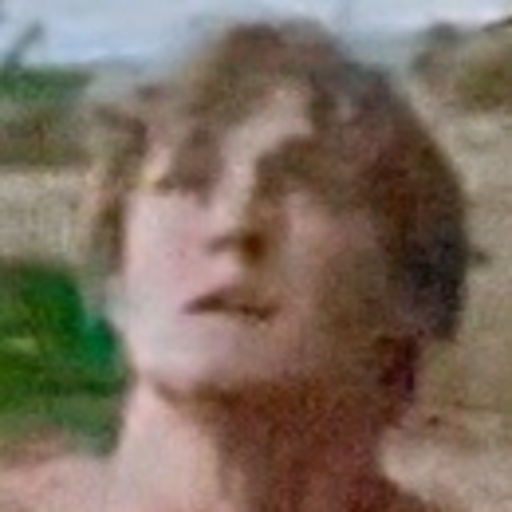}
    \caption{\textbf{Limitation.} If the subject face is too small, \ultrawide~ is not able to capture enough facial details for deblurring.
    }
    \label{fig:limitation_small_face}
\end{figure}

The quality of our fusion results mainly depends on the quality of \ultrawide~ images.
Our method may fail to restore a clear face under the following situations: 1) \textit{extreme low-light}: As the \ultrawide~ image is captured at a faster exposure time of \wide, the \ultrawide~ image may appear excessively noisy under the extreme low-light scenes. %
Our FusionNet may transfer the noise from the reference image and lead to painterly artifacts, as shown in~\figref{limitation}.
2) \textit{Small face}: When the face is too small, \ultrawide~ may not be able to capture facial details for restoration, as shown in~\figref{limitation_small_face}. Our method is not able to recover a sharp face under such a case.
In the above cases, the fallback mechanism in \secref{fallback} would trigger to avoid artifacts. \figref{limitation} and~\ref{fig:limitation_small_face} are created by disabling the fallback mechanism. 

When the face mask includes part of the clothes, the blending boundary may be visible if the clothes have regular texture, as shown in the baby's apron in the first row of~\figref{our_results} .
This issue could be mitigated by applying a stronger mask boundary smoothing or a better face segmentation that contains face regions only.

Finally, our system currently applies deblurring to a single face (the largest face) due to the device computing budget.
A multi-face deblurring solution will require more justifications on system stability and is considered as our future work.

\section{Conclusion}

In this work, we have presented a robust system to address motion blur on faces in portrait photography. 
Our system runs efficiently on Google Pixel 6 smartphones at the capturing time and produces visually more pleasing results than state-of-the-art single-image, multi-frame, video, and face-specific deblurring methods through extensive evaluations. 
By leveraging the dual camera system available on major premium phones, we have shown that combining optical flow for alignment, residual UNet for image fusion, and training on synthetic data can address the challenges that would be difficult to approach through a single camera setup.  
We have specified a list of system requirements including adaptive stream and camera ISP to make the dual camera fusion practical on mobile devices. We hope our work encourages further progress on reference-based image deblurring.

\section*{Appendix - Additional Ablation Studies}

\begin{figure}
    \centering
    \footnotesize
    \renewcommand{\tabcolsep}{1pt} 
	\renewcommand{\imagewidth}{0.32\linewidth} 
    \newcommand{\addimage}[5]{
        \includegraphics[width=\imagewidth]{figures/ablation/perceptual_loss/#1_source_crop.jpg} &
        \begin{overpic}[width=\imagewidth]{figures/ablation/perceptual_loss/#1_fusion_no_perceptual_crop.jpg}
            \linethickness{1pt}
            \put(#2,#3){\color{yellow}\vector(#4,#5){10}}
        \end{overpic} &
        \includegraphics[width=\imagewidth]{figures/ablation/perceptual_loss/#1_fusion_tot_crop.jpg}
	}
    \begin{tabular}{ccc}
        \addimage{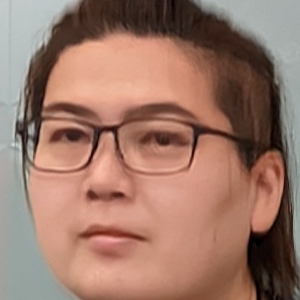}{8}{28}{1}{1} \\
        \addimage{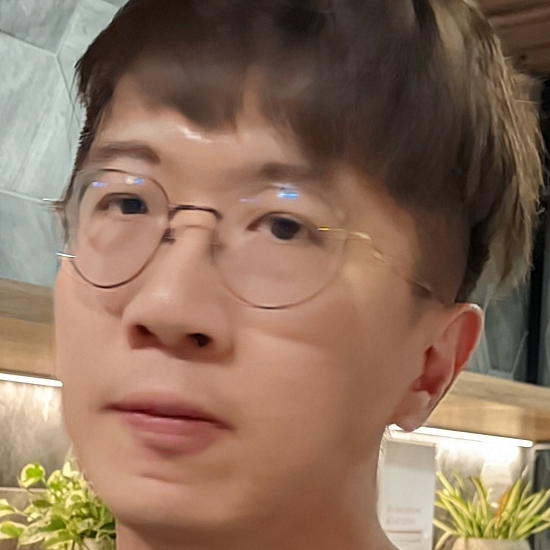}{70}{32}{-1}{1}\\
        (a) Source & (b) Ours w/o $\vggloss$ & (c) Ours 
    \end{tabular}
    \vspace{-2mm}
    \caption{\textbf{Effectiveness of perceptual loss.} Without the perceptual loss $\vggloss$, our FusionNet tends to generate overly-smooth results and leaves ringing artifacts or blurriness on eyeglasses frames.
    }
    \label{fig:perceptual_loss}
\end{figure}

\paragraph{Perceptual loss.}
Without the perceptual loss $\vggloss$, our deblurred results often look overly smooth.
Since motion blur cannot be completely removed, the results contain ringing or blurriness around the eyeglasses frame, as shown in~\figref{perceptual_loss}(b).
By training with the perceptual loss, our FusionNet generates sharper results with richer facial details, as shown in~\figref{perceptual_loss}(c).

\begin{figure}
    \centering
    \footnotesize
    \renewcommand{\tabcolsep}{1pt} 
	\renewcommand{\imagewidth}{0.32\linewidth} 
    \newcommand{\addimage}[1]{
        \begin{tabular}{ccc} &
            \includegraphics[width=\imagewidth]{figures/ablation/mask_smoothing/#1/source_mask_without_smoothing.jpg} &
            \includegraphics[width=\imagewidth]{figures/ablation/mask_smoothing/#1/source_mask_with_smoothing.jpg} \\
            & Mask w/o smoothing & Mask with smoothing \\
            \includegraphics[width=\imagewidth]{figures/ablation/mask_smoothing/#1/source.jpg} &
            \includegraphics[width=\imagewidth]{figures/ablation/mask_smoothing/#1/fusion_without_smoothing.jpg} &
            \includegraphics[width=\imagewidth]{figures/ablation/mask_smoothing/#1/fusion_with_smoothing.jpg} \\
            Source & Ours w/o mask smoothing & Ours \\
        \end{tabular}
	}
    \addimage{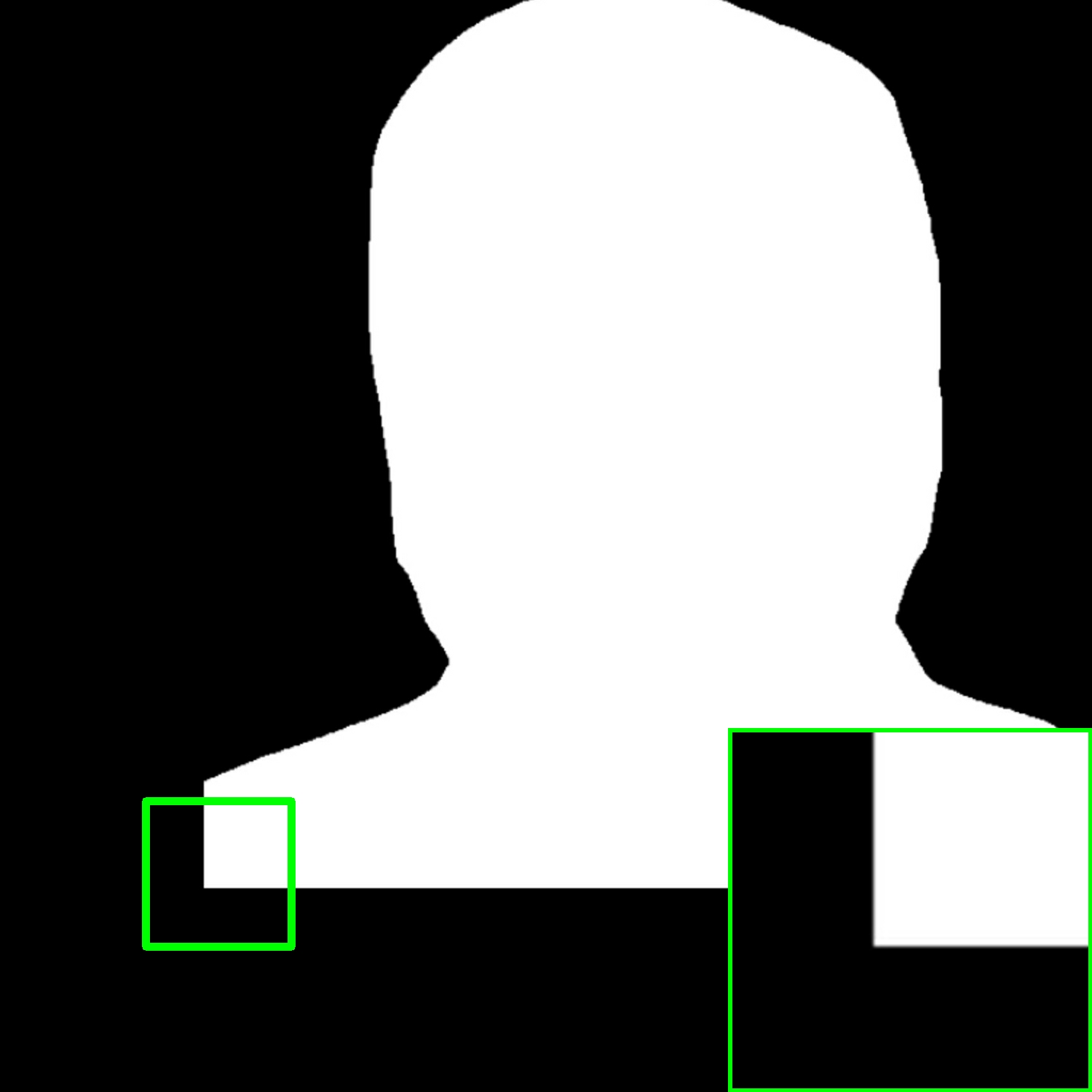}
    \vspace{-2mm}
    \caption{\textbf{Effectiveness of mask smoothing.}
    Without mask boundary smoothing, the result shows a visible border between the face and clothes.
    }
    \label{fig:mask_smoothing}
\end{figure}

\paragraph{Face mask smoothing.}
We apply an alpha blending to blend our deblurred face with the full-resolution source image.
Without the mask boundary smoothing, a clear border is visible in the deblurred image, as shown in~\figref{mask_smoothing}.
The smoothed mask for alpha blending helps to avoid such a sharp transition.

\begin{figure}
    \centering
    \footnotesize
    \renewcommand{\tabcolsep}{1pt} 
	\renewcommand{\imagewidth}{0.32\linewidth} 
    \newcommand{\addimage}[1]{
        \includegraphics[width=\imagewidth]{figures/ablation/polysharp/#1_source_crop.jpg} &
        \includegraphics[width=\imagewidth]{figures/ablation/polysharp/#1_fusion_no_polysharp_crop.jpg} &
        \includegraphics[width=\imagewidth]{figures/ablation/polysharp/#1_fusion_tot_crop.jpg} 
	}
    \begin{tabular}{ccc}
        \addimage{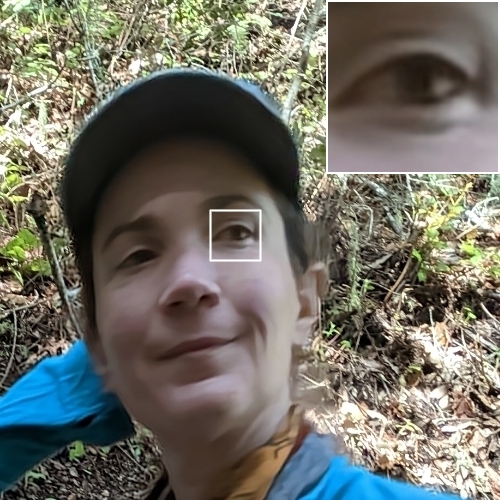} \\
        Source & Ours w/o Polyblur & Ours
    \end{tabular}
    \vspace{-2mm}
    \caption{\textbf{Effectiveness of Polyblur.} Polyblur removes mild blur and improves the details sharpness of our fusion results.
    }
    \label{fig:polysharp}
\end{figure}

\paragraph{Polyblur.}
As our fusion results may look overly-smooth, we adopt the Polyblur~\cite{delbracio2021polyblur} method to enhance the sharpness.
Polyblur mainly removes mild blur caused by lens blur or small out-of-focus, while our FusionNet can reduce large motion blur caused by subject motion.
\figref{polysharp} shows an example that Polyblur improves the detail sharpness around the eyes.

\begin{figure*}
    \centering
    \footnotesize
    \renewcommand{\tabcolsep}{1pt} 
	\renewcommand{\imagewidth}{0.24\linewidth} 
    \newcommand{\addimage}[1]{
        \begin{tabular}{cccc}
        \includegraphics[width=\imagewidth]{figures/ablation/pwcnet/#1/source.jpg}  &
        \includegraphics[width=\imagewidth]{figures/ablation/pwcnet/#1/warped_reference_and_flow_pwcnet_full.jpg} &
        \includegraphics[width=\imagewidth]{figures/ablation/pwcnet/#1/warped_reference_and_flow_pwcnet_small.jpg} &
        \includegraphics[width=\imagewidth]{figures/ablation/pwcnet/#1/warped_reference_and_flow_pwcnet_mobile.jpg} \\
        (a) Source & 
        (b) Warped-reference and flow from & 
        (c) Warped-reference and flow from & 
        (d) Warped-reference and flow from \\
        &
        PWC-Net (full-size input) & 
        PWC-Net($\frac{1}{4}\times$-size input) & 
        optimized PWC-Net ($\frac{1}{4}\times$-size input) \\
        \includegraphics[width=\imagewidth]{figures/ablation/pwcnet/#1/reference.jpg} &
        \includegraphics[width=\imagewidth]{figures/ablation/pwcnet/#1/fusion_pwcnet_full.jpg} &
        \includegraphics[width=\imagewidth]{figures/ablation/pwcnet/#1/fusion_pwcnet_small.jpg} &
        \includegraphics[width=\imagewidth]{figures/ablation/pwcnet/#1/fusion_pwcnet_mobile.jpg} \\
        (e) Reference & 
        (f) Deblurred result from & 
        (g) Deblurred result from & 
        (h) Deblurred result from \\
        &
        PWC-Net (full-size input) & 
        PWC-Net($\frac{1}{4}\times$-size input) & 
        optimized PWC-Net ($\frac{1}{4}\times$-size input)
        \end{tabular}
	}
    \addimage{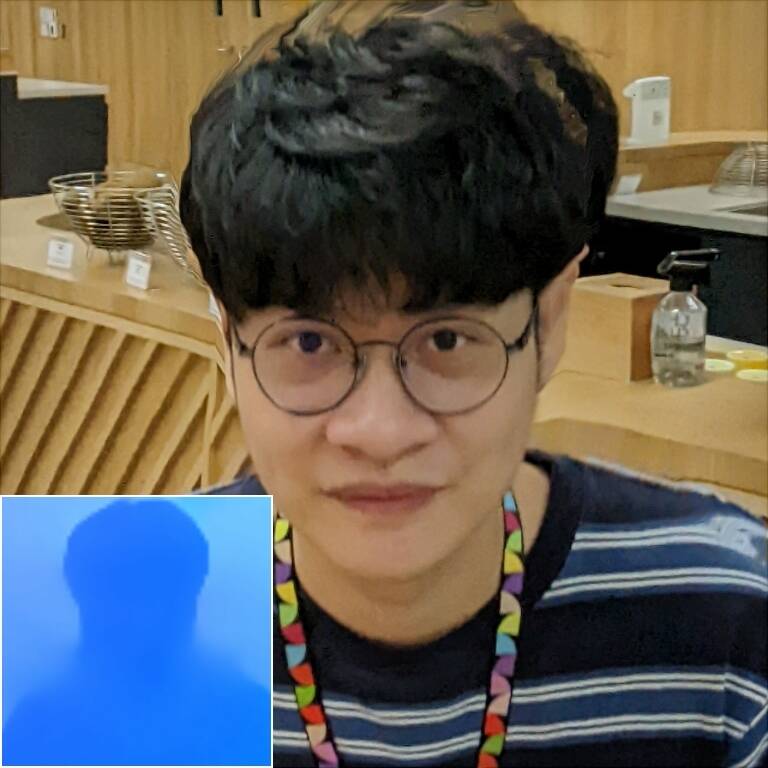}
    \vspace{-2mm}
    \caption{\textbf{Comparisons on PWC-Net.}
        The original PWC-Net may generate inaccurate flow due to the resolution mismatch between training and inference images.
        By estimating flow from $\frac{1}{4}\times$ resolution input images, we obtain more robust flow to align face regions more accurately.
        We further optimize the architecture to reduce latency and memory footprint. 
        Our optimized PWC-Net can generate similar flow and fusion results as the original PWC-Net. 
    }
    \label{fig:pwcnet}
    \vspace{-1mm}
\end{figure*}

\paragraph{Optical flow.}
The original PWC-Net may generate noisy flow and cannot properly align the face due to the resolution mismatch between training and inference images (the second column of~\figref{pwcnet}).
By estimating flow from $\frac{1}{4}\times$ resolution input images, we obtain a more robust flow to align face more accurately, as shown in the third column of~\figref{pwcnet}.
We further optimize the architecture to reduce latency and memory footprint. 
As shown in the fourth column of~\figref{pwcnet}, the optimized PWC-Net generates similar flow and fusion results as the original PWC-Net.

\begin{figure*}
    \centering
    \footnotesize
    \renewcommand{\tabcolsep}{1pt} 
	\renewcommand{\imagewidth}{0.24\linewidth} 
    \begin{tabular}{cccc}
        \includegraphics[width=\imagewidth]{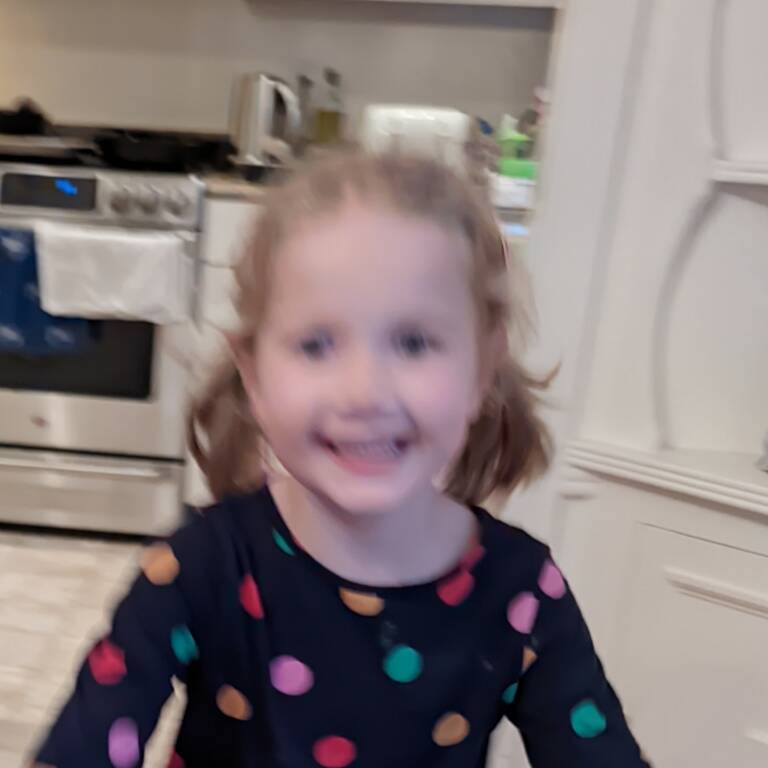} &
        \includegraphics[width=\imagewidth]{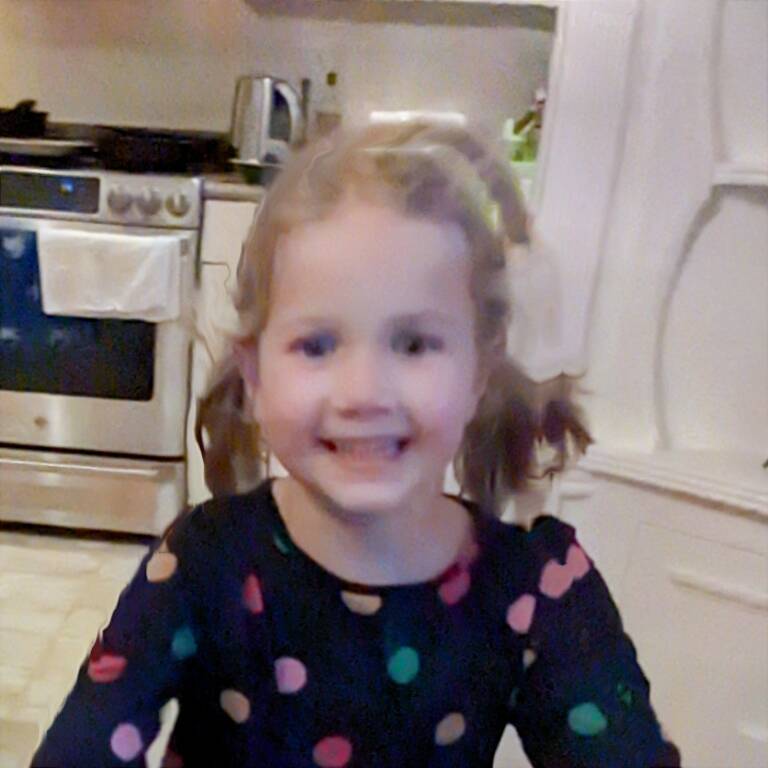} &
        \includegraphics[width=\imagewidth]{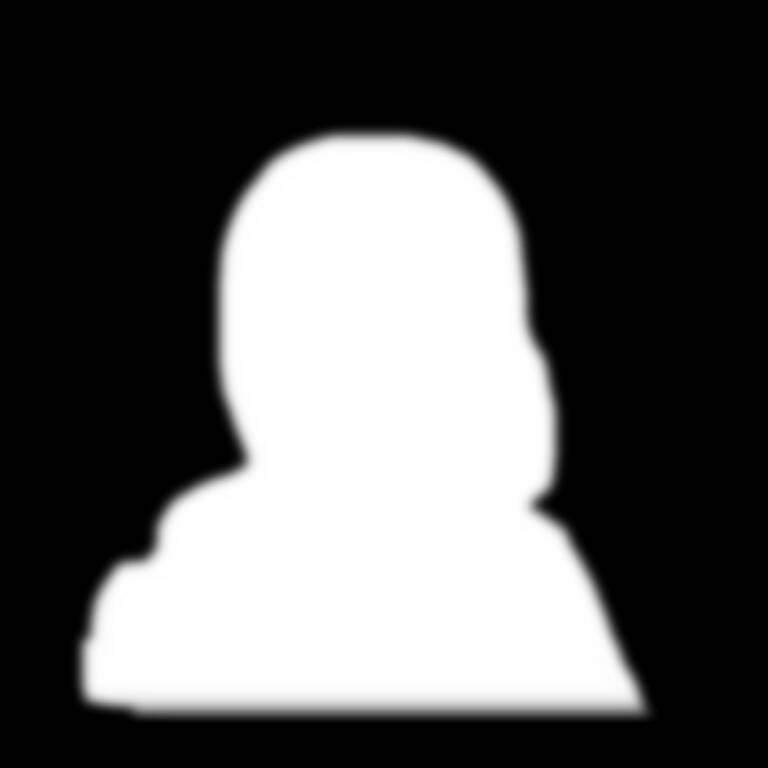} &
        \includegraphics[width=\imagewidth]{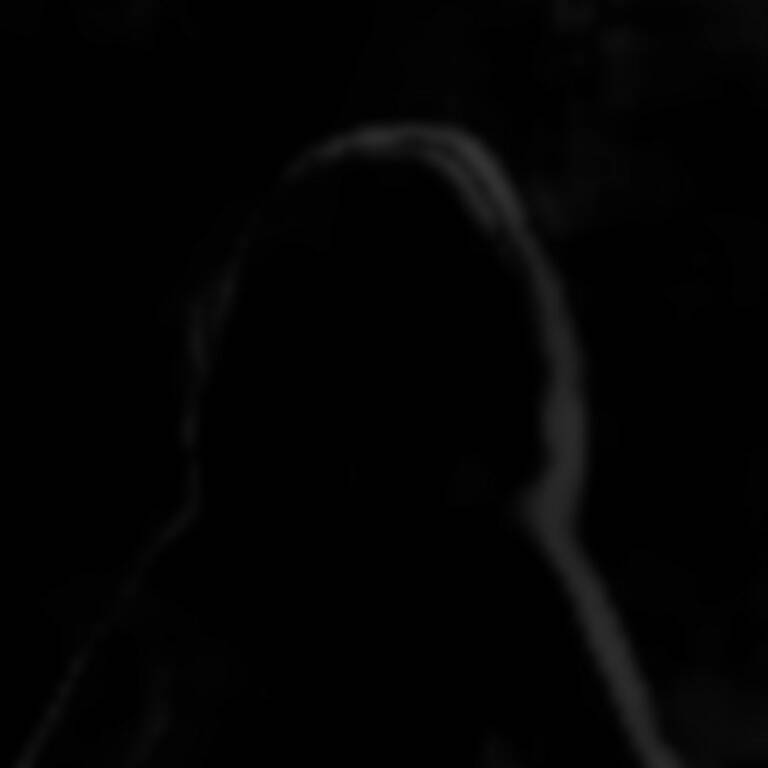} \\
        (a) Source & (b) Warped reference & (c) Face mask & (d) Occlusion mask \\
        \includegraphics[width=\imagewidth]{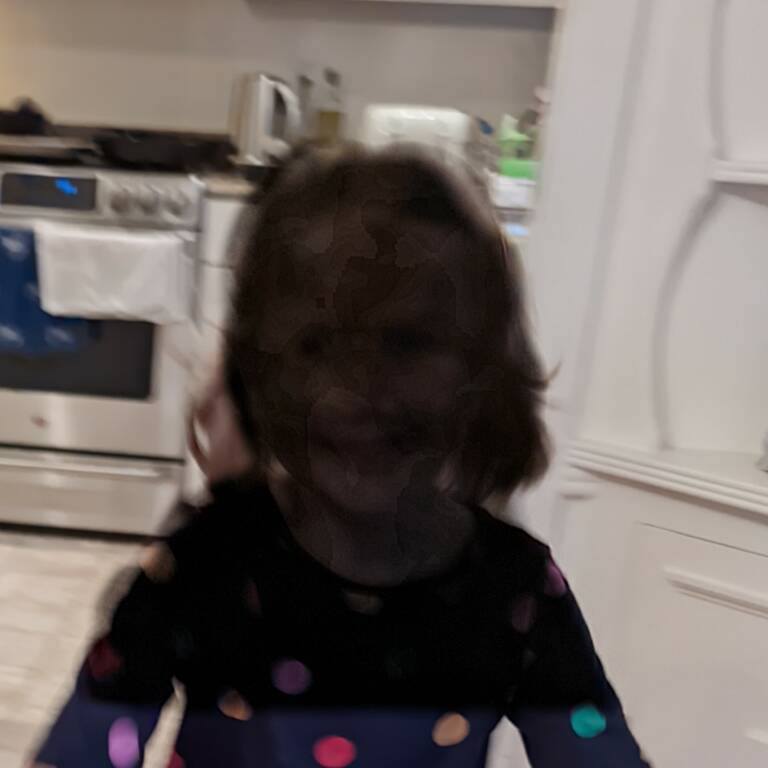} &
        \includegraphics[width=\imagewidth]{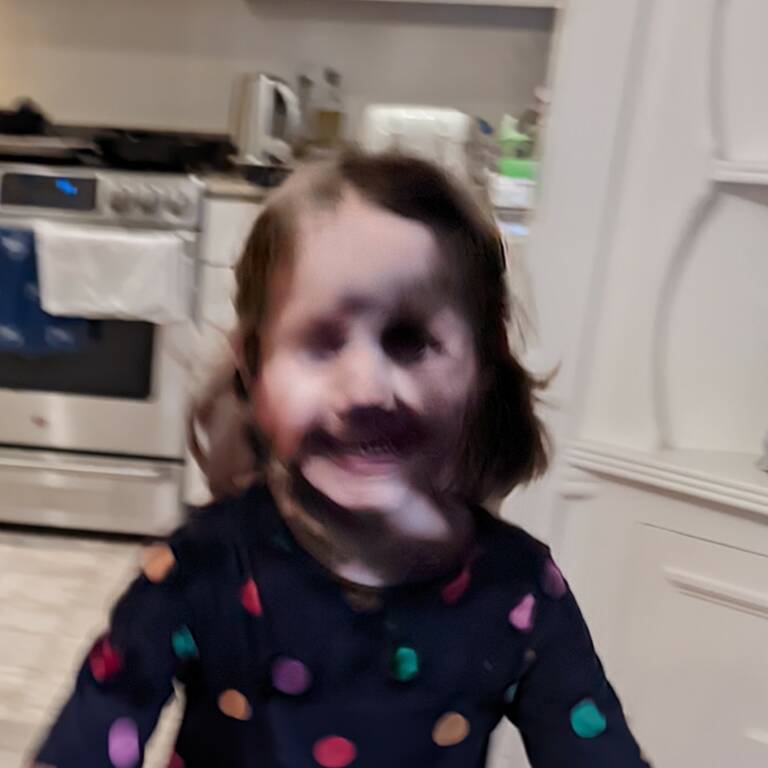} &
        \includegraphics[width=\imagewidth]{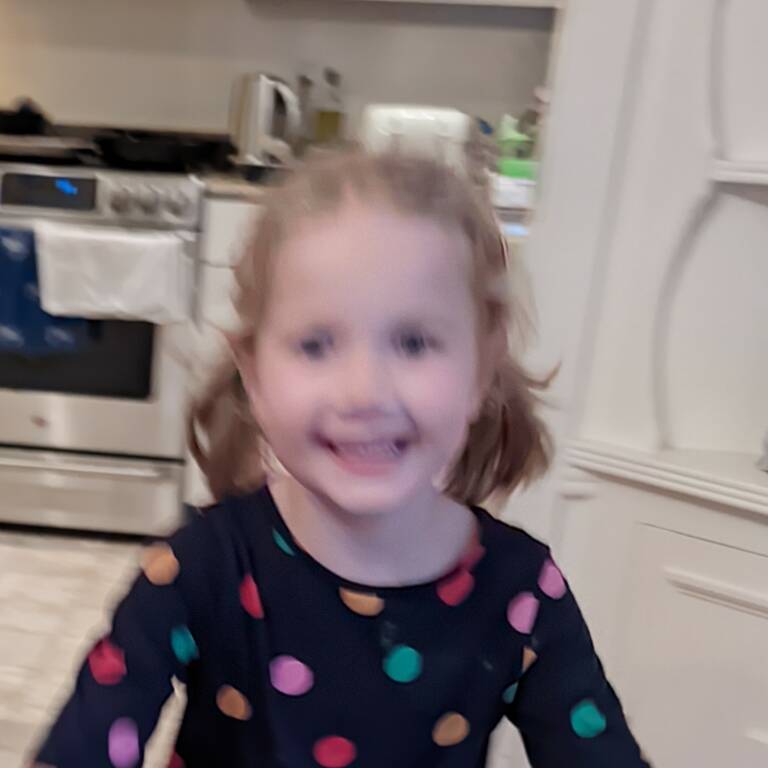} &
        \includegraphics[width=\imagewidth]{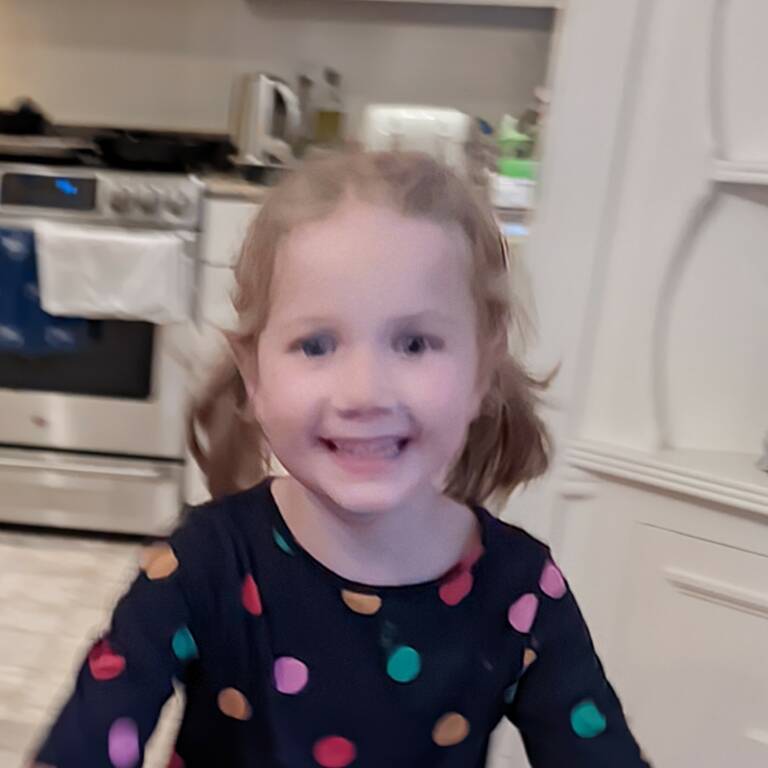}
        \\
        (e) Ours w/o source & (f) Ours w/o reference & (g) Ours w/o face mask & (h) Ours
    \end{tabular}
    \vspace{-2mm}
    \caption{\textbf{Analysis on the FusionNet.}  We apply our FusionNet by setting one of the input tensors to zeros.
    Without the source or reference image, the FusionNet generates unnatural result that does not contain any facial details.
    Without the face mask, the FusionNet does not apply deblurring at all.
    }
    \vspace{-1mm}
    \label{fig:fusion_net}
\end{figure*}

\begin{figure}[h!]
    \centering
    \footnotesize
    \renewcommand{\tabcolsep}{1pt} 
	\renewcommand{\imagewidth}{0.32\linewidth} 
    \newcommand{\addimage}[5]{
        \includegraphics[width=\imagewidth]{figures/ablation/occlusion/#1_source.jpg} &
        \begin{overpic}[width=\imagewidth]{figures/ablation/occlusion/#1_warped_reference.jpg}
            \linethickness{1pt}
            \put(#2,#3){\color{yellow}\vector(#4,#5){10}}
        \end{overpic} &
        \includegraphics[width=\imagewidth]{figures/ablation/occlusion/#1_source_mask.jpg} \\
        Source & Warped-reference & Face mask \\
        \includegraphics[width=\imagewidth]{figures/ablation/occlusion/#1_occlusion_mask.jpg} &
        \begin{overpic}[width=\imagewidth]{figures/ablation/occlusion/#1_fusion_no_occlusion_mask.jpg}
            \linethickness{1pt}
            \put(#2,#3){\color{yellow}\vector(#4,#5){10}}
        \end{overpic} &
        \begin{overpic}[width=\imagewidth]{figures/ablation/occlusion/#1_fusion.jpg}
            \linethickness{1pt}
            \put(#2,#3){\color{yellow}\vector(#4,#5){10}}
        \end{overpic}
        \\
        Occlusion mask & Ours w/o occlusion mask & Ours
	}
    \begin{tabular}{ccc}
        \addimage{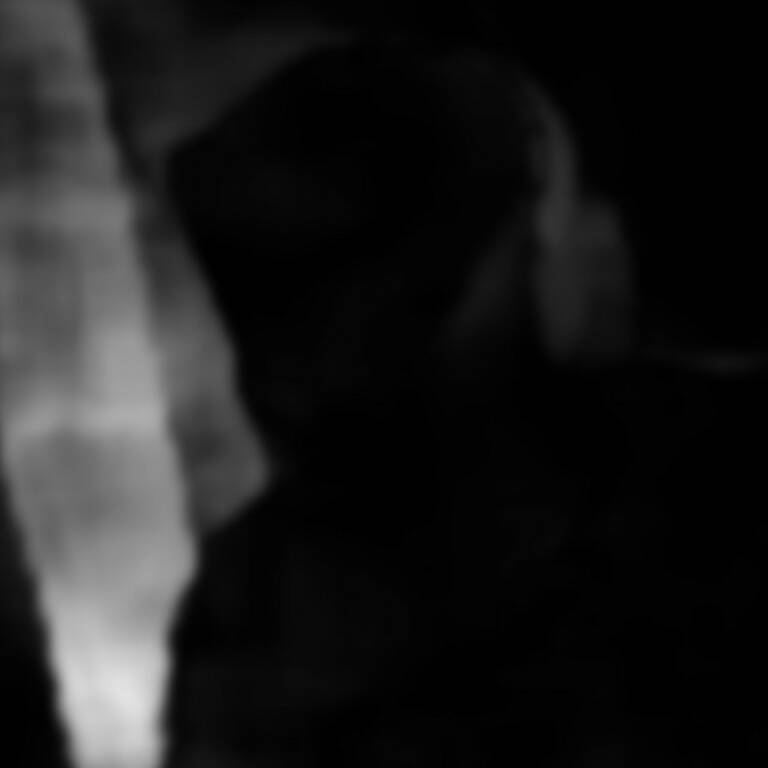}{8}{40}{1}{1} \\
    \end{tabular}
    \caption{\textbf{Effectiveness of occlusion mask.} The occlusion mask can avoid transferring warping artifacts from the warped reference image to the result, especially around the face boundary or background.
    }
    \label{fig:occlusion}
\end{figure}

\paragraph{Understanding FusionNet.}
To understand how our FusionNet utilizes source, reference, face mask, and occlusion mask, we remove one of the input tensors by setting the tensor to zeros, and show the results in~\figref{fusion_net} and~\ref{fig:occlusion}.
We observe the following from this experiment:
\begin{compactenum}
\item Without the source image, the result looks all black without any facial details (\figref{fusion_net}(e)).
\item Without the reference image, FusionNet smears facial details aggressively and generates an unnatural result (\figref{fusion_net}(f)).
\item Without the face mask, FusionNet does not deblur the source image at all (\figref{fusion_net}(g)).
\item The contribution of the occlusion mask is mainly at face boundaries or background. Without the occlusion mask, the face boundary may look sharper, e.g., the edge of face mask in~\figref{occlusion}. However, some warping artifacts on the background could be fused to the output when the face mask falsely includes some background pixels, e.g., the straight lines in the wall of~\figref{occlusion}.
\end{compactenum}
From our analysis, the FusionNet does not blindly copy everything from the reference image to the output. 
Instead, the FusionNet learns to apply fusion only when the source and reference have correspondences, \eg the same face, within the face mask.
While using occlusion mask could sacrifice the sharpness on face boundary, it is crucial to prevent any unnatural artifacts in the results.
Minimizing artifacts is an important goal of our system, and we are willing to be conservative in the amount of sharpening. 

\begin{acks}
This work would be impossible without close collaboration between Android Pixel Camera and Google Research teams.
We particularly thank David Massoud, Gabriel Nava, Blossom Chiang, Firman Prayoga, Xu Han, Sam Chang, Lida Wang, Peter Liu, Navinprashath Rajagopal, Jinglun Gu, Gary Sun, and Bhushan Mondkar for their product and engineering contributions on integrating the dual fusion system into Pixel 6 and Google Camera App.
We thank Jinwei Yuan, Alex Hong, Sam Hasinoff, Dillon Sharlet, Kiran Murthy, Mauricio Delbracio, Damien Kelly, and Peyman Milanfar for their supports and constructive feedback on algorithm development and HDR+ integration.
We thank Karl Rasche, Andy Wang, Tianfan Xue, Bart Wronski, and Janne Kontkanen for their helpful discussions on the paper.
We also thank Hsin-Fu Wang, Lala Hsieh, Yun-Wen Wang, Domi Yeh, Cort Muller, Michael Milne, Nicholas Wilson, James Adamson, Christopher Farro for the effort on data collection and image quality reviewing.
Finally, we thank all the photography models in this work for supporting data collection.

\end{acks}

\bibliographystyle{ACM-Reference-Format}
\bibliography{references}

\end{document}